\documentclass{article}

\usepackage[preprint]{neurips_2025}

\usepackage[utf8]{inputenc} 
\usepackage[T1]{fontenc}    
\usepackage{hyperref}       
\usepackage{url}            
\usepackage{booktabs}       
\usepackage{amsfonts}       
\usepackage{nicefrac}       
\usepackage{microtype}      
\usepackage{xcolor}         

\usepackage{hyperref}
\definecolor{Green}{HTML}{228B22}

\usepackage{amsmath}
\usepackage{amssymb}
\usepackage{mathtools}
\usepackage{amsthm}

\usepackage[capitalize,noabbrev]{cleveref}

\theoremstyle{plain}
\newtheorem{theorem}{Theorem}[section]

\newtheorem{lemma}[theorem]{Lemma}

\theoremstyle{definition}

\theoremstyle{remark}

\usepackage[textsize=tiny]{todonotes}

\usepackage{microtype}
\usepackage{pifont} 

\usepackage{graphicx}
\usepackage{subfigure}
\usepackage{booktabs} 

\usepackage[]{xcolor}
\definecolor{ForestGreen}{HTML}{228B22}
\usepackage{microtype}

\newcommand{\revision}[1]{\textcolor{black}{#1}}

\usepackage{graphicx}
\DeclareGraphicsExtensions{.pdf,.jpeg,.png}
\graphicspath{{Images/}} 
\usepackage{subfigure}
\usepackage{rotating} 
\usepackage{dblfloatfix} 

\usepackage{booktabs} 
\usepackage{multirow}
\usepackage{afterpage} 
\usepackage[normalem]{ulem} 
\usepackage{placeins} 

\usepackage{diagbox}
\usepackage[]{amsmath} 
\usepackage{amssymb} 
\usepackage{mathtools}
\usepackage{extarrows} 
\usepackage{amsfonts}
\usepackage{mathrsfs} 
\usepackage{bm} 
\usepackage{array} 
\usepackage{relsize} 

\usepackage[toc,page,header]{appendix}
\usepackage{minitoc}

\usepackage{hyperref}
\usepackage{url}

\usepackage{caption}

\newcommand{\pd}{p_d}

\newcommand{\pg}{p_g}
\newcommand{\pz}{p_{\bm{z}}}

\newcommand{\bmd}{\bm{d}}
\newcommand{\bme}{\bm{e}}

\newcommand{\bmg}{\bm{g}}

\newcommand{\bmy}{\bm{y}}

\newcommand{\bmW}{\bm{W}}
\newcommand{\bmX}{\bm{X}}
\newcommand{\bmY}{\bm{Y}}

\newcommand{\mcalD}{\mathcal{D}}
\newcommand{\mcalE}{\mathcal{E}}

\newcommand{\mcalN}{\mathcal{N}}

\newcommand{\mcalP}{\mathcal{P}}

\newcommand{\mcalW}{\mathcal{W}}
\newcommand{\mcalX}{\mathcal{X}}
\newcommand{\mcalY}{\mathcal{Y}}
\newcommand{\mcalZ}{\mathcal{Z}}

\newcommand{\rmd}{\mathrm{d}}

\newcommand{\rmJ}{\mathrm{J}}

\newcommand{\rmT}{\mathrm{T}}

\newcommand{\mbbE}{\mathbb{E}}

\newcommand{\mbbI}{\mathbb{I}}

\newcommand{\mbbR}{\mathbb{R}}

\newcommand{\mfrakC}{\mathfrak{C}}

\newcommand{\w}{\bm{w}}

\newcommand{\x}{\bm{x}}

\newcommand{\y}{\bm{y}}
\newcommand{\z}{\bm{z}}

\newcommand{\loss}{\mathcal{L}}

\newcommand{\fracpartial}[2]{\frac{\partial #1}{\partial  #2}}
 \newcommand{\DataScore}[1]{ \nabla_{#1} \ln \left(\pd (#1)\right)}

\newcommand{\Esub}{\mathop{\mathbb{E}}}

\usepackage{array}
\newcolumntype{P}[1]{>{\centering\arraybackslash}p{#1}}

\makeatletter
\newcommand{\subalign}[1]{%
  \vcenter{%
    \Let@ \restore@math@cr \default@tag
    \baselineskip\fontdimen10 \scriptfont\tw@
    \advance\baselineskip\fontdimen12 \scriptfont\tw@
    \lineskip\thr@@\fontdimen8 \scriptfont\thr@@
    \lineskiplimit\lineskip
    \ialign{\hfil$\m@th\scriptstyle##$&$\m@th\scriptstyle{}##$\hfil\crcr
      #1\crcr
    }%
  }%
}
\makeatother

\usepackage{algorithmic}

\title{Insights into Closed-form IPM-GAN Discriminator Guidance for Diffusion Modeling}

\author{%
 Aadithya Srikanth\thanks{Denotes equal contribution.}\(^{*~1}\) \\
School of Electrical and Computer Engineering\\
 Purdue University College of Engineering\\
 West Lafayette, Indiana, USA \\
 \texttt{srikanth.aadithya@gmail.com} \\
 \And
 Siddarth Asokan\(^{*~2}\) \\
 Microsoft Research\\
 \#9 VIGYAN, Lavelle Road, \\
 Bengaluru - 560001, India\\
 \texttt{siddarth.asokan@microsoft.com} \\
 \And
 Nishanth Shetty \\
Department of Electrical Engineering\\
 Indian Institute of Science\\
 Bengaluru - 560012, India\\
 \texttt{nishanths@iisc.ac.in} \\
 \And
 Chandra Sekhar Seelamantula\\
 Department of Electrical Engineering\\
 Indian Institute of Science\\
 Bengaluru - 560012, India\\
 \texttt{css@iisc.ac.in} \\
}

\begin{document}

\maketitle

\doparttoc 
\faketableofcontents 

\begin{abstract}
Diffusion models are a state-of-the-art generative modeling framework that transform noise to images via Langevin sampling, guided by the score, which is the gradient of the logarithm of the data distribution. Recent works have shown empirically that the generation quality can be improved when guided by classifier network, which is typically the discriminator trained in a generative adversarial network (GAN) setting. In this paper, we propose a theoretical framework to analyze the effect of the GAN discriminator on Langevin-based sampling, and show that the IPM-GAN optimization can be seen as one of \textit{smoothed score-matching}, wherein the scores of the data and the generator distributions are convolved with the kernel function associated with the IPM. The proposed approach serves to unify score-based training and optimization of IPM-GANs. Based on these insights, we demonstrate that closed-form kernel-based discriminator guidance, results in improvements (in terms of CLIP-FID and KID metrics) when applied atop baseline diffusion models. We demonstrate these results on the denoising diffusion implicit model (DDIM) and latent diffusion model (LDM) settings on various standard datasets. We also show that the proposed approach can be combined with existing accelerated-diffusion techniques to improve latent-space image generation. 

\end{abstract}

\footnotetext[1]{~Work done as a Research Assistant at the Spectrum Lab, Department of Electrical Engineering, Indian Institute of Science, Bengaluru - 560012.}
\footnotetext[2]{~Corresponding Author. Work done during Ph.D. studentship at the Robert Bosch Center for Cyber-Physical Systems, Indian Institute of Science, Bengaluru - 560012.}

\section{Introduction} \label{Sec:Intro}

Generative modeling is the process of learning the underlying distribution of data with the aim of generating new unseen samples from the underlying distribution. Over the past few years, diffusion models~\citep{NCSN19,DDPM20} have become the {\it de facto} approach for generative modeling. Diffusion models treats image generation as a denoising process, and models the transformation by means of a stochastic differential equation (SDE)~\citep{NCSNv220}. The sampling process involves learning the denoising function, or equivalently, the gradient of the logarithm of the data distribution, known as the \textit{score}~\citep{ScoreMatching05}, and subsequently discretizing the SDE. Diffusion models achieve state-of-the-art performance for image generation~\citep{DiscGuidance23,Innate24}. Prior to diffusion models, generative adversarial networks (GANs,~\citet{SGAN14}) were the most popular framework for image generation, owing to their superior single-step sampling performance~\citep{ADAStyleGAN20, StyleGAN321,StyleGANXL22}. As shown by ~\citet{DiscGuidance23}, standard GANs (SGANs)~\citep{SGAN14} and diffusion models can be unified, wherein the gradients of an SGAN discriminator can improve the score. Considering this setting, we develop strong foundations to IPM-GAN-based discriminator guidance for diffusion. \par

\begin{figure*}[!t]
\begin{center}
  \begin{tabular}[!t]{P{.45\linewidth}|P{.45\linewidth}}
  CelebA-HQ & FFHQ \\
\includegraphics[width=0.99\linewidth]{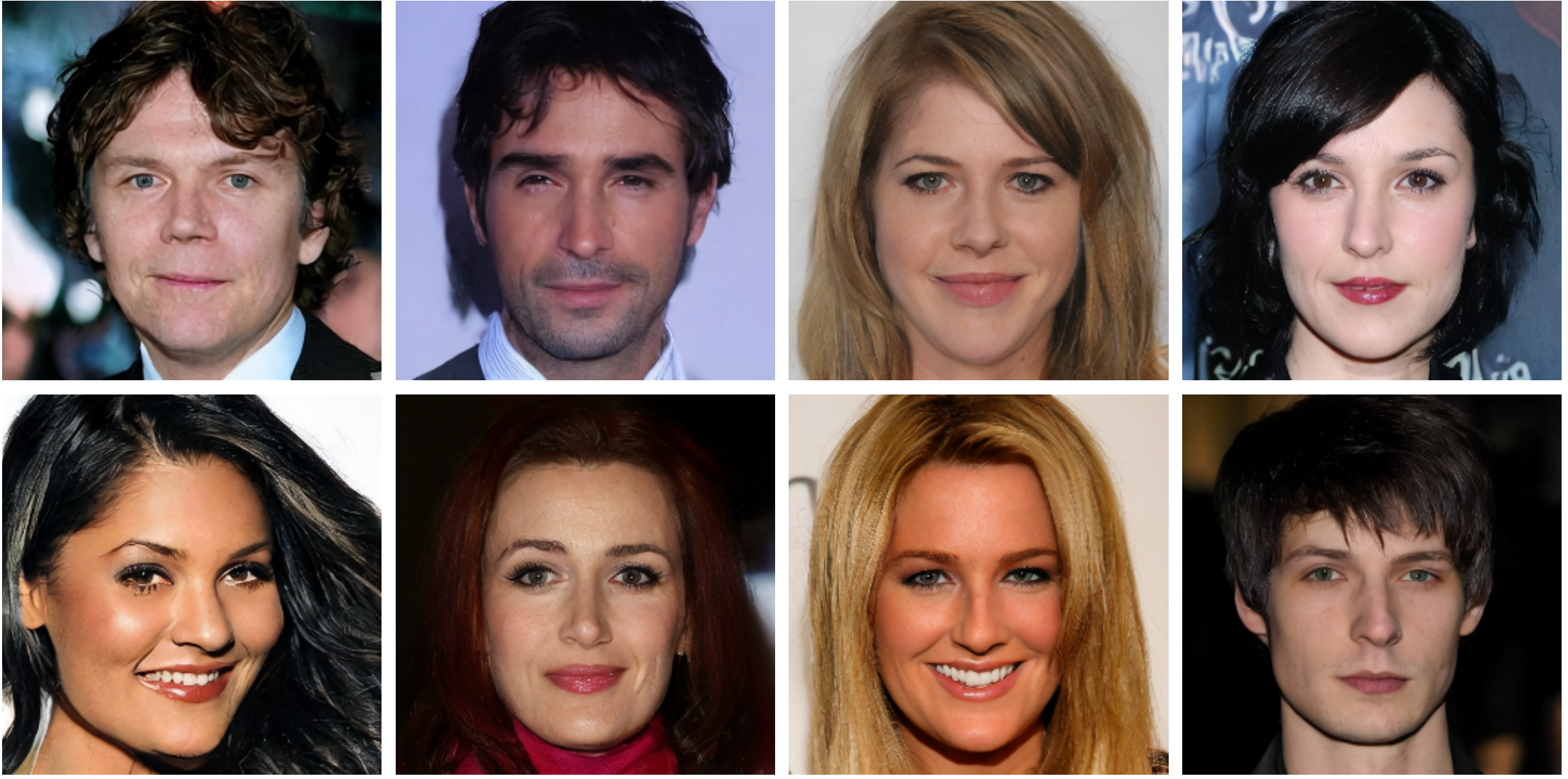} & \includegraphics[width=0.99\linewidth]{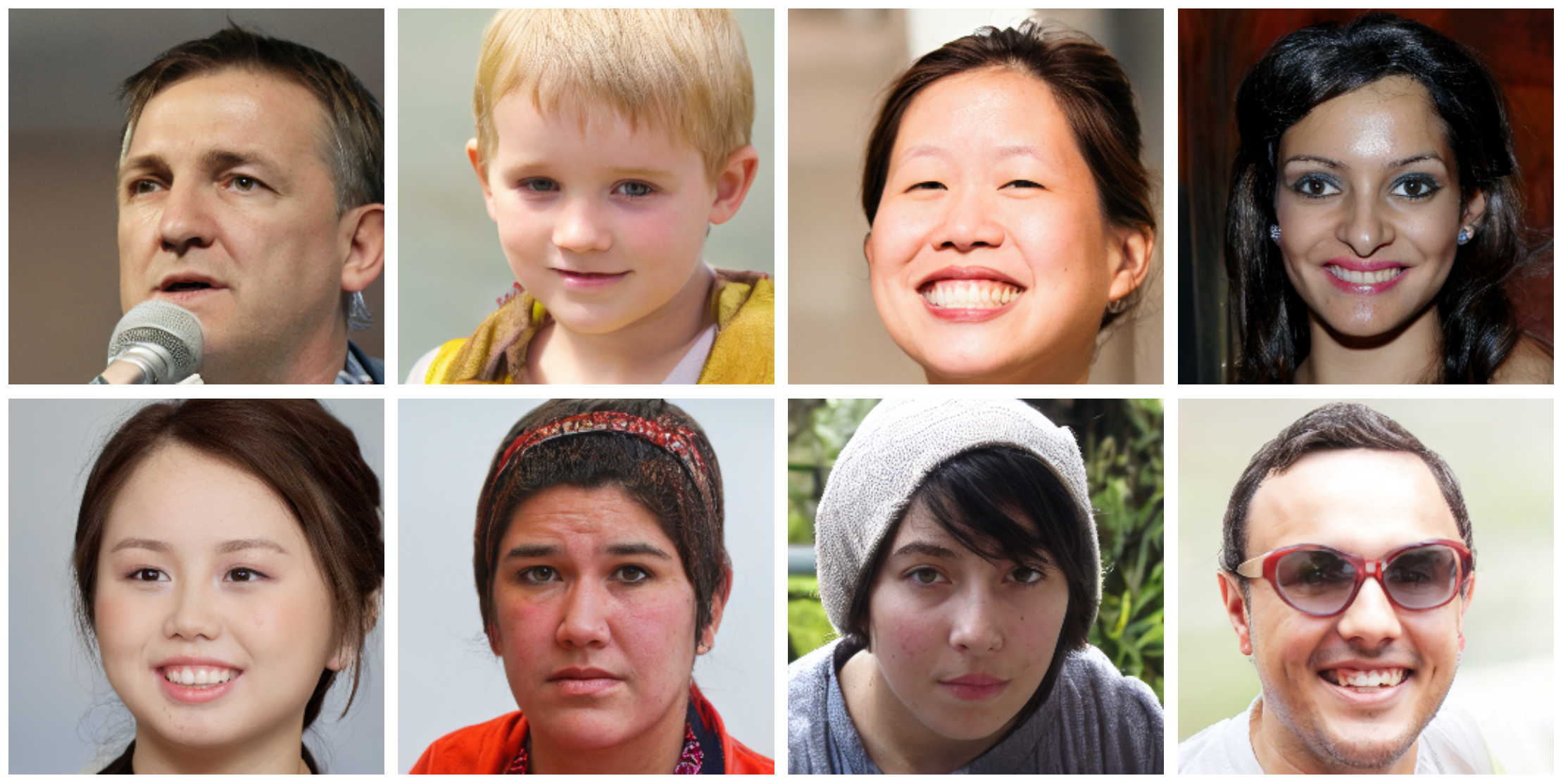} 
  \end{tabular} 
\caption[]{Images generated by the proposed closed-form discriminator guidance (DG$^*$) approach for the latent difusion model (LDM) on the 256-dimensional CelebA-HQ and FFHQ datasets.}
\label{Fig_LDMs}  
\end{center}
\vskip-1em
\end{figure*}

\textbf{Score-based Diffusion Models}:  Score matching was originally proposed by~\citet{ScoreMatching05} in the context of independent component analysis. Let the underlying distribution of the data to be modeled be denoted by \(\pd(\x)\). The {\it Stein score}~\citep{SteinScore16} is the gradient of logarithm of the density function with respect to the data, i.e., \(\DataScore{\x}\). It generates a vector field that points in the direction where the data density grows most steeply. In score matching, the score can be approximated by a parametric function \(S^{\mcalD}_{\phi}(\x)\) obtained by minimizing the Fisher divergence between the true score and the score estimated by the network~\citep{InfoTheory06}.
The output of the trained network is used to generate samples through annealed Langevin dynamics in noise-conditioned score networks (NCSN)~\citep{NCSN19}. Recent approaches  accelerate sampling by improving either the approximation quality of the score network~\citep{SSM20,DDPM20,NCSNv220,NCSNPP21,ScoreFlow21}, or the discretization of the underlying differential equations~\citep{GGF21,EDM22}. Upon discretization of the SDE, the evolution of the images, indexed by time \(t\), is denoted as \(\x_t\in\mbbR^n\), with \(\x_0 \sim \pd;\), and \(\x_T \sim \mathcal{N}(\bm{0},\mbbI)\), which is the standard Gaussian distribution. Image generation follows the reverse process, and is equivalent to sequentially denoising the sample \(\x_T\), to ultimately generate a realistic image that ideally comes from the distribution $p_d$.

\textbf{Generative Adversarial Networks (GANs)}:  GANs are a two-player game between a generator network \(G\colon\mbbR^d\rightarrow\mbbR^n\) and a discriminator network \(D\colon\mbbR^n\rightarrow\mbbR,~n\gg d\). Similar to the reverse process in diffusion, the generator transforms a noise vector \(\z \sim \pz;\,\z\in\mbbR^d\), typically standard Gaussian, into a {\it fake} sample \(G(\z)\), with the push-forward distribution \(\pg = G_{\#}(\pz)\). The discriminator accepts an input drawn either from the target distribution, \(\x \sim \pd;\,\x\in\mbbR^n\), or from the output of a generator, and learns a {\it real versus fake} classifier. The objective is to learn the {\itshape optimal generator} that can create realistic samples, which is equivalent to modeling the reverse process in a single step. GAN literature considers two main classes of loss functions: (a) \(f\)-divergence-based losses, and (b) integral probability metric (IPM) based losses. The standard GAN (SGAN,~\citet{SGAN14}), least-squares GAN (LSGAN,~\citet{LSGAN17}) and \(f\)-GANs~\citep{fGAN16} formulations, fall into the first category, wherein the discriminator models a chosen {\it divergence} metric between the target and generator distributions, while the generator network is trained to minimize this divergence. In IPM-GANs, the discriminator performs the role of a {\it critic}, and approximates the IPM, which in turn relates to a constraint class. For example, in Wasserstein GAN (WGAN),~\citep{WGAN17} consider Lipschitz-1 critics, while variants such as the Sobolev GAN~\citep{SobolevGAN18}, BWGAN~\citep{BWGAN18}, and PolyGAN~\citep{PolyGAN23} consider discriminator functions drawn from Sobolev spaces, with a corresponding penalty on the energy in the gradient.~\citet{KernelTest12} showed that the minimization of IPM losses can be equivalently solved through the minimization of kernel-based statistics in a reproducing-kernel Hilbert space (RHKS). Maximum-mean discrepancy GANs (MMD-GANs)~\citep{MMDGAN17,DemistifyMMD18} and Coulomb GAN~\citep{CoulombGAN18} are examples of kernel-based GANs.

\noindent {\it {\bfseries Discriminator Guidance (DG) in Diffusion Models}}: \citet{DMsbeatGANs21} and~\citet{CFG} use classifier gradients in conjunction with the score estimate of a diffusion model to improve the diversity of conditional image generation. \citet{DiscGuidance23} were the first to leverage the GAN discriminators, and showed that the score learnt at the time instant \(t\) in NCSN~\citep{NCSN19} could be improved by a correction term involving the SGAN discriminator gradients. Subsequently, \citet{pmlr-v235-naderiparizi24a,MinGuidance24,UnivGuidance} and~\citet{pmlr-v235-yang24h} have also explored discriminator guidance for superior coverage of the image manifold in diffusion, while~\citet{pmlr-v238-ekstrom-kelvinius24a} and \citet{kerby2024trainingfreeguidancediscretediffusion} combine DG with discrete diffusion models for molecular graph generation. However, these approaches typically either consider only the SGAN discriminator, or are unable to provide an explanation for the effectiveness of discriminator guidance when going beyond the SGAN setting.


\noindent {\it {\bfseries Unifying GANs and Diffusion Models}}: There has been a significant research focus on the optimality of the GAN discriminator function, with \citet{SobolevGAN18,DeconstructGAN20,HowWellGANs21,NTKGAN22,MonoFlow23} and \citet{ANON_JMLR}~considering a functional approach to derive the differential equations that govern the optimal discriminator, given the generator. Along another vertical,~\citet{WhatWGAN18},~\citet{WGANsFail21} and~\citet{WGANnotOT22} showed that, in practical gradient-descent-based training, the optimal discriminator is not attained. In the recent past, there has been a strong push to develop a unifying theory to explain GAN optimization, potentially leveraging results from flow-based approaches. For example,~\citet{MonoFlow23,DeepWFlow23} propose a unifying theory for all \(f\)-GANs under the umbrella of Wasserstein flows, while~\citep{fGANsScores23} link the generator optimization in SGANs to score-based sampling, and~\citet{UnifyingParticles23,DiffFlow23} formulate both GANs and score-based diffusion models as special cases of particle flows. While in most scenarios, the generator can be linked to minimizing the chosen divergence or IPM, the actual functional optimization has not been thoroughly explored. Motivated by the strong links between the guidance in diffusion and the GANs discriminator~\citep{DiscGuidance23}, and the equivalences between GAN training and Langevin sampling~\citep{UnifyingParticles23}, in this paper, we seek to answer the question: {\bf How does the closed-form optimization of the GAN generator link to discriminator guidance for diffusion?}

\subsection{Our Contributions} \label{Sec:Contrib} 
In this paper, we analyze the links between GAN optimization and score-based diffusion, and provide a principled approach to applying IPM-GAN discriminator guidance for diffusion models. The contributions of this paper are along two axes -- GANs and diffusion models. \par

First, considering the GAN optimization setting, we draw parallels between the generator optimization in IPM-GANs and score-based diffusion. Using {\it Variational Calculus}, we show that the generator optimality condition in IPM-GANs closely resembles the score-matching condition seen in diffusion models. We extend the analysis of~\cite{fGANsScores23} to the optimization of the generator loss in IPM-GANs, given the optimal discriminator. We show that the optimal generator in these settings minimizes a {\it smoothed score-matching} term, where the scores are conditioned by means of the kernel associated with the reproducing kernel Hilbert space (RKHS) from which the IPM discriminator is drawn, akin to noise-conditioned score networks (NCSN)~\citep{NCSN19}. That is, given an IPM-GAN, there exists a kernel associated with it's RKHS, and therefore, a corresponding kernel-smoothed score-matching formulation. Further, we show that, in IPM-GANs, the {\it smoothed score-matching} formulation is equivalent to minimizing a flow induced by the gradient field of a kernel function (\textbf{cf. Section~\ref{Sec:WGANs}}). These results can be viewed as a generalizations of Sobolev descent~\citep{SobolevDescent19}, MMD-Flows~\citep{MMDFlow19} and MonoFlows~\citep{MonoFlow23}. Leveraging these insights, we employ the closed-form IPM-GAN discriminator guidance in score-based diffusion. \par

Along the axis of Diffusion model, we demonstrate a closed-form discriminator guidance framework leveraging the kernel-based IPM-GAN discriminator (abbreviated DG$^*$) for existing Langevin sampling frameworks. We consider (a) Noise-free discriminator-only ODE flow; (\textbf{cf. Section~\ref{Sec:DiscInfusion}}) (b) Discriminator-only Langevin flow (\textbf{cf. Section~\ref{Sec:DiscInfusion}}), wherein we replace the score with DG$^*$ and (c) Closed-form discriminator guidance for score-based Langevin diffusion (both in the image or the latent space, \textbf{cf. Section~\ref{Sec:LDM}}). Theoretically, we show that the proposed approaches results in improved convergence over the classical score-based diffusion (\textbf{cf. Section~\ref{Sec:DiscGuidance}}), and that applying DG$^*$ can be viewed as introducing a second-order term to the update equation, thereby accelerating convergence in the Polyak heavy-ball momentum sense~\citep{StatNotes18} (\textbf{cf. Section~\ref{Sec:DiscGuidance}}). Lastly, we show that DG$^*$ can be coupled with existing approaches for accelerated diffusion, considering two example frameworks: (a) The time-step-shifted diffusion~\citep{TimeShift24}, and (b) The accelerated DPM Solver~\citep{DPMSolver22} (\textbf{cf. Section~\ref{Sec:LDM}}). We show that the inclusion of DG$^*$ can further accelerate the denoising process, allowing for larger jumps in noise levels when in time-step-shifted diffusion, and superior FID scores, given comparable sampling steps, when using the DPM solver. \par 
To summarize, our \textbf{key contributions} are two-fold: We develop a strong theoretical foundation for employing closed-form IPM-GAN discriminators for guidance, by establishing equivalences between GAN-generator optimality and smoothed score-matching. We leverage these insights to develop a novel closed-form discriminator guidance framework that can be applied in a \textit{plug-and-play} fashion with an existing diffusion model, demonstrated through experimentation on multiple baseline such as NCSN~\citep{NCSN19}, and LDMs~\citep{LDM22}, DPM-Solver~\citep{DPMSolver22}, etc.

\section{Background on Diffusion and GANs} \label{Sec:MathPrelims}
In this section, we introduce diffusion probabilistic models and GANs.
\textbf{Diffusion Probabilistic Models (DPMs)} primarily model the {\it forward process} wherein Gaussian noise is progressively added to an image \(\x\sim\pd\). The noise is modelled as adhering to a fixed variance schedule $\beta(t)$. The generative task is one of modeling the reverse process, essentially iterated denoising. Given the data distribution $\pd$ and a fixed noise schedule $\beta(t) \in (0,1), \forall t = 1 \ldots T$, the forward process, structured as a Markov process, is expressed as $p(\x_{1,2,\ldots, T}|\x_{0}) = \prod^{T}_{t=1} p(\x_{t}|\x_{t-1}).$ In the DPM setting, the forward transition kernel at time \(t\), given by $p(\x_{t}|\x_{t-1})$ can be defined as a Gaussian \(\mathcal{N}(\sqrt{\alpha_{t}}\x_{t-1}, \beta_{t}\mbbI)\), centered around the sample \(\sqrt{\alpha_{t}}\x_{t-1}\), where $\alpha_{t} = 1 - \beta_{t}$~\citep{DDPM20}. Via the re-parameterization trick, the conditional distribution is given by \(p(\x_{t-1}|\x_{t}, \x_{0}) = \mathcal{N}(\Tilde{\mu}_{t}, \Tilde{\beta_{t}})\), wherein, $\Bar{\alpha}_{t} = \prod^{t}_{i=1} \alpha_{i}$ and $\epsilon_{t} \sim \mathcal{N}(\bm{0}, \mbbI)$, $\Tilde{\mu}_{t} = \frac{1}{\sqrt{\alpha_{t}}}\left(\x_{t} - \frac{1 - \alpha_{t}}{\sqrt{1 - \Bar{\alpha}_{t}}}\epsilon_{t}\right)$,  $\Tilde{\beta_{t}} = \dfrac{(1 - \Bar{\alpha}_{t-1})}{1 - \Bar{\alpha}_{t}}\beta_{t}$ and \(p(\x_0) = \pd\). Training DPMs involves learning a neural network $\epsilon_{\theta}$ to approximate $\epsilon_{t}$, with the following mean-squared-error loss~\citet{DDIM21}:
\begin{align}
    \loss_{\mathrm{DPM}} &= \mathbb{E}_{t, \x_t, \epsilon_{t} \sim \mathcal{N}(0, \mathbb{I})} [\left\| \epsilon_{\theta}(\x_{t}, t) - \epsilon_{t}\right\|_{2}^{2}]
\label{eqn: ddpm_loss}
\end{align}
In practice, the model is trained on a variational lower bound of the negative log-likelihood loss. Consequently, generation starts by sampling \(\x_T\) from a standard Gaussian and progressively generating samples according to the recursion:
\begin{align*}
    &\x_{t-1} = \mu_{\theta}(\x_t,t) + \Sigma_{\theta}(\x_t,t)\z_t,~t = T,T-1,\ldots,0,
\end{align*}
where \(\z_t\sim\mcalN(\bm{0},\mbbI)\), and \(\mu_{\theta}\) and \(\Sigma_{\theta}\) are the estimates of the noise mean and covariance, as output by \(\epsilon_{\theta}\). \revision{The SDE governing the above process was generalized by~\citet{DDIM21}, and is given by:
\begin{align}
    \rmd \bmX_t &= \left( f(t) + g^2(t) \nabla_{\bmX} \ln p_t^*(\bmX_t) \right)\rmd t + g(t) \rmd\bmW_t, \label{Eqn_DiffusionBase}
\end{align} 
for suitable function \(f\) and \(g\), where \(\rmd\bmW\) refers to the standard Wiener process. We refer the reader to~\citep{DDIM21} for an in-depth analysis for the choice of these functions.} The discretized update is then given by:
\begin{align}
        \x_{t-1} = \underbrace{\sqrt{\frac{\alpha_{t-1}}{\alpha_t}}\x_t - \sqrt{\frac{\alpha_{t-1}}{\alpha_t}}\sqrt{(1-\alpha_t)}\epsilon_{\theta}(\x_t,t)}_{\hat{\x}_0} + \sqrt{(1-\alpha_{t-1}) - \sigma^2_{t}}\cdot\epsilon_{\theta}(\x_t,t) +  \sigma_{t}\epsilon_{t}
\label{eqn:ddim_update}
\end{align}
where $\hat{\x}_0$ can be viewed as the \textit{prediction} of  $\x_0$, $\epsilon^{t}_{\theta}(\x_t)$ represents the direction pointing towards $\x_{t}$ with $\alpha_{0}=1$, and $\sigma_{t}\epsilon_{t}$ is the diffusion term with $\epsilon_t \sim \mathcal{N}(0, \mathbb{I})$ being standard Gaussian. Different values of $\sigma$ lead to different generative processes while keeping $\epsilon_{\theta}$ fixed. In general, we can set \(\sigma_{\tau(\eta)} = \eta \sqrt{(1-\alpha_{t-1})/(1-\alpha_{t})} \sqrt{(1-\alpha_{t}/\alpha_{t-1})}\), where setting \(\eta = 1\) results in the DDPM framework~\citet{DDPM20}, and for $\eta = 0$, the samples generated obey a deterministic procedure, giving rise to the denoising diffusion implicit model (DDIM) sampling~\citep{DDIM21}. In this work, we explore the inclusion of closed-form discriminator guidance in the DDIM setting.

\textbf{Optimality of GANs}: GAN optimization can be viewed as minimizing either the \(f\)-divergence~\citet{fGAN16} between the target distribution \(\pd\) and the distribution of the generated samples (denoted as \(\pg\)), or an integral probability metric (IPM) between \(\pd\) and \(\pg\)~\citep{WGAN17}. For completeness, we recall the optimality result for $f$-GANs derived by~\citet{fGANsScores23}, wherein the authors showed that the optimal $f$-GAN generators performed score-matching. Detailed discussions on this result are provided in Appendix~\ref{App_fGANs}. 
\begin{theorem} \label{Theorem_fGAN_OptG}~\citep{fGANsScores23}
(\textbf{Informal}) Consider the optimization in \(f\)-GANs. The {\bfseries optimal \(f\)-GAN generator} satisfies the following score-matching condition:
\(
\nabla_{\x} \ln \left(p_{t-1}(\x) \right)\big|_{\x = G^*_t(\z)} =  \DataScore{\x} \big|_{\x = G^*_t(\z)},
\)
where \(G^*_t\) is the optimal generator at time \(t\), \(\DataScore{\x}\) is the score of the data distribution at \(\x\), and \(p_{t-1}\) is the push-forward distribution at \(t-1\).
 \end{theorem}
 In the IPM-GAN setting,~\citet{WGAN17} proposed Wasserstein GANs (WGANs) as an alternative to divergence-minimizing GANs. Motivated by {\it optimal transport}, the discriminator (also called the {\it critic}) approximates the Wasserstein-1 distance between \(\pd\) and \(\pg\). The optimization is then defined through the Kantorovich--Rubinstein duality as:
\begin{align}
    \min_{\pg} \max_{D} \left\{ \Esub_{\x \sim \pd}[D(\x)] -  \Esub_{\x \sim \pg}[D(\x)] + \Omega_D \right\}, \label{eqn:WGAN_optimization}
\end{align}
where \(\Omega_D\) is an appropriately chosen regularizer. We let \(D^*(\x)\) denote the optimal discriminator. During training,~\citet{WGAN17} ensure a Lipschitz discriminator by clipping the network weights. Subsequent variants considered regularizers that bound the energy in the discriminator gradient~\citep{WGANLP18,SobolevGAN18,BWGAN18,PolyGAN23}, resulting in Sobolev constraint spaces. In practice, this optimization is an alternating one, wherein \(D_t\), the discriminator at time \(t\), is derived given the generator of the previous iteration \(G_{t-1}\), and the subsequent generator optimization involves computing \(G_{t}\), given \(D^*_{t}\) and \(G_{t-1}\). The optimal discriminator in these variants has been shown to be the solution to partial differential equations (PDEs)~\citep{SobolevGAN18,PolyGAN23}, which can be represented via kernel-based convolutions: 
\begin{align}
D_t^*(\x) &= \mathfrak{C}_{\kappa} \left( \left( p_{t-1} - \pd \right) * \kappa \right) (\x),\label{eqn_OptD}
\end{align}
where the kernel \(\kappa\) is the Green's function to the differential operator and \(\mathfrak{C}_{\kappa}\) is a positive constant. For example, in Poly-WGAN~\citep{PolyGAN23}, the kernel corresponds to the family of polyharmonic splines (PHS), given by
\begin{align*}
\kappa(\x) = \begin{cases}
\|\x\|^{\mathit{k}} &\text{if}~~\mathit{k}<0~~\text{or}~~n~\text{is odd}, \\
\|\x\|^{\mathit{k}} \ln(\|\x\|) & \text{if}~~\mathit{k}\geq0~~\text{and}~~n~\text{is even},%
\end{cases}
\end{align*}
where in turn, \(\mathit{k} = 2m - n\), \(m\) being a hyperparameter that controls to smoothness of the discriminator and \(n\) is the dimensionality of the data, and the authors showed that setting \(m=\lceil \frac{n}{2}\rceil\) results in optimal performance in GANs. We now extend the results derived for \(f\)-GANs~\citep{fGANsScores23} to the IPM-GAN setting.

\section{The Optimal Generator in IPM GANs} \label{Sec:WGANs}
To motivate our results, consider the solution to Theorem~\ref{Theorem_fGAN_OptG}. We observe that the optimal \(f\)-GAN generator is the one that matches the score of the generator push-forward distribution  to the score of the data distribution. While this results in the classical discriminator guidance framework~\citep{DiscGuidance23}, \(f\)-GANs are known to be unstable to train~\citep{PrincipledMethods17,DiscGuidance23}. Furthermore, as noted by~\citep{MonoFlow23}, \(f\)-GANs can be viewed as a special case of IPM-GANs. Therefore, we derive the general solution to generator optimality that holds for all IPM-GANs. Consider the IPM-GAN optimization problem given in Eqn.~\eqref{eqn:WGAN_optimization}. Then, the following theorem holds:
\begin{theorem} \label{Theorem_IPMGAN_OptG}
Consider the generator loss given by \(\loss^{\kappa}_G(G;D_t^*,G_{t-1}) = - \Esub_{\z\sim\pz} [ D_t^*\left( G(\z)\right)]\), and the optimal discriminator given in Equation~\ref{eqn_OptD}. The {\bfseries optimal IPM-GAN generator} satisfies
 \begin{align}
\mathfrak{C}_{\kappa} & \left( \Esub_{\y \sim p_{t-1}}\left[\nabla_{\y} \ln{p_{t-1}(\y)} \kappa (\x-\y) \right] - \Esub_{\y \sim \pd}\left[\nabla_{\y} \ln{\pd(\y)} \kappa (\x-\y) \right] \right) \bigg|_{\x = G_t^*(\z)}=\bm0,
 \label{Eqn_IPMGAN_GtStar_Cond}
 \end{align}
for all \(\x = G_{t}^*(\z)\), \(\z\sim\pz\), where \(\mathfrak{C}_{\kappa}\) is a non-zero constant dependent on the kernel \(\kappa\).
 \end{theorem}
 The above theorem shows that the optimal generator in IPM GANs is also one of score-matching, where the score is conditioned by the kernel function, centered around \(\x\). As the following lemma shows, Theorem~\ref{Theorem_IPMGAN_OptG} can equivalently be reformulated using the kernel gradient as follows:
\begin{lemma} \label{Lemma_IPMGAN_OptG}
Consider the optimality condition for the IPM generator, presented in Theorem~\ref{Theorem_IPMGAN_OptG}. The condition can be written equivalently as:
\(
\mathfrak{C}_{\kappa}  \left( \left(\pd - p_{t-1} \right) * \nabla_{\x} \kappa \right)(\x) \big|_{\x = G_t^*(\z)}=\bm0,\)
where \(\nabla_{\x} \kappa\) denotes the gradient vector of the kernel, and the convolution must be interpreted element-wise, {\it i.e.,} \(\pd(\x) - p_{t-1}(\x)\) is convolved with each entry of~\(\nabla_{\x} \kappa\).
 \end{lemma}
The proof of Theorem~\ref{Theorem_IPMGAN_OptG} and Lemma~\ref{Lemma_IPMGAN_OptG} are presented in detail in Appendix~\ref{App_KernelGAN}. The optimal IPM-GAN generator can be seen as minimizing a proxy to the score --- similar to the Stein score --- where the gradient field induced by the kernel \(\kappa\) is maximized at locations where data samples are present. As observed in Coulomb GANs, these are akin to charge-potential fields, with {\it attractive} data samples and {\it repulsive} generator samples. While we use the polyharmonic spline (PHS) kernel \(\kappa\) due to its stability~\citep{PolyGAN23}, other choices are discussed in Appendix~\ref{App_IPMGANs}. \par


\subsection{Linking the Optimal IPM-GAN Generator to Score-based Diffusion}
Based on the theoretical insights, we see that, given the optimal discriminator $D_t^*$ that admits a kernel-based interpolation form at training iteration $t-1$, the optimal generator at the subsequent iteration $G_t^*$ can be derived as the one that minimizes the value of the convolution between the density difference, and the gradient of the optimal discriminator kernel, \textit{i.e.,} minimize $((p_d-  p_t) *\nabla\kappa)$. For most popular positive-definite kernels $\kappa$, this term would be minimized when the generator distribution $p_t$ moves towards the data distribution $p_d$. Furthermore, from Lemma~\ref{Lemma_IPMGAN_OptG}, we see that the gradient field of the kernels convolved with the density difference, and the data score \(\DataScore{\x}\), serve similar purposes: output an arbitrarily large value at data sample location, and low values elsewhere. Unlike the score, however, the kernel gradients produce a repulsive force at the location of generator samples, resulting in a {\it push-pull} framework -- The target distribution creates a {\it pull}, while the generator distribution creates the {\it push}. \par

These results serve to validate why IPM GANs typically do not suffer from vanishing gradients~\citep{PrincipledMethods17}, as opposed to the \(f\)-divergence counterparts. When \(p_0(\x)\) is initialized far from the target, although the {\it influence} of the score is weak, the repulsive force of the kernel-based loss is strong. The derived solution can also be used to explain denoising diffusion GANs (DDGAN,~\citet{DDGAN22}), wherein a GAN is trained to model the reverse diffusion process, with the generator and discriminator networks conditioned on the time index. DDGAN can be seen as a special instance of our approach, with Langevin updates over the gradient field of the time-conditioned discriminator (cf. Appendix~\ref{App_IPMGANs}). The kernel-convolved score-matching condition can also be viewed as generalized score matching~\citep{GeneralizedScore09} where the IPM-GAN generators minimize a {\it generalized score}, {\it i.e.,} given an IPM GAN, an equivalent diffusion model exists, with the flow field induced by the kernel of the discriminator, and vice versa. We demonstrate this approach in Section~\ref{Sec:DiscInfusion}. \par

\section{Closed-form IPM-GAN Discriminator Guided Langevin Diffusion} \label{Sec:DiscGuidance}
\revision{The results derived above allows us to explore Langevin sampling, wherein the score of the data is either replaced, or guided using the gradient of the kernel-based discriminator. In particular, we can explore three approaches to closed-form discriminator guidance: (a) Noise-free discriminator-only ODE flow; (b) Discriminator-only Langevin flow, and (c) Closed-form discriminator guidance for score-based Langevin diffusion (either in the image or the latent space). Additionally, given the \textit{push-pull} nature of the discriminator, we intuit, and subsequently show, that the applied discriminator guidance leads to an accelerated sampling strategy that is orthogonal to existing acceleration techniques to improve the discretization of the Langevin SDE.} While the score of the data possesses a {\it strong attractive force} in regions close to the target data, it does not significantly influence samples that are far away. On the other hand, the kernel gradients possess a repulsive term that {\it pushes} particles away from where they previously were, thereby accelerating convergence. \par

First, in the discriminator-only flow setting, we consider the following update scheme:
\begin{align*}
  \x_{t+1} = \x_{t} - \alpha_t \nabla_{\x}D_t^*(\x_t) + \gamma_{t}\z_t,
\end{align*}
where \(\z_t\sim\mcalN(\bm{0}_n,\mbbI_n)\) and \( \nabla_{\x}D_t^*(\x_t)\) denotes an \(N\)-sample estimate of the discriminator gradient with centers \(\bmd^{i}\sim\pd\), and the set of samples generated at the previous iteration \(\{\x_{t-1}\,|\,\x_{t-1}\sim p_{t-1}\}\):
\begin{align}
     \nabla_{\x}D_t^*(\x_t) = \mathfrak{C}^{\prime}_k\!\!\sum_{\bmg^j \sim \{\x_{t-1}\}} \!\!\!\nabla_{\x}\kappa(\x_t-\bmg^j) - \mathfrak{C}^{\prime}_k\!\!\sum_{\bmd^i \sim \pd} \!\!\nabla_{\x}\kappa(\x_t-\bmd^i).
  \label{Eqn_TimedGradD}
\end{align}
Typically, \(\gamma_t = \sqrt{2\alpha_t}\), while \(\alpha_t\) is decayed geometrically~\citep{NCSN19}, while setting \(\gamma_t=0\) results in the ODE-flow scenario. \revision{The reverse process associated with the discriminator-guidance framework can be written as:
\begin{align}
\rmd \bmX_t = \left( f(t) + g^2(t) \right){\epsilon}_{\theta}(\bmX_t)~\rmd t  + h(t) \nabla_{\bmX} D_t^*(\bmX_t)~\rmd t + g(t)\rmd\bmW_t,   \label{Eqn_DiffusionOurs}
\end{align}
where \(h(t)\) models the weight associated with the discriminator guidance term. In practice, we denote $h(t) = w_{dg, t}$ for simplicity. The following Lemma bounds the error in the DG$^*$ setting:}
\begin{lemma}
    Consider the reverse diffusion processes associated with the base score-based approach, and the proposed closed-form discriminator (DG$^*$) guidance model. Let the probability densities associated with these two processes be \(p^*_t\) and \(p_t\), with \(p_T^* = \mcalN(\bm{0},\mbbI)\), \(p_T = \pi\), \(p_0^*=p_d\) and \(p_0 = p_m\), denoting the data, and the \textit{modeled} target and data distributions, respectively. Then,
    \begin{align*}
        \mcalD_{KL,\mathrm{DG}^*}(\pd\|p_m) \leq \mcalD_{KL}(p_T^*\|\pi) + \varepsilon_{D^*},
    \end{align*}
    where \(p_m\) is the \textit{modeled} data distribution and the error is:
    \begin{align}
         \varepsilon_{D^*} = \frac{1}{2} \mbbE_{p^*_t}\!\left[\int\!\!g^2(t)\Big\|E_{S^*}\!-\!h(t) \nabla_{\bmX} D_t^*(\bmX_t)\Big\|^2\!\rmd t\right]\!, \label{Eqn_Gain1}
    \end{align}
    where in turn, \(E_{S^*} = \nabla \ln p_t^*(\bmX_t) - {\epsilon}_{\theta}(\bmX_t)\) is the error in the standard score-based Langevin sampler, and \(D_t^*\) denotes the closed-form kernel-based discriminator at time $t$, with either the Gaussian kernel or the PHS kernel with \(k\leq0\).
\end{lemma}
The proof of the above Lemma is provided in Appendix~\ref{App_Convergence}, where we show that, when \(\bmX_t \sim p_t\) far from \(p_d\), the discriminator gradients are positive, and we see a gain in the KL-divergence over the standard score-based sampler. The above result shows that the discriminator-guided Langevin diffusion process converges to the data distribution, with an error lower than that achieved by the standard score-based Langevin diffusion. In addition, the proposed solution can also be viewed as accelerating convergence, as discussed by the following Lemma:
\begin{lemma}
    Consider the Langevin SDE-based update:
    \begin{align*}
      \bmX_{t+1} = \alpha_{1,t} \bmX_t - \alpha_{2,t} \epsilon_{\theta}(\bmX_t) - \alpha_{3,t} \nabla D_t(\bmX_t) + \alpha_{4,t} \mathbf{Z}_t,
    \end{align*}
    where \(\alpha_{i,t},~i=1,2,3,4\) denote the coefficient of various terms involved. Let $\bmd$ be a random sample drawn from the target data distribution, used to define a 1-sample approximation of the polyharmonic-kernel discriminator gradient with $k=1$. Then, the above update is equivalent to: 
\begin{align*}
\bmX_{t+1}\!=\!\beta_{1,t} \bmX_t \!-\! \alpha_{2,t} \epsilon_{\theta}(\bmX_t) \!-\!\beta_{3,t} \bmX_{t-1} + \alpha_{4,t} \mathbf{Z}_t + \beta_{5,t}
\end{align*}
where $\beta_{1,t} = \alpha_{1,t} - \frac{\alpha_{3,t}\mathfrak{C}^{2}_k}{\Vert \bmX_{t} - \bmX_{t-1} \Vert} + \frac{\alpha_{3,t}\mathfrak{C}^{2}_k}{\Vert \bmX_{t} - \bmd \Vert}$, $\beta_{3,t} = \frac{\alpha_{3,t}\mathfrak{C}^{2}_k}{\Vert \bmX_{t} - \bmX_{t-1} \Vert}$ and \(\bm{\beta}_{5,t} = \left(\frac{\alpha_{3,t}\mathfrak{C}^{2}_k}{\Vert \bmX_{t} - \bmd \Vert}\right) \bmd\). 
\end{lemma}  
\revision{Detailed discussions are provided in Appendix~\ref{App_WANDAConv}. While an in-depth analysis of second-order acceleration in diffusion is outside of the score of this paper, the above result shows that the closed-form discriminator guidance terms can be viewed as a second-order update that resembles the Polyak heavy-ball momentum update found in the literature~\citep{StatNotes18,RechtBook22,FastDiff23} and can be attributed to being the source for the acceleration. This acceleration is orthogonal to existing methods that develop improved SDE discretization techniques to accelerate sampling~\citep{DPMSolver22,FastDiff23,TimeShift24,FastODE24} and can therefore be combined with these techniques to further improve the sampling efficiency. We demonstrate this considering the DPM solver~\citep{DPMSolver22}, and time-shifted sampling~\citep{TimeShift24} (cf. Section~\ref{Sec:LDM}).} \par

\subsection{Experimental Results} \label{Sec:DiscInfusion}

\begin{figure*}[t!]
  \begin{center}
    \begin{tabular}[b]{P{.01\linewidth}|P{.09\linewidth}P{.09\linewidth}P{.09\linewidth}P{.09\linewidth}l}
      \multirow{3}*{\rotatebox{90}{Circular \enskip}} &
      \multirow{3}*{\includegraphics[width=1.1\linewidth]{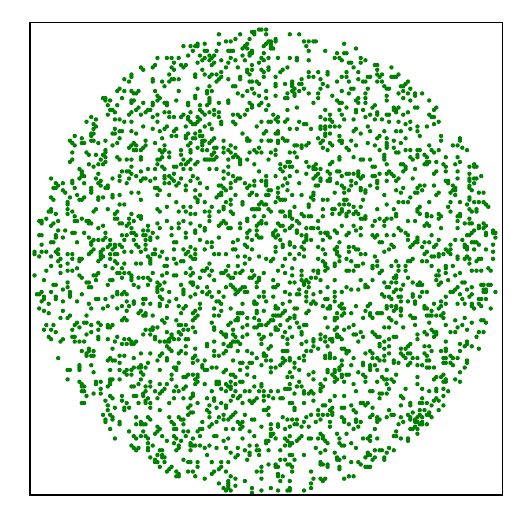}} & 
      \multirow{3}*{\includegraphics[width=1.1\linewidth]{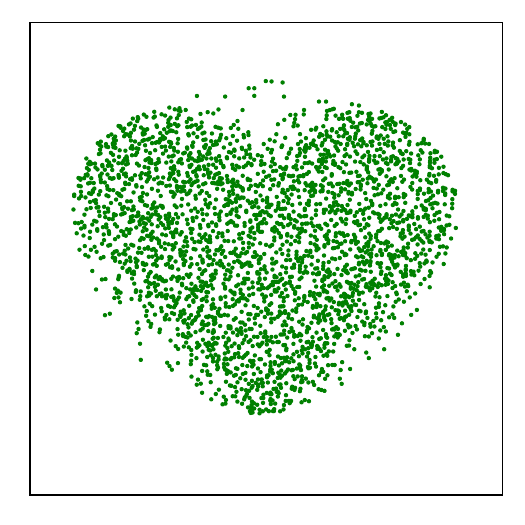}} & 
      \multirow{3}*{\includegraphics[width=1.1\linewidth]{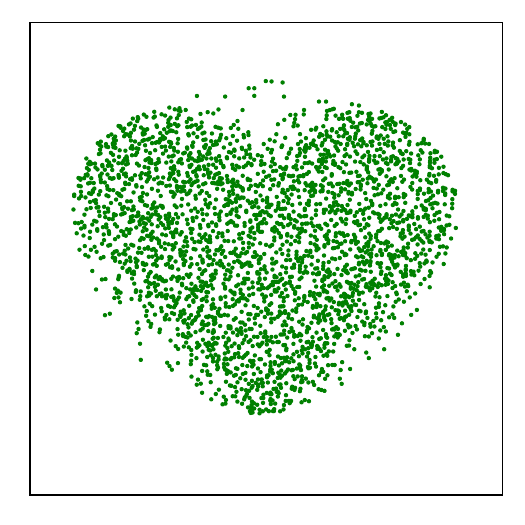}} & 
      \multirow{3}*{\includegraphics[width=1.1\linewidth]{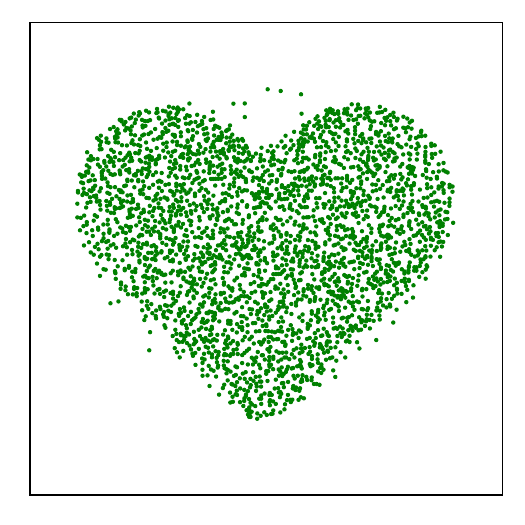}} & 
      \multirow{4}*{\includegraphics[width=0.47\linewidth]{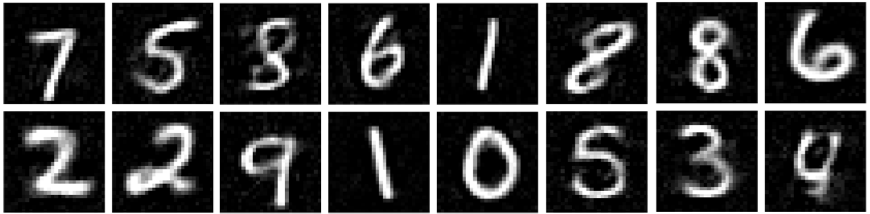}} \\ \\ \\ \\[1pt] 
      \multirow{3}*{\rotatebox{90}{ Spiral \enskip}} &
      \multirow{3}*{\includegraphics[width=1.1\linewidth]{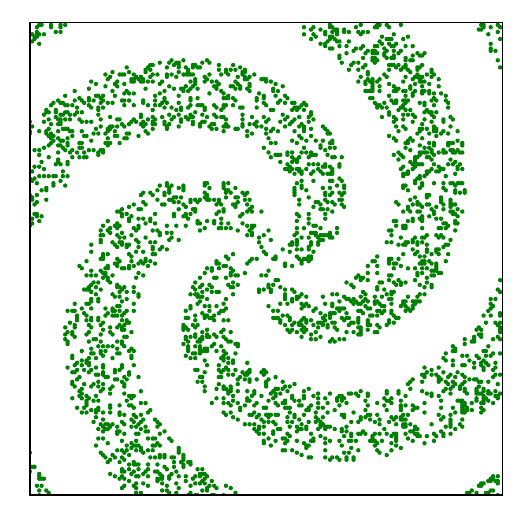}} & 
      \multirow{3}*{\includegraphics[width=1.1\linewidth]{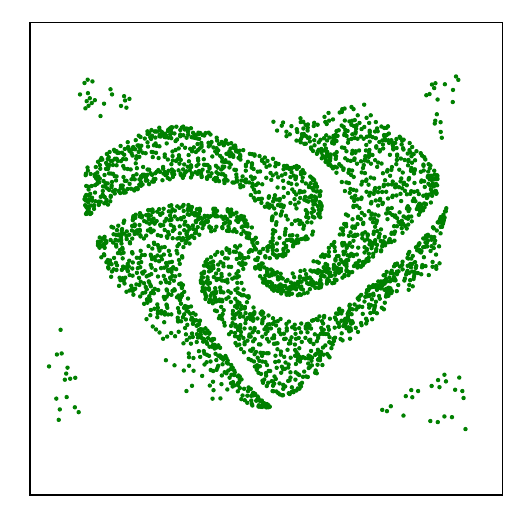}} & 
      \multirow{3}*{\includegraphics[width=1.1\linewidth]{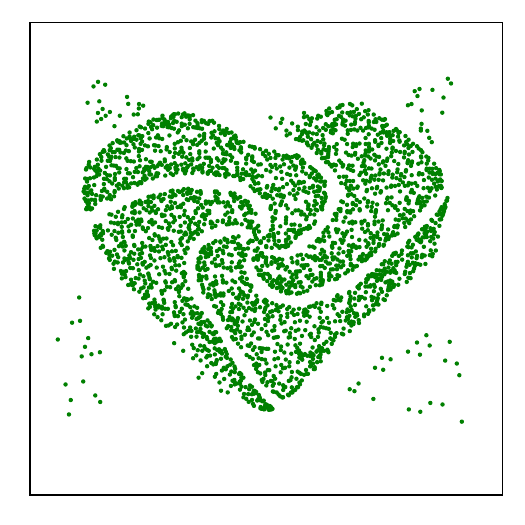}} & 
      \multirow{3}*{\includegraphics[width=1.1\linewidth]{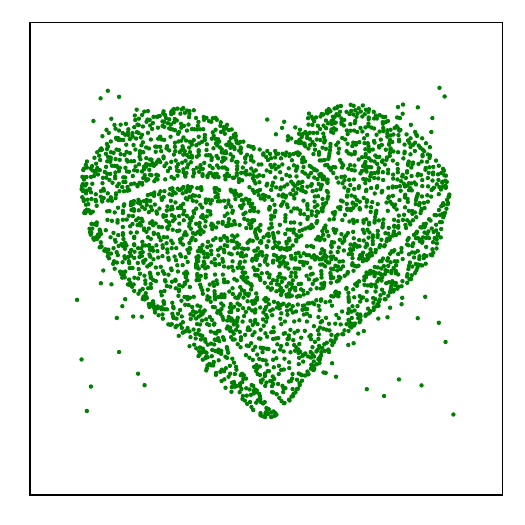}} & 
      \multirow{2}*{\includegraphics[width=0.47\linewidth]{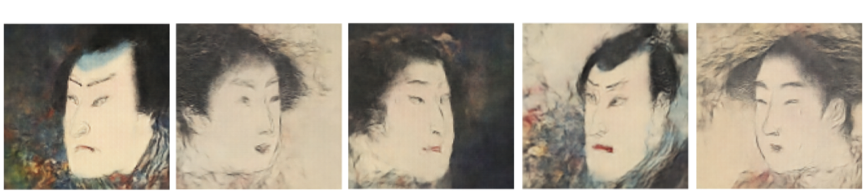}} \\ \\ \\ \\[-1pt] 
      & \scriptsize{\(t=1\) } &\scriptsize{\(t=50\) } & \scriptsize{\(t=100\) } &\scriptsize{\(t=250\) } & \multirow{2}*{\enskip\qquad\qquad\qquad\qquad(b)}\\[-1pt] 
      & \multicolumn{4}{c}{(a)}  \\[-5pt]
    \end{tabular} 
  \caption[]{(\includegraphics[height=0.009\textheight]{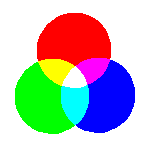} Color online) (a) Shape morphing using the proposed discriminator-guided Langevin sampler. For relatively simpler input shapes, such as the circular pattern, the sampler converges in about 100 iterations, while in the  spiral case, the sampler converges in about 500 steps. (b) Images generated using the discriminator-guided Langevin sampler on MNIST and Ukiyo-E faces datasets.} 
    \label{Plot_Morphing}
    \end{center}
  \vskip-0.5em
  \end{figure*}

To demonstrate the performance of the discriminator-guided Langevin flow, we consider shape morphing, proposed by~\citet{SobolevDescent19}. The source and target samples are drawn uniformly from the interior regions of pre-defined shapes. Figure~\ref{Plot_Morphing}(a) depicts two such scenarios, where the target shape is a heart, and the input shapes are a disk, and a spiral, respectively. Additional combinations are presented in Appendix~\ref{App_DiscInfusion}. The discriminator-guided Langevin sampler converges in about 500 iterations in all the scenarios considered, compared to the 800 iterations reported in Sobolev descent~\citep{SobolevDescent19,USobolevDescent20}. We extend the proposed approach to images, considering MNIST, SVHN and Ukiyo-E~\citep{UkiyoE} datasets. Ablation experiments on the choice of \(\alpha_t\) and \(\gamma_t\), and extensions to the EDM sampler~\citep{EDM22} are provided in Appendix~\ref{App_DiscInfusion}. Figure~\ref{Plot_Morphing}(b) presents the samples generated by this discriminator-guided Langevin sampler on MNIST and 256-dimensional Ukiyo-E faces. The model converges to realistic images in as few as 300 steps of sampling, resulting in performance comparable to baseline NCSN~\citep{NCSN19}. Subsequent iterations, as in NCSN, serve to {\it clean} the noisy images generated. 
 Additional experiments are provided in Appendix~\ref{App_DiscInfusion}. \par

These motivating experiments provide two key observations. First, since diffusion models such as NCSN work directly on the pixel space, the evaluation of the closed-form discriminator computationally expensive. Scaling the discriminator-guided Langevin sampler is therefore infeasible on high-resolution datasets such as CelebA-HQ~\citep{PGGAN18} and FFHQ~\citep{StyleGAN19}. Second, we observe that the inclusion of the discriminator guidance over all iterations may not be necessary, and we could fall back to score-based sampling once the discriminator guidance brings us close to the image distribution. We now present approaches to leverage these insights to apply the closed-form IPM-GAN discriminator guidance for accelerating diffusion models in the latent space.

\begin{table*}[!t]
  \fontsize{7.5}{12}\selectfont
  \begin{center}
  \caption{A comparison of the proposed LDM+DG$^*$ and WANDA samplers and the baselines on CelebAHQ and FFHQ datasets. LDM+DG$^*$  outperforms the baseline on the Clean-FID, CLIP-FID and KID metrics. \(^*\)While the FID reported by~\citep{LDM22} is 5.11, we were unable to reproduce these numbers (even with pre-trained models) using standard metric libraries (Clean-FID~\citep{CleanFID21} and Torch Fidelity~\citep{TorchFid20}). A \(\dagger\) denotes a metric computed via Torch Fidelity, and \(\ddagger\) denotes a metric computed via Clean-FID. } 
   \label{Table_LDM} 
   \begin{tabular}{P{.005\linewidth}P{2.2cm}||P{0.93cm}|P{1.5cm}|P{1.5cm}|P{1.4cm}|P{1.4cm}|P{1.15cm}}
   \toprule \toprule 
  &Method & *FID\(\dagger\) \(\downarrow\) & Clean-FID\(\ddagger\) \(\downarrow\)& CLIP-FID\(\ddagger\) \(\downarrow\)& KID\(\ddagger\) \(\downarrow\)& Precision\(\dagger\) \(\uparrow\) & Recall\(\dagger\) \(\uparrow\)\\[1pt]
  \midrule
  \multirow{3}{*}{\rotatebox{90}{~\underline{CelebAHQ}}} & LDM & \textbf{18.21} & 21.53	 	 & 7.17		 & \(0.0221\)	& 0.5434 & 0.4406	\\[-1pt]
  & LDM+DG$^*$ ({\bf Ours})  	  & 18.46 & {\bfseries20.49	}	 & {\bfseries6.48} 	 &  \({\bf 0.0204}\)	& 0.4932 & 0.4806  \\[-1pt]
  & WANDA ({\bf Ours}) & 19.84 & 22.76 & 7.98 & \(0.0227\) & 0.4570 & \textbf{0.4990}  \\
  \midrule
  \multirow{3}{*}{\rotatebox{90}{\underline{~~~FFHQ~~}}} & LDM & \textbf{10.97} & 8.65 & 7.16	 & \(0.0034\)	& 0.545 & 0.563 \\[-1pt]
  & LDM+DG$^*$ ({\bf Ours})  & 11.05 & \textbf{7.92} & \textbf{6.51} & \({\bf 0.0030}\)	& 0.537 & \textbf{0.571} \\[-1pt]
  & WANDA ({\bf Ours})  & 11.78 & 8.79 & 7.06 & \(0.0034\)	& 0.540 & 0.568  \\[-2pt]
  \bottomrule\bottomrule
   \end{tabular}
   \end{center}
   \end{table*}


\section{Extension to Latent Diffusion Models}\label{Sec:LDM}
Given the limitations of the pixel-space generation given above, we extend the closed-form discriminator-guidance approach to latent diffusion models (LDMs)~\citep{LSGM21,LDM22}, wherein the score, and the closed-form discriminator guidance (DG$^*$) term are defined over \(\bme_{\x} = \mcalE_{\mathrm{LDM}}(\x)\), the LDM-encoded representation of \(\x\). The resulting LDM baseline is therefore a DDIM sampler working on encoder representations. Experimentally, we found that setting the temporal weighting factor $w_{dg, T} = 5$ with an exponential decay resulted in superior image generation quality. Ablations on this choice are discussed in Section~\ref{App_Ablations} \par

Figure \ref{Fig_LDMs} presents the samples generated using vanilla LDM update and LDM+DG$^{*}$ approach sampled using the equation above, on  CelebA-HQ. Similar comparisons on the FFHQ dataset are provided in Appendix~\ref{App_DiscInfusion}. Both approaches are initialized with the deterministic sampler ($\eta$ = 0) on the CelebA-HQ dataset while with the stochastic sampler ($\eta$ = 1) on the FFHQ dataset. We observe that the LDM-DG$^{*}$ sampler converges to visually superior images in comparison to the vanilla DDIM. We compare performance on standard metrics --- FID~\citep{CleanFID21}, KID~\citep{DemistifyMMD18}, CLIP-FID~\citep{ClipFID}, and precision-recall~\citep{ImprovedPR19} scores. As we can observe from Table~\ref{Table_LDM}, LDM+DG$^{*}$ outperforms the baseline in CLIP-FID, Clean-FID and KID. We also carried out comparisons when using a trainable discriminator for guidance in LDM, similar to the LSGM-G++ setting proposed by~\citet{DiscGuidance23} on CelebA-HQ, where this baseline achieves a CLIP-FID value of 7.08, which is worse than that achieved by the proposed LDM+DG$^*$. Details are provided in Appendix~\ref{App_Ablations}. Given the results in Section~\ref{Sec:DiscInfusion} and the theoretical acceleration shown by DG$^{*}$, we also explore accelerating LDM+DG$^{*}$ using time-step shifted~\citep{TimeShift24} and DPM~\citep{DPMSolver22} solvers. \par

{\it {\bfseries Time-Shifted Sampling}}:~\citet{TimeShift24} proposed the time-shifted sampler to mitigate \textit{exposure bias} in DPMs caused due to poor inference-time generalization, \textit{i.e.,} $\epsilon_{\theta}$ is trained on ground-truth samples $\x_{t}$, but inference is performed on $\hat{\x}_{t-1}$, diverting samples from the intended trajectory. To mitigate this issue, given the sample $\hat{\x}_{t}$, an estimate of the noise variance in the image is used to evaluate and transition to a new coupling time $t_{s}$. Further, they also show that diffusion models basically contain \textit{two stages} -- The initial phase, wherein the input Gaussian distribution moves towards the image space, and the second phase, wherein patterns and structure emerge from latching onto a specific image to generate. Time-step shifting and the proposed DG$^*$ therefore operate in the first stage, which is where we focus the discriminator guidance.

Motivated by the above setting, and the observation in Section~\ref{Sec:DiscInfusion} that applying LDM+DG$^*$ for all time steps may be unnecessary, we adopt the time-shifted discriminator-guided diffusion strategy to ensure that the effect of discriminator guidance is restricted to the earlier, exploratory step. We also improve upon the noise-variance estimation technique proposed in the baseline. In particular, based on image denoising literature~\citet{MallatWavelets,Donoho95} we use the Haar wavelet representation to estimate noise as \(\Tilde{\sigma} = \frac{M_{\x}}{0.6745}\), wherein \(M_{\x}\) is the median of the absolute of the wavelet coefficients of the image \(\x\), and one level of decomposition suffices. The details are presented in Appendix~\ref{App:NoiseVar}. We refer to the wavelet-based noise estimation for DG$^*$-guided acceleration as WANDA. Table~\ref{Table_LDM} presents various evaluation metrics, when sampling using WANDA, compared against the baseline LDM, and LDM+DG$^*$ approaches. Figure~\ref{Fig_LDMs} presents the images generated by the proposed approach. WANDA achieves comparable performance, while running fewer sampling steps than the baseline.

\begin{figure*}[!t]
  \begin{center}
    \begin{tabular}[!t]{P{.29\linewidth}|P{.29\linewidth}|P{.29\linewidth}}
     LDM & LDM+DG$^*$ (\textbf{Ours}) & WANDA (\textbf{Ours}) \\
     \includegraphics[width=0.99\linewidth]{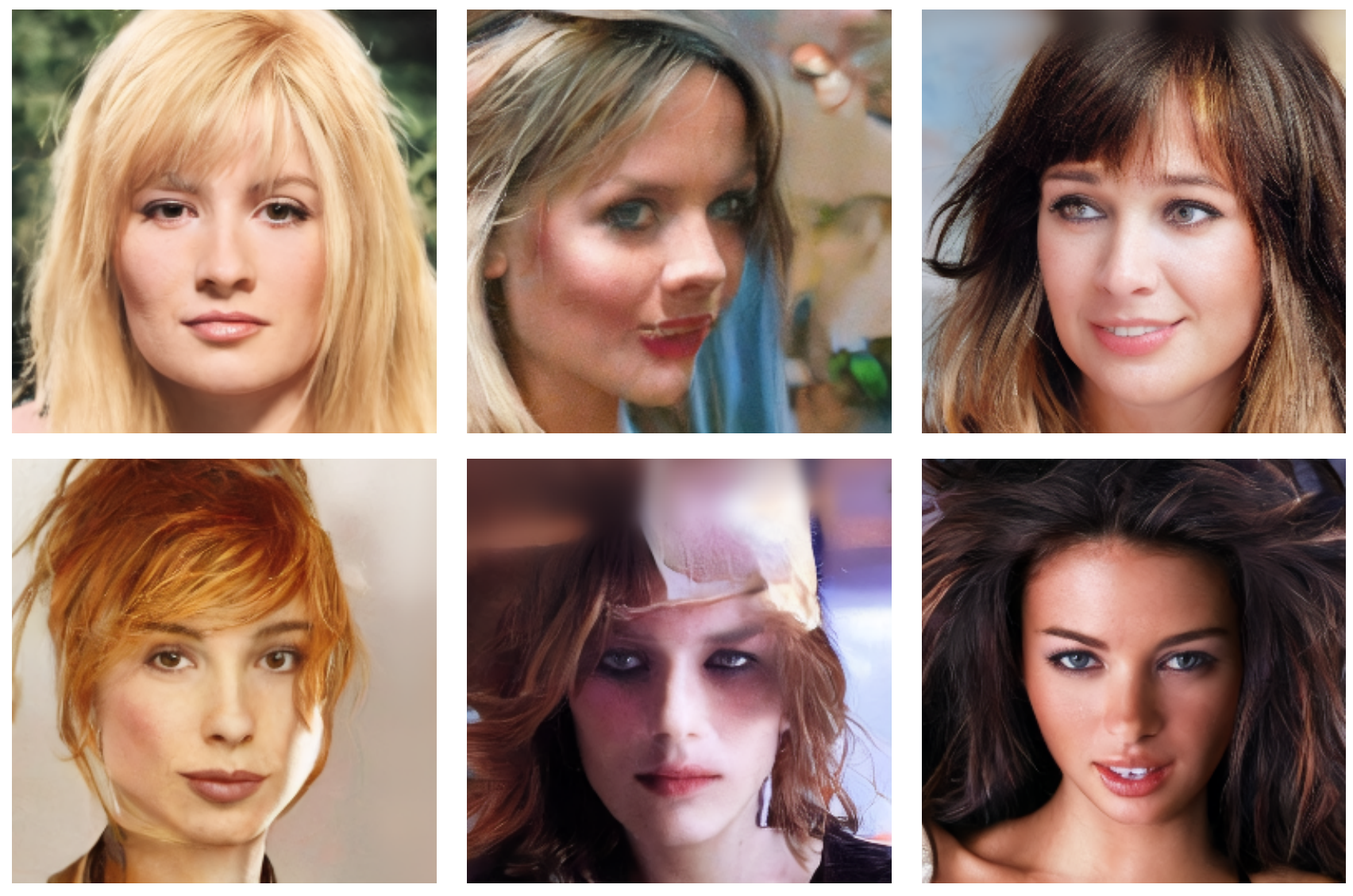}  &\includegraphics[width=0.99\linewidth]{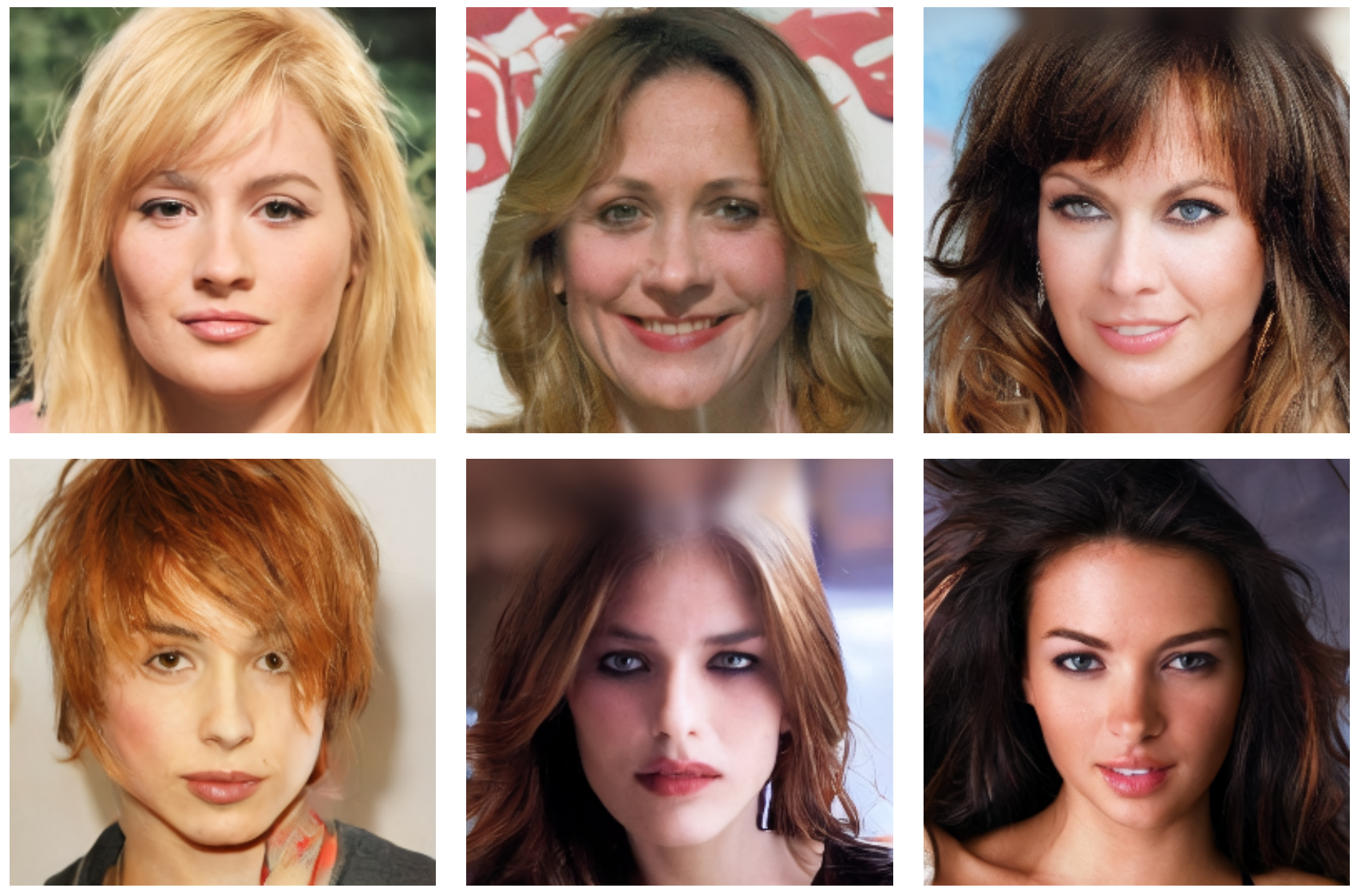}  &\includegraphics[width=0.99\linewidth]{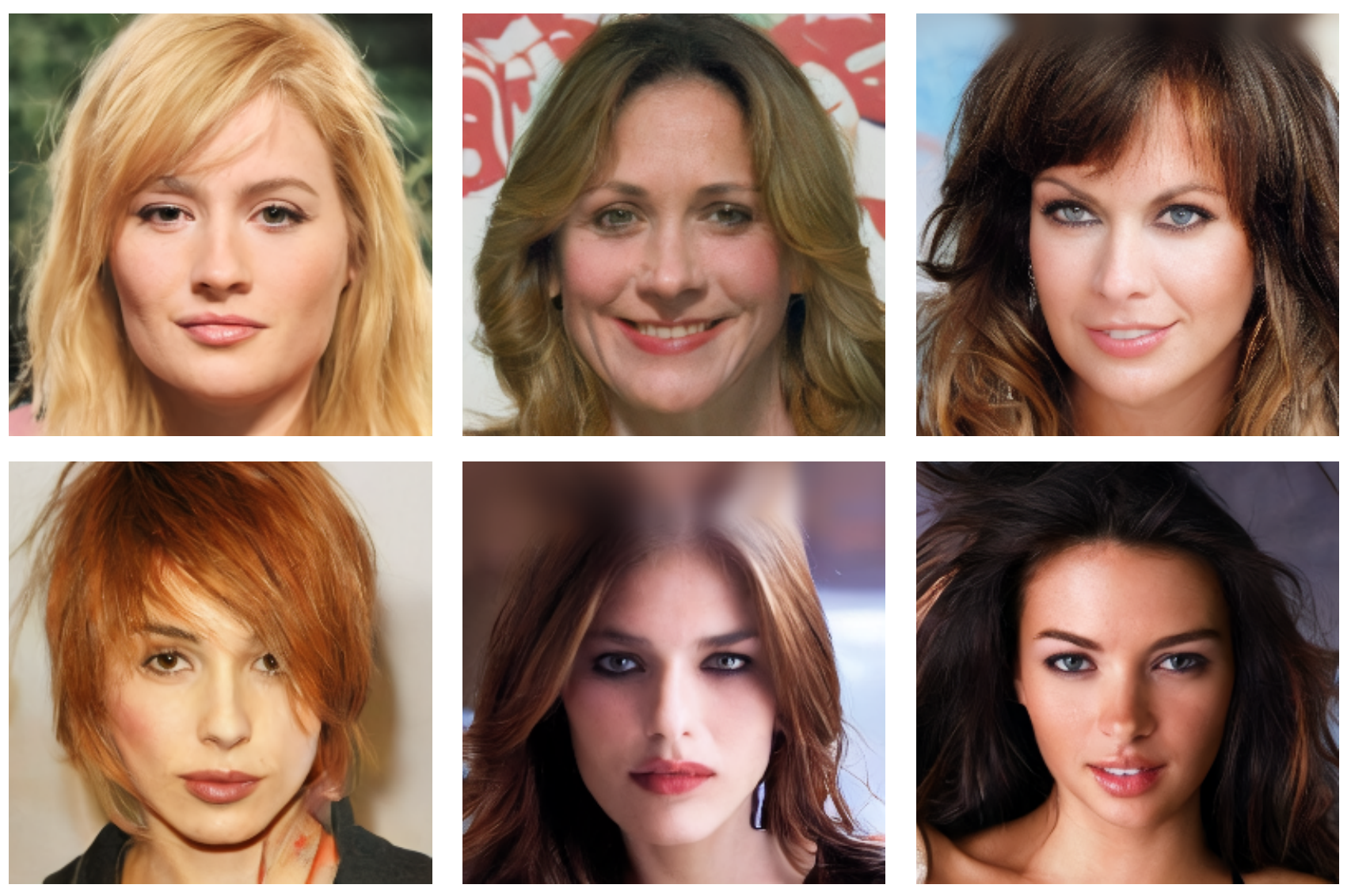} 
    \end{tabular} 
  \caption[]{(\includegraphics[height=0.009\textheight]{Rgb.png} Color online)~A comparison of the 256-dimensional CelebA-HQ images generated (given the same input) by the baseline LDM, and the proposed closed-form discriminator guidance models without and with time-step-shifted sampling (LDM-DG$^*$ and WANDA, respectively). LDM-DG$^*$ significantly improves the generated image quality, by removing artifacts. WANDA generates images with a quality comparable to that of LDM-DG$^*$, with relatively fewer function evaluations.}
  \label{Fig_LDMs}  
  \end{center}
  \vskip-1em
  \end{figure*}

{\it {\bfseries DPM Solver:}} The proposed DG$^*$ term is orthogonal to baselines acceleration schemes such as~\citet{DPMSolver22,FastODE24}, wherein better ODE solvers are used to accelerate sampling. As such, DG$^*$ can be combined with these techniques as well. As a proof of concept, we present an ablation on CelebA-HQ, considering the DPM solver~\citep{DPMSolver22}, with and without +DG$^*$. Exhaustive results are provided in Table~\ref{Table_DPMSolver} of the Appendix. We observe that, for $T=20$, the baseline achieved a CLIP-FID of 9.5. Sampling including discriminator guidance allows us to further accelerate the sample generation process, with the DPM+DG$^*$ sampler achieving comparable performance (CLIP-FID or 9.71) in $T=15$ steps (1 discriminator step with 14 DPM solver steps). On the other hand, the DPM+DG$^*$ with $T=20$ outperforms the baseline, with a CLIP-FID of 9.22. \par

We also report comparisons on the LSUN-Churches and CIFAR-10 datasets, and ablations on the choice of the decay parameter, \(w_{dg,t}\) and linear vs. exponential decay, the number of discriminator guidance steps \(T_D\), etc. are provided in Appendix~\ref{App_Ablations}.

\section{Conclusion}
\label{sec:conclusion}
In this paper, we considered the setting of discriminator guidance in diffusion models, and developed strong theoretical links between IPM-GAN generator optimization and the smoothed score-matching condition. Based on this novel insight, we developed a kernel-based closed-form discriminator guidance framework (DG$^*$) that can be applied in a \textit{plug-and-play} fashion to any existing diffusion model. We demonstrated the feasibility of this approach by applying DG$^*$ to DDIMs and LDMs, resulting in superior image quality at no additional training cost. We also demonstrated the interoperability of DG$^*$ with existing acceleration schemes such as time-step-shifted diffusion, or other solvers such as DPM. While the presented experiments demonstrate the versatility of the closed-form IPM-GAN discriminator guidance approach, applications to other state-of-the-art diffusion models and acceleration techniques are promising directions for future research. 

\section*{Acknowledgments}
\label{app:ack}
Siddarth Asokan was supported by the Microsoft Research PhD Fellowship, Robert Bosch Centre for Cyber Physical System (RBCCPS) Ph.D. Fellowship 2020 and 2021 and the Qualcomm Innovation Fellowship 2019, 2021, 2022 and 2023 during his tenure as a Ph.D. student at the Indian Institute of Science (IISc). Nishanth Shetty is supported by the Qualcomm Innovation Fellowship 2023 and 2024, and the Prime Minister's Research Fellowship.
\section*{Impact Statement}
\label{app:ethics}

In this paper, we present results with goal is to advance the field of understanding diffusion and GAN models, by providing a theoretical framework that unifies diffusion models and optimization in GANs. The broader societal impact of this research is not beyond those inherent in generative modeling research. These include potential applications in AI-assisted content creation, digital media, and creative industries, where high-fidelity image synthesis can be either beneficial or harmful. As with all generative modeling techniques, there exist ethical considerations regarding potential misuse, including the generation of deepfakes and synthetic content that could be used for misinformation, and we emphasize that these decisions are to be taken by the researchers that partake in the usage of these models, and that our work, on better understanding how these models work, builds upon existing ethical safeguards in generative AI research.

\bibliography{references}
\bibliographystyle{icml2025}

\newpage
\appendix

\appendix

\addcontentsline{toc}{section}{Appendix} 
\part{Appendix} 
\parttoc
\section{Computational Resources}

All experiments were carried out using TensorFlow 2.0~\citep{TF} and PyTorch ~\citep{PyTorch} backend. Experiments on NCSN, EDM, and LDM were built atop publicly available implementations (URL: \url{https://github.com/Xemnas0/NCSN-TF2.0}, \url{https://github.com/NVlabs/edm}, and \url{https://github.com/CompVis/latent-diffusion}, respectively). Experiments were performed on SuperMicro workstations with 256 GB of system RAM comprising two NVIDIA GTX 3090 GPUs, each having 24 GB VRAM, and NVIDIA RTX A6000 with 8 GPUs.

\section{Code Repository and Animations}
The TF 2.0~\citep{TF} based source code for implementing discriminator-guided Langevin diffusion and LDM-based experiments are accessible at \url{https://github.com/DarthSid95/ScoreFloWGANs}. Additionally, we have also provided animations corresponding to the {\it Shape Morphing} experiments presented in Figure~\ref{PlotApp_Morphing}, and the images generated in Figures~\ref{Fig_BetaCompares_MNSIT}--\ref{Fig_BetaCompares_CelebA}, Figure~\ref{Fig_DiscDiffusion} and Figure~\ref{Fig_LDMs}. Full-resolution versions of images presented in the paper will also be made accessible in the GitHub Repository.

\newpage

\section{Preliminaries and Background}
\subsection{Mathematical Preliminaries}\label{App_MathPrelims}
Consider a vector \(\z = [z_1, z_2,\,\ldots\,,z_n]^{\rmT} \in \mbbR^n\) and the generator \(G:\mbbR^n \rightarrow \mbbR^n\), {\it i.e,.}, \(G(\z) = [G_1(\z),G_2(\z),\ldots;G_n(\z)]^{\rmT}\), where $G_i(\z)$ denotes the $i^{th}$ entry of $G$. The notation \(\nabla_{\z} G(\z)\) represents the gradient matrix of the generator, with entries consisting of the partial derivatives of the entries of \(G\) with respect to the entries of $\z$ and is given by
\begin{align*}
\nabla_{\z} G(\z) &= \left[\begin{matrix}
\frac{\partial G_1}{\partial z_1} & \frac{\partial G_2}{\partial z_1} & \ldots &\frac{\partial G_n}{\partial z_1} \\[3pt]
\frac{\partial G_1}{\partial z_2} & \frac{\partial G_2}{\partial z_2} & \ldots &\frac{\partial G_n}{\partial z_2} \\[3pt]
\vdots&\vdots&\ddots&\vdots\\
\frac{\partial G_1}{\partial z_n} & \frac{\partial G_2}{\partial z_n} & \ldots &\frac{\partial G_n}{\partial z_n} 
\end{matrix} \right].
\end{align*}
The Jacobian \(\rmJ\) {\it measures} the transformation that the function imposes locally near the point of evaluation and is given as the transpose of the gradient matrix, {\it i.e.,} \(\ \rmJ_G(\z) = (\nabla_{\z}G(\z))^\rmT\). 

{\it{\bfseries Calculus of Variations}}: Our analysis centers around deriving the optimal generator in the functional sense, leveraging the {\it Fundamental Lemma of the Calculus of Variations}~\citep{HistoryOfCalcVar,CalcVarPhy04}. Consider an integral cost \(\loss\), to be optimized over a function \(h\):
\begin{align}
\mathcal{L}\left( h, h^{\prime}\right)=\int \limits_\mcalX \mathcal{F}\left(\x, h(\x), h^{\prime}(\x)\right)~\rmd \x\,, 
\label{integral_cost}
\end{align}
where \(h\) is assumed to be continuously differentiable or at least possess a piecewise-smooth derivative \( h^\prime(\x)\) for all \(\x\in\mcalX\). If $h^*(\x)$ denotes the optimum, The {\it first variation} of \(\loss\), evaluated at \(h^*\), is defined as the derivative \(\delta\loss(h^*;\eta) = \fracpartial{\loss_{\epsilon}(h^*)}{\epsilon} \) evaluated at \(\epsilon=0\), where \(\loss_{\epsilon}(h^*) \) denotes an \(\epsilon\)-perturbation of the argument \(h\) about the optimum \(h^*\), given by
\begin{align*}
 \loss_{h,\epsilon}(\epsilon) &= \mathcal{L}\left( h^*(\x) + \epsilon\,\eta(\x), h^{*{\prime}}(\x) + \epsilon \,\eta^{\prime}(\x)\right) 
\end{align*}
where, in turn, \(\eta(\x)\) is a family of {\it perturbations} that are compactly supported, infinitely differentiable functions, and vanishing on the boundary of \(\mcalX\). Then, the optimizer of the cost \(\loss\) satisfies the following first-order condition: 
\begin{align*}
\frac{\partial  \loss_{h,\epsilon}(\epsilon)}{\partial \epsilon} \bigg|_{\epsilon = 0} = 0
\end{align*}
Another core concept in deriving functional optima is the {\it Fundamental Lemma of Calculus of Variations}, which states that, if a function \(g(\x)\) satisfies the condition
\begin{align*}
\int_{\mcalX} g(\x)\,\eta(\x)~\rmd \x = 0
\end{align*}
for all compactly supported, infinitely differentiable functions \(\eta(\x)\), then \(g\) must be identically zero almost everywhere in \(\mcalX\). Together, these results are used to derive the condition that the optimal generator transformation satisfies, within various GAN formulations. \par

\subsection{Diffusion Probabilistic Models}
Diffusion probabilistic models (DPMs) primarily model the {\it forward process} wherein Gaussian noise is progressively added to an image \(\x\sim\pd\). The noise is modelled as adhering to a fixed variance schedule $\beta(t)$. The generative task is one of modeling the reverse process, essentially iterated denoising. Given the data distribution $\pd$ and a fixed noise schedule $\beta(t) \in (0,1), \forall t = 1 \ldots T$, the forward process, structured as a Markov process, is expressed as $p(\x_{1,2,\ldots, T}|\x_{0}) = \prod^{T}_{t=1} p(\x_{t}|\x_{t-1}).$ In the DPM setting, the forward transition kernel at time \(t\), given by $p(\x_{t}|\x_{t-1})$ can be defined as a Gaussian \(\mathcal{N}(\sqrt{\alpha_{t}}\x_{t-1}, \beta_{t}\mbbI)\), centered around the sample of the previous time instant \(\sqrt{\alpha_{t}}\x_{t-1}\), where $\alpha_{t} = 1 - \beta_{t}$~\citep{DDPM20}. By means of the reparameterization trick, the conditional distribution can be expressed as: 
\begin{align}
    \x_{t} = \sqrt{\Bar{\alpha}_{t}}\x_{0} + \sqrt{1 - \Bar{\alpha}_{t}}\epsilon_{t} \quad \Rightarrow \quad p(\x_{t-1}|\x_{t}, \x_{0}) = \mathcal{N}(\Tilde{\mu}_{t}, \Tilde{\beta_{t}})
\label{eqnSupp: nice_property}
\end{align}
wherein, $\Bar{\alpha}_{t} = \prod^{t}_{i=1} \alpha_{i}$ and $\epsilon_{t} \sim \mathcal{N}(\bm{0}, \mbbI)$, $\Tilde{\mu}_{t} = \dfrac{1}{\sqrt{\alpha_{t}}}\left(\x_{t} - \dfrac{1 - \alpha_{t}}{\sqrt{1 - \Bar{\alpha}_{t}}}\epsilon_{t}\right)$,  $\Tilde{\beta_{t}} = \dfrac{(1 - \Bar{\alpha}_{t-1})}{1 - \Bar{\alpha}_{t}}\beta_{t}$ and \(p(\x_0) = \pd\). Training DPMs involves learning a neural network $\epsilon_{\theta}$ to approximate $\epsilon_{t}$, with the following MSE loss~\citet{DDIM21}:
\begin{align}
    \loss_{\mathrm{DPM}} &= \mathbb{E}_{t, \x_t, \epsilon_{t} \sim \mathcal{N}(0, \mathbb{I})} [\left\| \epsilon_{\theta}(\x_{t}, t) - \epsilon_{t}\right\|_{2}^{2}]
\label{eqnSupp: ddpm_loss}
\end{align}
In practice, the model is trained on a variational lower bound of the negative log-likelihood loss. Consequently, generation starts by sampling \(\x_T\) from a standard Gaussian, {\it i.e.,} \(\x_T\sim\mcalN(\bm{0},\mbbI)\), and progressively generating samples according to the backward recursion:
\begin{align*}
    &\x_{t-1} = \mu_{\theta}(\x_t,t) + \Sigma_{\theta}(\x_t,t).\z_t,~t = T,T-1,\ldots,0,
\end{align*}
where \(\z_t\sim\mcalN(\bm{0},\mbbI)\), and \(\mu_{\theta}\) and \(\Sigma_{\theta}\) are the estimates of the noise mean and covariance, as output by \(\epsilon_{\theta}\). The SDE governing the above process was generalized by~\citet{DDIM21}, and in general, can be written as:
\begin{align}
    \rmd \bmX_t &= \left( f(t) + g^2(t) \nabla_{\bmX} \ln p_t^*(\bmX_t) \right)\rmd t + g(t) \rmd\bmW_t, \label{EqnSupp_DiffusionBase}
\end{align} 
for suitable function \(f\) and \(g\), where \(\rmd\bmW\) refers to the standard Weiner process. We refer the reader to~\citep{DDIM21} for an in-depth analysis for the choice of these functions. The discretized update is then given by:
\begin{equation}
    \begin{split}
        \x_{t-1} &= \underbrace{\sqrt{\frac{\alpha_{t-1}}{\alpha_t}}\x_t - \sqrt{\frac{\alpha_{t-1}}{\alpha_t}}\sqrt{(1-\alpha_t)}\epsilon_{\theta}(\x_t,t)}_{\hat{\x}_0} + \sqrt{(1-\alpha_{t-1}) - \sigma^2_{t}}\cdot\epsilon_{\theta}(\x_t,t) +  \sigma_{t}\epsilon_{t}
    \end{split}
\label{eqnSupp:ddim_update}
\end{equation}
where $\hat{\x}_0$ can be viewed as the \textit{prediction} of  $\x_0$; the term $\sqrt{(1-\alpha_{t-1}) - \sigma^2_{t}}\cdot\epsilon^{t}_{\theta}(\x_t)$ represents the direction pointing towards $\x_{t}$ with $\alpha_{0}=1$; and $\sigma_{t}\epsilon_{t}$ is the diffusion term with $\epsilon_t \sim \mathcal{N}(0, \mathbb{I})$ being standard Gaussian and independent of $\x_t$. Different values of $\sigma$ lead to different generative processes while keeping $\epsilon_{\theta}$ fixed, thus removing the necessity to retrain the models. When $\sigma_{t}$ is set to $\sqrt{(1-\alpha_{t-1})/(1-\alpha_{t})} \sqrt{(1-\alpha_{t}/\alpha_{t-1})}$, for all $t$, the resulting generative process becomes DDPM~\citet{DDIM21}. On the other hand, when $\sigma_t$ = 0 for all $t$, the samples generated obey a deterministic procedure and this specific generative trajectory is referred to as denoising diffusion implicit model (DDIM) sampling. DDIM sampling can generate high-quality samples with fewer time-steps $\tau < T$ with no changes in the training procedure of the DDPM denoiser \(\epsilon_{\theta}\) which was trained over $T$ timesteps. In general, we can set \(\sigma_{\tau(\eta)} = \eta \sqrt{(1-\alpha_{t-1})/(1-\alpha_{t})} \sqrt{(1-\alpha_{t}/\alpha_{t-1})}\) to interpolate between the DDPM and DDIM~\citep{DDIM21}. The choice of \(\eta\) controls the stochasticity in sampling, with $\eta =1$ and $\eta = 0$ corresponding to DDPM and DDIM, respectively. 

\subsection{Optimality of \(f\)-GANs} \label{App_fGANs}
GAN optimization can be viewed as minimizing either the \(f\)-divergence between the target distribution \(\pd\) and the distribution of the generated samples (denoted as \(\pg\)), or an integral probability metric (IPM) between \(\pd\) and \(\pg\). \citet{fGAN16} proposed \(f\)-GANs, considering \(f\)-divergences of the form:
\( \mathfrak{D}_f(\pd\Vert\,p_{t-1}) = \int_{\mcalX} f\left( r_{t-1}(\x)\right)\pd(\x) ~\rmd\x,\)
where \(f\colon\mbbR_{+}\rightarrow\mbbR\) is a convex, lower-semicontinuous function over the support \(\mcalX\) and satisfies \(f(1) = 0\) and \(r_{t-1}(\x)\) is the density ratio \(r_{t-1}(\x) = \frac{\pd(\x)}{p_{t-1}(\x)}\). The optimization is given by 
\begin{align}
    \min_{G} \max_{D} \left\{\mbbE_{\x \sim \pd}[T(\x)]-\mbbE_{\z \sim \pz}[f^{c}(T(G(\z)))] \right\} , \label{EqnSupp_fGAN_Opt} 
\end{align}
where \(T(\x) = g(D(\x))\), is the output of the discriminator $D$ subjected to the activation \(g\), and  \(D^*(\x)\) is the optimal discriminator, and \(f^c\) denotes the Fenchel conjugate of \(f\). In practice, the optimization is an alternating one, wherein the discriminator \(D_t\) is derived given the generator of the previous iteration \(G_{t-1}\), and the subsequent generator optimization involves computing \(G_{t}\), given \(D_{t}\) and \(G_{t-1}\). Within this setting,~\citep{fGANsScores23} presented the following result:
\begin{theorem} (\textbf{Formal},~\citep{fGANsScores23}
Consider the generator loss in \(f\)-GANs, given by Equation~\eqref{EqnSupp_fGAN_Opt}. The {\bfseries optimal \(f\)-GAN generator} satisfies the following score-matching condition:
\(r_{t-1}(\x) g^{\prime}(t)\big|_{t=D_t^*} D_t^{*\prime}(y)\big|_{y=\ln(r_{t-1})} \nabla_{\x} \left(\ln r_{t-1}(\x)\right)=\bm0,\) where \(g^{\prime}(t)\) denotes the derivative of the activation function with respect to \(D\) evaluated at \(D_t^*\),~  \(D_t^{*\prime}(y)\) denotes the derivative of the optimal discriminator function with respect to \(y = \ln(r_{t-1}(\x))\), evaluated at \(\ln(r_{t-1}(\x))\). For \(\z\) such that \(r_{t-1}(\x) g^{\prime}(t) D_t^{*\prime}(y) \neq 0\), the optimization yields the score-matching cost:
\begin{align*}
\nabla_{\x} \ln \left(p_{t-1}(\x) \right)\big|_{\x = G^*_t(\z)} &=  \DataScore{\x} \big|_{\x = G^*_t(\z)}.
\end{align*}
 \end{theorem}

\newpage

\section{Optimality of IPM-based GANs} \label{App_IPMGANs}
We now derive the proofs for theorems presented in the context of IPM GANs. The \(f\)-GAN counterparts are provided in~\citet{fGANsScores23}.
\subsection{Optimality of Kernel-based IPM-GANs (Proofs of Theorem~\ref{Theorem_IPMGAN_OptG} and Lemma~\ref{Lemma_IPMGAN_OptG})} \label{App_KernelGAN}

~\citet{SobolevGAN18}, in the context of SobolevGAN, showed that IPM-GANs with a gradient-based constraint defined with respect to a base density \(\mu(\x)\) results in the optimal discriminator solving the Fokker-Planck partial differential equation (PDE), given by:
\begin{align*}
    \mathrm{div.}\left(\mu\,\,\nabla D \right) \big|_{D = D_t^*(\x)} = \mathrm{c}\left(\pd(\x) - p_{t-1}(\x)\right),
\end{align*}
where \(\mathrm{div}\) denotes the divergence operator and \(\mathrm{c}\) is a constant. Considering a uniform base measure, ~\citet{ANON_JMLR} showed that the optimization results in a Poisson differential equation, while in the case of higher-order gradient penalties~\citep{BWGAN18,PolyGAN23}, the optimal discriminator is the solution to an iterated Laplacian equation, and generalizes the SobolevGAN formulation. The optimal discriminator that satisfies the iterated-Laplacian operator was shown to be~\citep{PolyGAN23}:
\begin{align*}
D_t^*(\x) &= \mathfrak{C}_{\kappa}  \left( \left( p_{t-1} - \pd \right) * \kappa \right) (\x),
\end{align*}
where \(\mathfrak{C}_{\kappa}  = \frac{(-1)^{m+1}\varrho}{2\lambda} \) and \(\varrho\) are positive constants, and the kernel \(\kappa\) is the Green's function associated with the differential operator. In Poly-WGAN, the kernel corresponds to the family of polyharmonic splines, given by
\begin{align*}
\kappa(\x) = \begin{cases}
\|\x\|^{\mathit{k}} &\text{if}~~\mathit{k}<0~~\text{or}~~n~\text{is odd}, \\
\|\x\|^{\mathit{k}} \ln(\|\x\|) & \text{if}~~\mathit{k}\geq0~~\text{and}~~n~\text{is even},%
\end{cases}
\end{align*}
where in turn, \(\mathit{k} = 2m - n\). The above was also shown to be an \(m^{th}\)-order generalization to the Plummer kernel considered in Coulomb GANs~\citep{CoulombGAN18}. Given the optimal discriminator, consider the generator optimization. Only the terms involving \(G(\z)\) influence the alternating optimization in practice, and the other terms can be neglected. Then, the cost is given by:
\begin{align*}
\loss^{\kappa}_G(G;D_t^*,G_{t-1}) &= - \Esub_{\z\sim\pz} [ D_t^*\left( G(\z)\right)] = -\int_{\mcalZ} D_t^*(G(\z))\,\pz(\z)~\rmd\z
\end{align*}
Let \(\loss_{G,i,\epsilon}\) denote the loss considering an $\epsilon$ perturbation of the \(i^{th}\) entry about the optimum, given by:
\begin{align*}
G_{t,i,\epsilon}^*(\z) = [G_{1,t}^*(\z), G_{2,t}^*(\z),~\ldots,~ G_{i,t}^*(\z) + \epsilon \eta(\z),~\ldots,~G_{n,t}^*(\z)]^{\rmT},
\end{align*}
where \(\eta(\z)\) is drawn from a family of compactly supported, infinitely differentiable functions. The loss can then be written as a function of \(\epsilon\). Consider the perturbed optimal generator \(G_{t,i,\epsilon}^*(\z)\), and the corresponding cost \(\loss_{G,i,\epsilon}(\epsilon)\). Substituting for \(D_t^*\) and expanding the convolution integral yields:
\begin{align}
  \loss^{\kappa}_{G,i,\epsilon}(\epsilon) &= -\int_{\mcalZ} \mathfrak{C}_{\kappa}\,\pz(\z) \int_{\mcalY} \left( p_{t-1}(G_{t,i,\epsilon}^*(\z) - \y) - \pd(G_{t,i,\epsilon}^*(\z) - \y)\right) \kappa(\y)~\rmd\y~\rmd\z,
  \label{Eqn_ExpandConv_IPM}
  \end{align}
where \(\mcalY\) is the union of the supports of \(\pd\) and \(p_{t-1}\) when they are overlapping, and the convex hull of their supports when non-overlapping. Differentiating the above with respect to \(\epsilon\) and setting it to zero at \(\epsilon = 0\) gives:
\begin{align*}
\frac{\partial \loss^{\kappa}_{G,i,\epsilon}(\epsilon)}{\partial \epsilon} \Bigg|_{\epsilon=0} &= -\int_{\mcalZ} \mathfrak{C}_{\kappa} \,\pz(\z) \int_{\mcalY} \left( p_{t-1}(\y) - \pd(\y)\right) \frac{\partial \kappa( G_{t,i,\epsilon}^*(\z) - \y)}{\partial \epsilon}\Bigg|_{\epsilon=0}~\rmd\y~\rmd\z \\
&= -\int_{\mcalZ} \mathfrak{C}_{\kappa} \,\pz(\z) \int_{\mcalY} \left( p_{t-1}(\y) - \pd(\y)\right) \frac{\partial \kappa( \w)}{\partial x_i}\Bigg|_{\w = G_{t}^*(\z) - \y} \!\!\!\frac{\partial [G_{t,i,\epsilon}^*(\z)]_i }{\partial \epsilon}~\rmd\y~\rmd\z \\
&= -\int_{\mcalZ} \mathfrak{C}_{\kappa} \,\pz(\z) \int_{\mcalY} \left( p_{t-1}(\y) - \pd(\y)\right) \frac{\partial \kappa( \w)}{\partial w_i}\Bigg|_{\w = G_{t}^*(\z) - \y} \eta(\z)~\rmd\y~\rmd\z = 0.
 \end{align*}
The inner integral represents a convolution, given by
\begin{align*}
\frac{\partial \loss^{\kappa}_{G,i,\epsilon}(\epsilon)}{\partial \epsilon} \Bigg|_{\epsilon=0} &= -\mathfrak{C}_{\kappa}  \int_{\mcalZ}  \left(\left( p_{t-1}- \pd \right) * \kappa_i^{\prime}\right)(\x)\bigg|_{\x = G_{t}^*(\z)} \pz(\z)\eta(\z)~\rmd\z = 0,
\end{align*}
where \(\kappa_i^{\prime}\) is the partial derivative of the kernel \(\kappa\) with respect to its \(i^{th}\) entry. From the {\it Fundamental Lemma of Calculus of Variations}, we have
\begin{align}
\mathfrak{C}_{\kappa} \left(\left( p_{t-1}- \pd \right) * \kappa_i^{\prime}\right)(\x)\bigg|_{\x = G_{t}^*(\z)} = 0,\qquad \forall~~\z\in\mcalZ. \label{Eqn_OptCond_Lemma}
\end{align}
Since the above holds for all \(i\), the above can be written compactly as
\begin{align*}
\mathfrak{C}_{\kappa} \left(\left( p_{t-1}- \pd \right) * \nabla_{\x}\kappa\right)(\x)\bigg|_{\x = G_{t}^*(\z)} = \bm{0},\qquad \forall~~\z\in\mcalZ,
\end{align*}
where the convolution between a scalar- and vector-valued function is carried out element-wise. This completes the proof of Lemma~\ref{Lemma_IPMGAN_OptG}. Table~\ref{Table_KenrelGradients} lists a few common kernels used across GAN variants and their corresponding gradient vectors.  \par

{\it \bfseries Proof of Theorem~\ref{Theorem_IPMGAN_OptG}}: An alternative approach to solving the aforementioned optimization, is to leverage the properties of convolution in Equation~\eqref{Eqn_OptCond_Lemma}. Consider the convolution integral:
\begin{align*}
 \left(\left( p_{t-1}- \pd \right) * \kappa_i^{\prime}\right)(\w) &=
   \int_{\mcalY} \left( p_{t-1}(\y) - \pd(\y)\right) \frac{\partial \kappa( \w)}{\partial w_i} \,\rmd\y\Bigg|_{\w = G_{t}^*(\z) - \y}  \\
  &= \frac{\partial }{\partial w_i} \left( \int_{\mcalY} \left( p_{t-1}(\y) - \pd(\y)\right) \kappa(\w)\,\rmd\y\right)\Bigg|_{\w = G_{t}^*(\z) - \y}\!\!\!\! = 0, \forall~~\z\in\mcalZ.
\end{align*}
From the property of convolutions, we have:
\begin{align*}
 \left(\left( p_{t-1}- \pd \right) * \kappa_i^{\prime}\right)(\w) &= \frac{\partial }{\partial w_i} \left( \int_{\mcalY} \left( p_{t-1}(\w) - \pd(\w)\right) \kappa(\y)\,\rmd\y\right)\Bigg|_{\w = G_{t}^*(\z) - \y}  \\
  &= \left( \int_{\mcalY} \left( \frac{\partial p_{t-1}(\w)}{\partial w_i}  - \frac{\partial\pd(\w) }{\partial w_i} \right) \kappa(\y)\,\rmd\y\right)\Bigg|_{\w = G_{t}^*(\z) - \y} \!\!\!\!\!\!= 0, \forall~~\z\in\mcalZ.
\end{align*}

Using the identity \( \dfrac{\partial p(\w)}{\partial w_i} = p(\w)\dfrac{\partial \ln p(\w)}{\partial w_i}\), we obtain:
\begin{align*}
 \left(\left( p_{t-1}- \pd \right) * \kappa_i^{\prime}\right)(\w) &=  \left( \int_{\mcalY} \left( \frac{\partial p_{t-1}(\w)}{\partial w_i}  - \frac{\partial\pd(\w) }{\partial w_i} \right) \kappa(\y)\,\rmd\y\right)\Bigg|_{\w = G_{t}^*(\z) - \y}\\
  &=  \left( \int_{\mcalY}\left(p_{t-1}(\y) \frac{\partial \ln(p_{t-1}(\y))}{\partial y_i}  - \pd(\y)\frac{\partial\ln(\pd(\y)) }{\partial y_i} \right) \kappa(\x- \y )\,\rmd\y\right) =0, \end{align*}
for all \(\z\in\mcalZ\) and \(\x = G_t^*(\z)\). Rewriting the integrals as expectations yields
  \begin{align*}
\Esub_{\y\sim p_{t-1}}\left[ \frac{\partial \ln(p_{t-1}(\y))}{\partial y_i} \kappa(G_t^*(\z) - \y ) \right] - \Esub_{\y\sim \pd}\left[ \frac{\partial \ln(\pd(\y))}{\partial y_i} \kappa(G_t^*(\z) - \y ) \right] = 0,\qquad \forall~~\z\in\mcalZ. 
\end{align*}
Stacking the above, for all \(i\), as a vector, we obtain:
\begin{align*}
  \Esub_{\y\sim p_{t-1}}\left[ \nabla_{\y}\ln(p_{t-1}(\y)) \kappa(G_t^*(\z) - \y ) \right] - \Esub_{\y\sim \pd}\left[ \nabla_{\y} \ln(\pd(\y))\kappa(G_t^*(\z) - \y ) \right] = \bm0,\qquad \forall~~\z\in\mcalZ. 
  \end{align*} 
 This completes the proof of Theorem~\ref{Theorem_IPMGAN_OptG}. \par

{\it {\bfseries Explaining Denoising Diffusion GANs}}: To derive a general solution to IPM-GANs (both network-based, or otherwise), consider the discriminator given at iteration \(t\),  \(D_t(\x)\). Then, the generator optimization is given by:
\begin{align*}
\loss^{IPM}_G(G;D_t,G_{t-1}) &= - \Esub_{\z\sim\pz} [ D_t\left( G(\z)\right)] = -\int_{\mcalZ} D_t(G(\z))\,\pz(\z)~\rmd\z
\end{align*}
The loss defined about the perturbed optimal generator is then given by:
\begin{align*}
\loss^{IPM}_{G,i,\epsilon}(\epsilon) &= -\int_{\mcalZ} D_t(G_{t,i,\epsilon}^*(\z))~\rmd\z \\
\Rightarrow\quad\quad  \frac{\partial \loss^{IPM}_{G,i,\epsilon}(\epsilon)}{\partial \epsilon} \Bigg|_{\epsilon=0} &=  \int_{\mcalZ}  \frac{\partial D_t(\x)}{\partial x_i}\bigg|_{\x = G_{t}^*(\z)} \pz(\z)\eta(\z)~\rmd\z = 0.
\end{align*}
A similar approach, as in the case of kernel-based IPM-GANs, to simplifying the above for all \(i\), results in the following optimality condition: 
\begin{align*}
 \nabla_{\x} D_t(\x)\big|_{\x = G_{t}^*(\z)} = \bm0,\quad\forall~\z\in\pz.
\end{align*}
While the above condition is essentially the optimality condition for gradient-descent over the discriminator in the context of gradient-descent-based training of GANs, it can be used to explain the optimality of GAN based diffusion models such as Denoising Diffusion GANs (DDGAN,~\citet{DDGAN22}). In DDGAN, a GAN is trained to approximate the reverse diffusion process, with time-embedding-conditioned discriminator and generator networks. While the approach results in superior sampling speeds as one only needs to sample from the sequence of generators, the underlying transformations that the generated images undergo, can be seen as the flow through the gradient field of the time-dependent discriminator as obtained above.
\begin{table}[t!]
\fontsize{8}{12}\selectfont
\begin{center}
\caption{Standard kernels considered in the GAN literature and their associated gradient fields.} \label{Table_KenrelGradients} 
\begin{tabular}{p{5.7cm}|p{2.75cm}|p{4.15cm}}
\toprule 
  Kernel &  \(\kappa(\x)\) & Gradient \(\nabla_{\x}\kappa(\x)\) \\ \midrule &&\\[-13pt]
  Radial basis function Gaussian (RBFG) \((\sigma > 0)\)  & \( \exp \left(  -\frac{1}{\sigma^2} \|\x\|^2 \right)\) & \(-\frac{1}{\sigma^2} \x \exp \left(  -\frac{1}{\sigma^2} \|\x\|^2 \right)\) \\[7pt]
Mixture of Gaussians (MoG) \(\left(\{\sigma_i>0\}_{i=1}^{\ell}\right)\)  &\( \sum_{\sigma_i}\exp \left(  -\frac{1}{\sigma_i^2} \|\x\|^2 \right)\) & \( -\x \left( \sum_{\sigma_i}\frac{1}{\sigma_i^2} \exp \left(  -\frac{1}{\sigma_i^2} \|\x\|^2 \right) \right)\) \\[7pt]
  Inverse multi-quadric (IMQ) \((c>0)\)
     & \( (\|\x\|^2 + c)^{-\frac12}\) & \( -\frac{1}{2}\x\,(\|\x\|^2 + c)^{-\frac{3}{2}}\)  \\[7pt]
 Polyharmonic spline (PHS) \((\mathit{k}<0~~\text{or}~~n~\text{is odd})\)    & \(\|\x\|^\mathit{k}\) & \((\mathit{k}-2)\x\|\x\|^{\mathit{k}-2}\) \\[7pt]
Polyharmonic spline (PHS) \((\mathit{k}\geq0~~\text{and}~~n~\text{is even})\)      &\(\|\x\|^\mathit{k}\ln(\|\x\|)\) & \(\x\|\x\|^{\mathit{k}-2}\left((\mathit{k}-2)\ln(\|\x\|) + 1 \right)\) \\[7pt]
\bottomrule
\end{tabular}
\end{center}
\end{table}

\textit{\textbf{Convergence of the Generator Distribution:}} Given the optimal discriminator \(D^*\),~\citet{PolyGAN23} showed that the generator distribution converges to the desired data distribution. For the sake of completeness, we summarize the Theorem here:
\begin{theorem}
    \citep{PolyGAN23}~(\textbf{Optimal generator density}\label{Lemma_pg}): Consider the minimization of the generator loss \(\loss_G\). The optimal generator density is given by \(\pg^*(\x) = \pd(\x),~\forall~\x\in\mcalX\). The optimal Lagrange multipliers are 
\begin{align*}
\lambda_p^* \in \mbbR\quad\text{and}\quad\mu^*_p(\x)  = \begin{cases}
0,&\forall~\x:~\pd(\x) > 0, \\
Q(\x) \in \mcalP_{m-1}^n(\x), &\forall~\x:~\pd(\x) = 0,
\end{cases}
\end{align*}
respectively, where \(Q(\x)\) is a non-positive polynomial of degree $m-1$, {\it i.e.,} \(Q(\x) \leq 0~\forall~\x\), such that \(\pd(\x) = 0\). The solution is valid for all choices of the homogeneous component \(P(\x)\in\mcalP_{m-1}^n(\x)\) in the optimal discriminator. 
\end{theorem}
\begin{proof}
As the cost function involves convolution terms, the Euler-Lagrange condition cannot be applied readily, and the optimum must be derived using the {\it Fundamental Lemma of Calculus of Variations}~\cite{GelfandCalcVar64}, as presented by\citet{PolyGAN23}. We recall a summary of the proof here for completeness. Consider the Lagrangian of the generator loss \(\loss_G\). Enforcing the first-order necessary conditions for a minimizer of the cost yields the following equation that the optimum solution \(\pg^*(\x)\) satisfies the equation
\(
\pg^*(\x) = \pd(\x) + \left(\frac{\lambda_d^*}{\xi} \right) \Delta^m \mu^*_p(\x).
\)
It is clear from the above solution that the optimum, \(\pg^*(\x)\), does not depend on the choice of the homogeneous component \(P(\x)\) in the optimal discriminator. The optimal Lagrange multipliers can be determined through dual optimization and enforcing the complementary slackness condition to obtain the result in above Theorem.
\end{proof}

\subsection{Sample Estimate of the Discriminator Gradient} \label{App_LemmaFloWGAN}
The proof follows closely the approach used in ~\citet{PolyGAN23}. Consider the optimality condition along a given dimension \(i\). We have:
\begin{align*}
\mathfrak{C}_{\kappa} \left(\left( p_{t-1}- \pd \right) * \kappa_i^{\prime}\right)(\x)\bigg|_{\x = G_{t}^*(\z)} = 0,\qquad \forall~~\z\in\mcalZ.
\end{align*}
Expanding the convolution integral yields
\begin{align*}
\mathfrak{C}_{\kappa} \int_{\mcalY}\left( p_{t-1}(\y)- \pd(\y) \right) \kappa_i^{\prime}( G_{t}^*(\z) -\y)~\rmd\y &= 0,\qquad \forall~~\z\in\mcalZ \\
\Rightarrow  \int_{\mcalY} p_{t-1}(\y)\,\kappa_i^{\prime}( G_{t}^*(\z) -\y)~\rmd\y -  \int_{\mcalY} \pd(\y)\,\kappa_i^{\prime}( G_{t}^*(\z) -\y)~\rmd\y &= 0,\qquad \forall~~\z\in\mcalZ \\
\Rightarrow \Esub_{\y\sim p_{t-1}} \left[ \kappa_i^{\prime}( G_{t}^*(\z) -\y) \right] - \Esub_{\y\sim \pd} \left[ \kappa_i^{\prime}( G_{t}^*(\z) -\y) \right] &= 0,\qquad \forall~~\z\in\mcalZ. 
\end{align*}
Replacing the expectations with their sample estimates yields
\begin{align*}
\sum_{\y_{\ell}\sim p_{t-1}} \kappa_i^{\prime}( G_{t}^*(\z) -\y_{\ell}) &= \sum_{\y_{\ell}\sim \pd}  \kappa_i^{\prime}( G_{t}^*(\z) -\y_{\ell}),\qquad \forall~~\z\in\mcalZ. 
\end{align*}
Evaluating the above at a sample level, for \(G_t^*(\z_t) = \x_t\), and stacking for all \(i\), we get the desired $N$-sample estimate of the discriminator gradient for the closed-form discriminator: 
\begin{align}
  \nabla_{\x}D_t^*(\x_t) &=   \mathfrak{C}^{\prime}_k\!\!\sum_{\bmg^j \sim \{\x_{t-1}\}} \nabla_{\x}\kappa(\x_t-\bmg^j)- \mathfrak{C}^{\prime}_k\!\!\sum_{\bmd^i \sim \pd} \nabla_{\x}\kappa(\x_t-\bmd^i).
\end{align}

\subsection{Choice of Discriminator Kernel} \label{App_Kernel}

\begin{figure*}[t!]
\begin{center}
  \begin{tabular}[b]{P{.02\linewidth}|P{.15\linewidth}P{.15\linewidth}P{.15\linewidth}P{.15\linewidth}P{.15\linewidth}P{.15\linewidth}}
    \rotatebox{90}{\footnotesize{ \qquad RBFG}} &
    \includegraphics[width=1.1\linewidth]{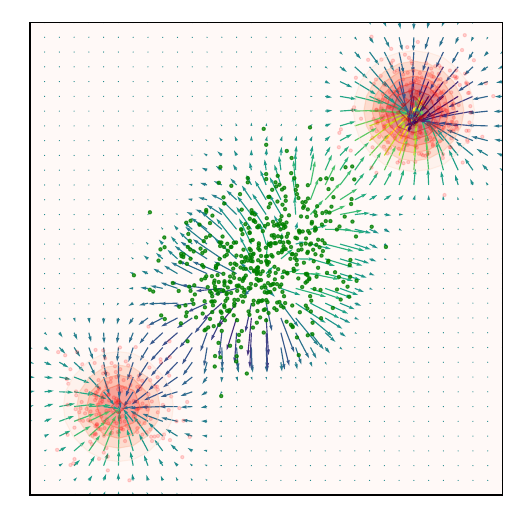} & 
    \includegraphics[width=1.1\linewidth]{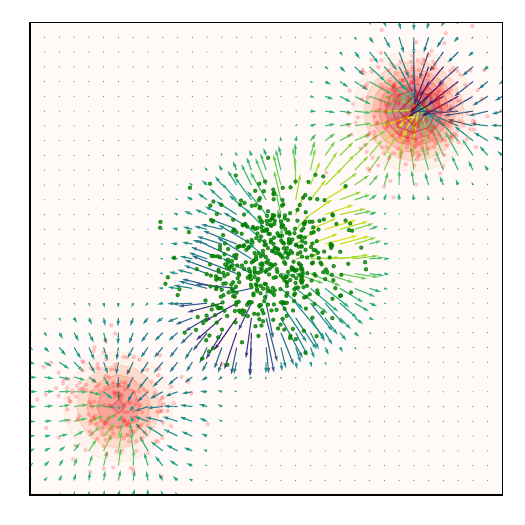} & 
     \includegraphics[width=1.1\linewidth]{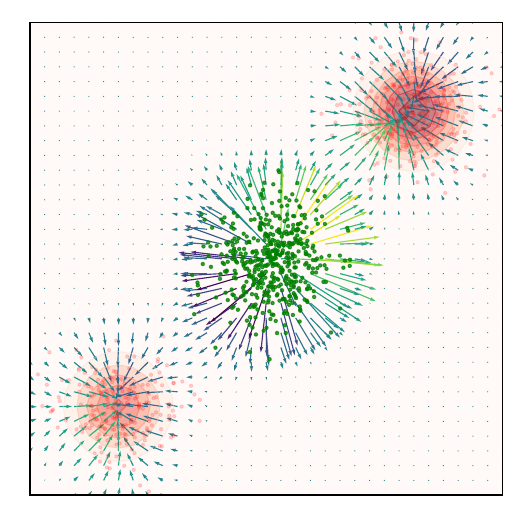} & 
     \includegraphics[width=1.1\linewidth]{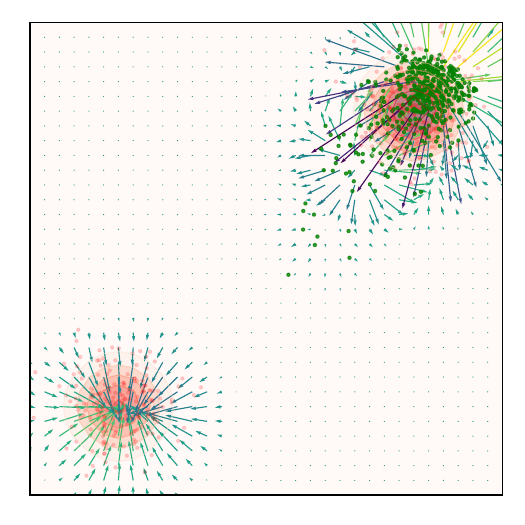} & 
    \includegraphics[width=1.1\linewidth]{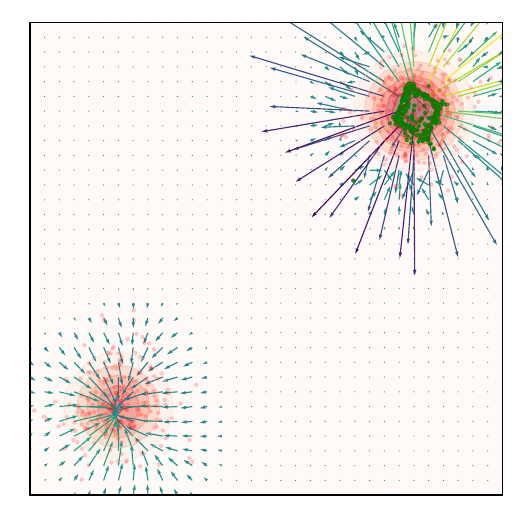} 
    \\[1pt] 
    
     \rotatebox{90}{\footnotesize{ \qquad MoG}} &
    \includegraphics[width=1.1\linewidth]{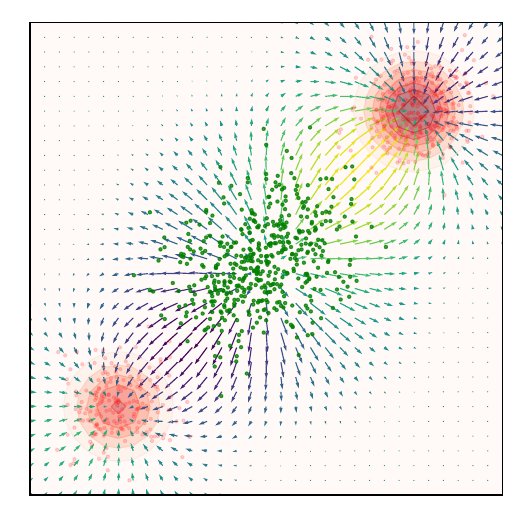} & 
    \includegraphics[width=1.1\linewidth]{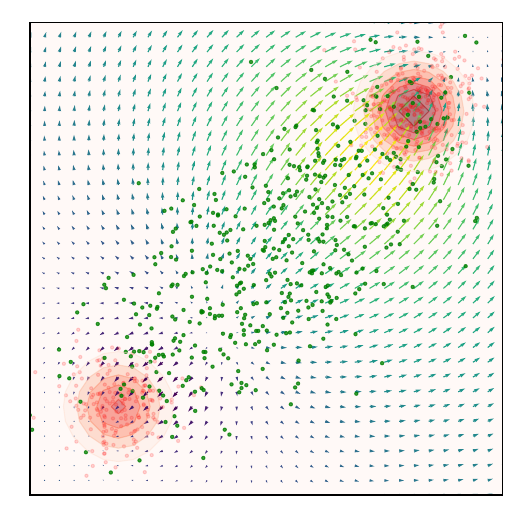} & 
     \includegraphics[width=1.1\linewidth]{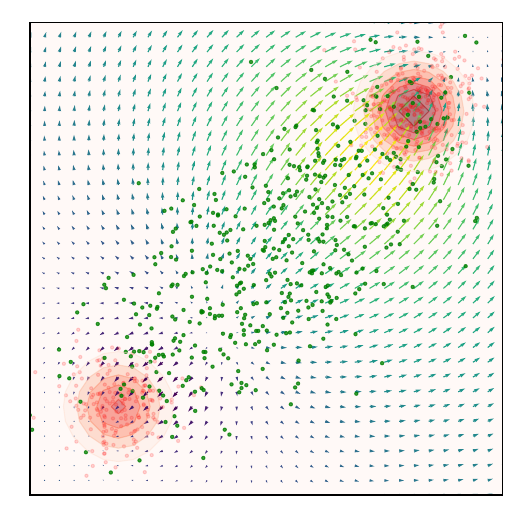} & 
     \includegraphics[width=1.1\linewidth]{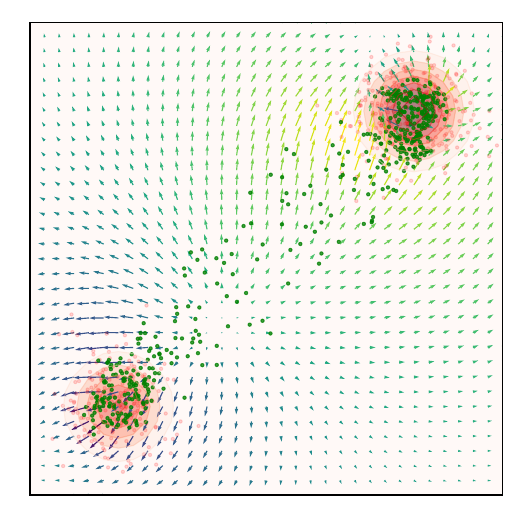} & 
    \includegraphics[width=1.1\linewidth]{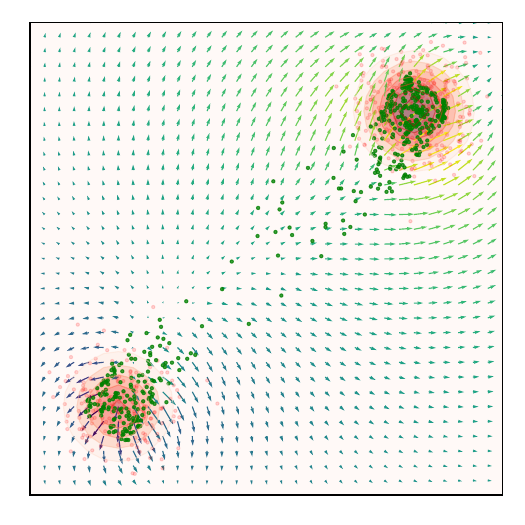}   \\[1pt] 
  
         \rotatebox{90}{\footnotesize{ \qquad\quad IMQ}} &
    \includegraphics[width=1.1\linewidth]{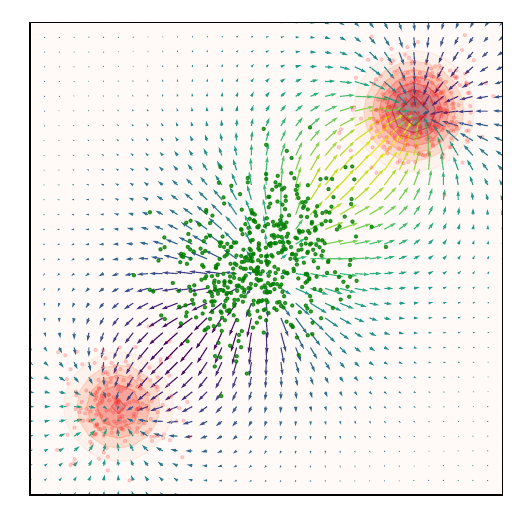} & 
    \includegraphics[width=1.1\linewidth]{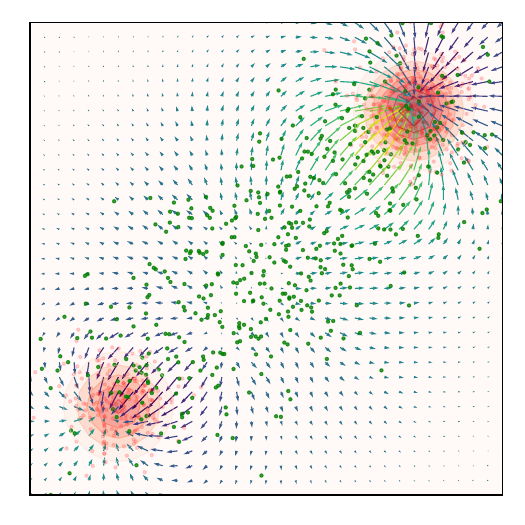} & 
     \includegraphics[width=1.1\linewidth]{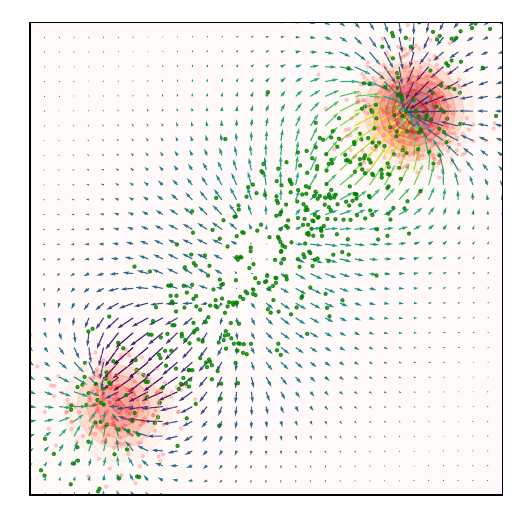} & 
     \includegraphics[width=1.1\linewidth]{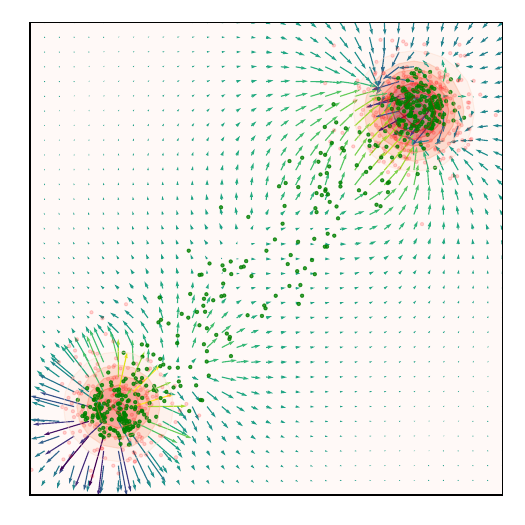} & 
    \includegraphics[width=1.1\linewidth]{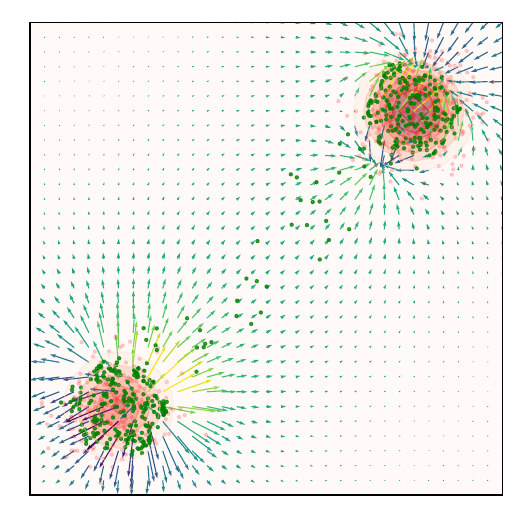}  \\[1pt] 
    
         \rotatebox{90}{\footnotesize{\qquad\quad PHS}} &
    \includegraphics[width=1.1\linewidth]{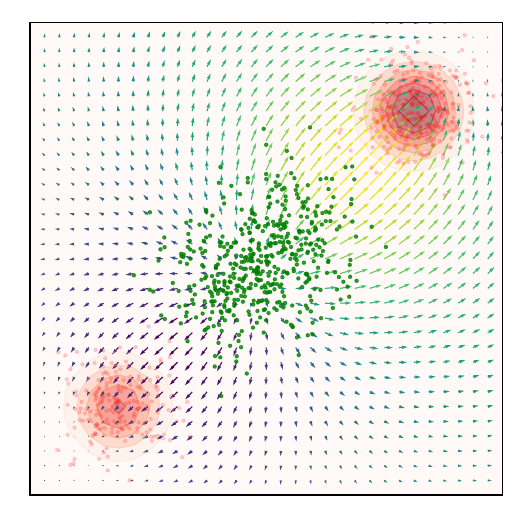} & 
    \includegraphics[width=1.1\linewidth]{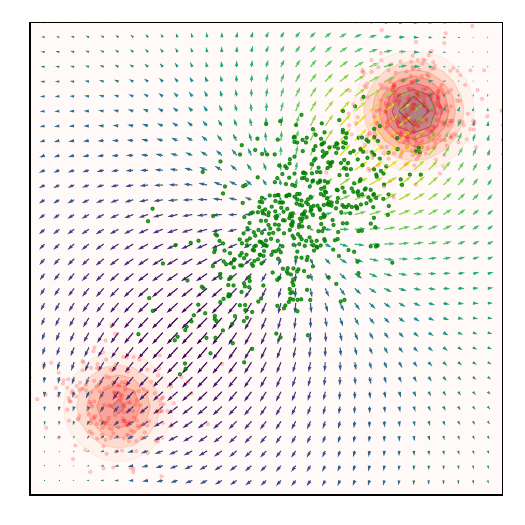} & 
     \includegraphics[width=1.1\linewidth]{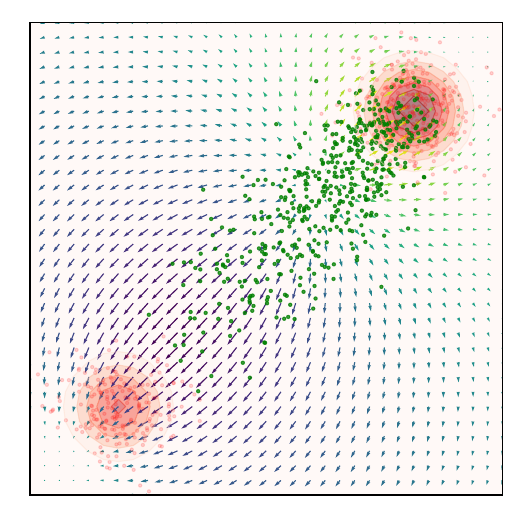} & 
     \includegraphics[width=1.1\linewidth]{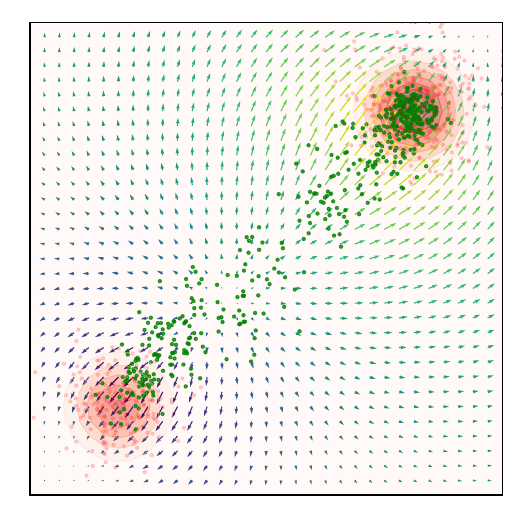} &
    \includegraphics[width=1.1\linewidth]{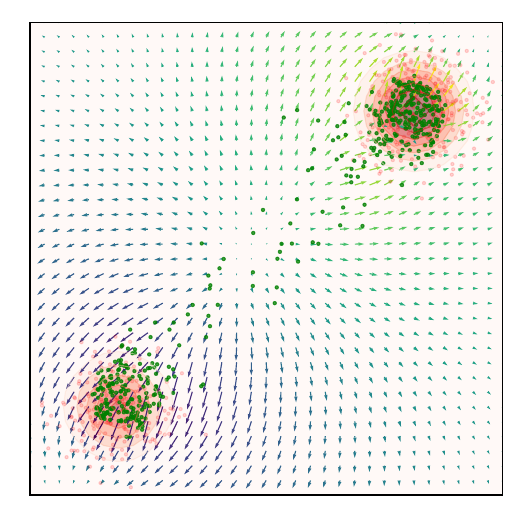}  \\[1pt] 
 
     & \scriptsize{10 iterations} & \scriptsize{50 iterations} &\scriptsize{500 iterations} & \scriptsize{1000 iterations} & \scriptsize{2500 iterations}  \\[1pt] 
    &\multicolumn{5}{c}{\includegraphics[width=0.75\linewidth]{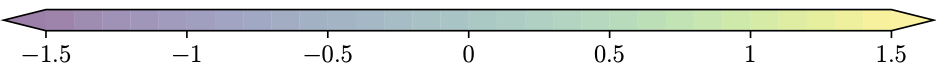}} \\[-1pt]  
  \end{tabular} 
\caption[]{(\includegraphics[height=0.009\textheight]{Rgb.png} Color online) Convergence of the generator samples (shown in \textcolor{ForestGreen}{green}) to the target two-component Gaussian (shown in \textcolor{red}{red}), \(\pd(\x) = \frac{1}{5}\mcalN(\x;-5\bm{1},\mbbI) + \frac{4}{5}\mcalN(\x;5\bm{1},\mbbI)\) considering various choices of the kernel function in FloWGAN. The quiver plot depicts the gradient field of the kernel convolved with the density difference. The single-component Gaussian kernel (RBFG) performs poorly if the chosen scale does not match the scale of the data. The mixture of Gaussians (MoG) kernel~\citep{MMDGAN17} alleviates this issue.  FloWGANs with the MoG, inverse multiquadric (IMQ) and Polyharmonic spline (PHS) kernel converge to the target data accurately.} 
  \label{Plot_GMMKernels_Iters}
  \end{center}
  \vskip-1.5em
\end{figure*}

Besides the Polyharmonic spline (PHS) kernel report in Main Manuscript, we also consider the radial basis function Gaussian (RBFG) and inverse multi-quadric kernels, as described in Table~\ref{Table_KenrelGradients}. As noted in the case of MMD-GANs~\citep{MMDGAN17}, the Gaussian kernel is sensitive to the scale parameter. Therefore, we consider two scenarios: (a) A single Gaussian kernel with \(\sigma = 1\); and (2) A mixture of five kernels with scale parameters \(\sigma \in \{0.5, 1, 2, 4, 8\}\). To simulate the performance of different kernels, we train a GAN generator, with the optimal, closed-form discriminators defined using the aforementioned kernel choices. Figure~\ref{Plot_GMMKernels_Iters} depicts the target and generated samples overlaid on the kernel gradient field. While the gradients in the IMQ kernel decay in regions far away from both \(\pd\) and \(\pg\), the gradient fields of the PHS and the {\it mixture of Gaussians} kernels is comparable. Since the polyharmonic function is not sensitive to a scale parameter, it converges to the target reliably for any input dynamic range. We therefore consider the PHS kernel in all experiments presented in Sections~\ref{Sec:DiscInfusion} and~\ref{Sec:LDM} and Appendix~\ref{App_DiscInfusion}.
 
\section{Theoretical Guarantees for closed-form IPM-GAN Discriminator Guidance} \label{App_WANDA}
\subsection{Convergence of Discriminator-guidance ODE}
An in-depth analysis of the convergence of discriminator-guided Langevin diffusion from the perspective of stochastic differential equations (SDEs) is outside the scope of this paper. However,~\citep{AdvReg18}, in the context of adversarial regularization for inverse problems, have extensively analyzed the following iterative algorithm:
\begin{align*}
  \x_{t+1} = \x_t - \eta \nabla_{\x} D^*_{t,\theta}(\x),
\end{align*}
where \(\eta\) is the learning rate, and \(D^*_{t,\theta}(\x)\) denotes the optimal discriminator at time \(t\) parameterized by \(\theta\). In particular, they show that~(\citet{AdvReg18}, Theorem 1):
\begin{align*}
\frac{\partial}{\partial \eta} \mcalW(\pd,p_t) = -\Esub_{\x\sim p_{t-1}} \left[ \| \nabla_{\x}D^*_{t,\theta}(\x)\|_2^2\right],
\end{align*}
where \(\mcalW\) denotes the Wasserstein-1 or Earthmover's distance. This shows that, the updated distribution \(p_{t}\) is closer in Wasserstein distance to the target distribution \(\pd\), in comparison to \(p_{t-1}\). For functions with \(\| \nabla_{\x}D^*_{t,\theta}(\x)\| = 1\), which is the condition under which the gradient-regularized GANs have been optimized, we have the decay \(\frac{\partial}{\partial \eta} \mcalW(\pd,p_t) = -1\). While we consider the updates 
\begin{align*}\x_{t+1} = \x_{t} - \alpha_t \nabla_{\x}D_t^*(\x_t) + \gamma_{t}\z_t\end{align*} in discriminator-guided Langevin diffusion, we will show, experimentally, that the update scheme \(\x_{t+1} = \x_{t} - \alpha_0 \nabla_{\x}D_t^*(\x_t)\) indeed performs the best, on image datasets  (cf. Appendix~\ref{App_DiscInfusion}). 
\newpage

\subsection{Convergence of Discriminator-guided Langevin Diffusion}\label{App_Convergence}
We provide a preliminary analysis of the convergence of the closed-form discriminator guided Langevin diffusion in a fashion similar to~\citep{DiscGuidance23}. For consistency with the literature, we fall back to the some of the notation of~\citep{DiscGuidance23}. Before we proceed, as a preliminary, we recall the Girsanov Theorem. Consider two diffusion process, 
\begin{align*}
    \rmd \bmX_t &= \mu_1(\bmX_t)\rmd t + \sigma(t) \rmd\bmW_t,~\text{and} \\
    \rmd \bmY_t &= \mu_2(\bmY_t)\rmd t + \sigma(t)\rmd\bmW_t,
\end{align*}
with identical diffusion terms, and associated densities \(p_1\) and \(p_2\). Then, the Girsanov theorem states that the Radon-Nikodym derivative (the ratio of probability densities) between these processes is given by:
\begin{align}
    \frac{\rmd p_1}{\rmd p_2} = \exp \left\{  \int \left(\frac{\mu_1 - \mu_2}{\sigma(t)}\right)\rmd\bmW_t +  \frac{1}{2} \int \left(\frac{\mu_1 - \mu_2}{\sigma(t)}\right)^2\rmd t\right\}.
\end{align}
Then, we have:
\begin{align*}
    \mathcal{D}_{KL}(p_1\Vert p_2) &= \mbbE_{p_1}\left[ \ln \left(\frac{\rmd p_1}{\rmd p_2}\right)\right] \\
    &= \mbbE_{p_1}\left[\int\left(\frac{\mu_1 - \mu_2}{\sigma(t)}\right)\rmd\bmW_t \right] + \frac{1}{2} \mbbE_{p_1}\left[ \int \left(\frac{\mu_1 - \mu_2}{\sigma(t)}\right)^2\rmd t\right] \\
    &= \frac{1}{2} \mbbE_{p_1}\left[ \int \left(\frac{\mu_1 - \mu_2}{\sigma(t)}\right)^2\rmd t\right] ,
\end{align*}
where the last equality is due to the martingale property of the \(\bmW_t\). In the context of the proposed discriminator guidance, we have the following two diffusion processes:
\begin{align}
    \rmd \bmX_t &= \left( f(t) + g^2(t) \nabla_{\bmX} \ln p_t^*(\bmX_t) \right)\rmd t + g(t) \rmd\bmW_t,~\text{and} \label{EqnApp_DiffusionBase}\\
    \rmd \bmY_t &= \left( f(t) + g^2(t) {\epsilon}_{\theta}(\bmY_t)+ h(t) \nabla_{\bmX} D_t^*(\bmY_t)  \right)  \rmd t + g(t)\rmd\bmW_t,  \label{EqnApp_DiffusionOurs}
\end{align}
associated with the target reverse process, and the discriminator guided score-based reverse process, respectively, where \(h(t)\) models the weight associated with the discriminator guidance term. The following Lemma gives us a convergence result on the discriminator guidance:
\begin{lemma}
    Consider the reverse diffusion processes associated with the base score-based approach, and the proposed closed-form discriminator guidance model. Let the probability densities associated with these two processes be \(p^*_t\) and \(p_t\), with \(p_T^* =\mcalN(\bm{0},\mbbI)\), \( p_T = \pi\), \(p_0^*=p_d\) and \(p_0 = p_m\), denoting the terminal and initial, data and \textit{modeled} data distribution, respectively. The, we have:
    \begin{align*}
        \mcalD_{KL,\mathrm{DG}^*}(\pd\|p_m) \leq \mcalD_{KL}(p_T^*\|\pi) + \varepsilon_{D^*},
    \end{align*}
    where
    \begin{align}
        \varepsilon_{D^*} &= \frac{1}{2} \mbbE_{p^*_t}\left[\int g^2(t)\Big\|E_{S^*} - h(t) \nabla_{\bmX} D_t^*(\bmX_t)\Big\|^2\rmd t\right] \label{EqnApp_Gain1}\\
        &= \frac{1}{2} \mbbE_{p^*_t}\left[\int g^2(t)\Big\| \nabla D_{SGAN,t}^*(\bmX_t) - \nabla D_{t}^*(\bmX_t)\Big\|^2\rmd t\right], \label{EqnApp_Gain2}
    \end{align}
    where in turn, \(E_{S^*} = \nabla \ln p_t^*(\bmX_t) - {\epsilon}_{\theta}(\bmX_t)\), which is the error present in the standard score-based Langevin sampler, and \(D_{SGAN,t}^*(\bmX_t) = \ln \frac{p_t^*}{p_t}\) is the optimal SGAN discriminator.
\end{lemma}
\begin{proof}
Let the probability densities associated with these two processes be \(p^*_t\) and \(p_t\), with \(p_T^*= \mcalN(\bm{0},\mbbI)\), the standard Gaussian distribution and \(p_0^*=p_d\) and \(p_0 = p_m\), denoting the data distribution and the \textit{modeled} data distribution, respectively. Following the procedure presented by~\citet{DiscGuidance23}, we apply the Girsanov theorem to obtain: 
\begin{align*}
    \mathcal{D}_{KL}(p_d\Vert p_m) & \leq \mathcal{D}_{KL}(p^*_T\Vert \pi) + \frac{1}{2} \mbbE_{p^*_t}\left[\int g^2(t)\Big\|\underbrace{\nabla \ln p_t^*(\bmX_t) - \left({\epsilon}_{\theta}(\bmX_t) + h(t) \nabla_{\bmX} D_t^*(\bmX_t)\right)}_{E_{D^*}}\Big\|^2\rmd t\right].
\end{align*}
Similarly, for the standard score-based sampler (without DG$^*$), we have: 
\begin{align*}
    \rmd \bmX_t &= \left( f(t) + g^2(t) \nabla_{\bmX} \ln p_t^*(\bmX_t) \right)\rmd t + g(t) \rmd\bmW_t,~\text{and}\\
    \rmd \bmY_t &= \left( f(t) + g^2(t){\epsilon}_{\theta}(\bmY_t) \right)\rmd t + g(t)\rmd\bmW_t.
\end{align*}
Applying the Girsanov theorem to the above setting, we get:
\begin{align*}
    \mathcal{D}_{KL}(p_d\Vert p_m) & \leq \mathcal{D}_{KL}(p^*_T\Vert \pi) + \frac{1}{2} \mbbE_{p^*_t}\left[\int g^2(t)\Big\|\underbrace{\ln p_t^*(\bmX_t) - {\epsilon}_{\theta}(\bmX_t)}_{E_{S^*}}\Big\|^2\rmd t\right].
\end{align*}
In order to analyze the gains obtained by introducing the closed-form discriminator guidance, we analyze the behavior of \(E_{D^*} - E_{S}\), and note that, when \(E_{D^*} - E_{S}\) is positive, the proposed discriminator-guided Langevin diffusion improves convergence, as the associated KL-divergence between \(p_d\) and its model \(p_m\), improved (reduced). Consider:
\begin{align}
    E_{D^*} &= \ln p_t^*(\bmX_t) - {\epsilon}_{\theta}(\bmX_t) - h(t) \nabla_{\bmX} D_t^*(\bmX_t) \nonumber \\
    \Rightarrow E_{D^*} &= E_{S^*} - h(t) \nabla_{\bmX} D_t^*(\bmX_t). \label{Eqn_Gain}
\end{align}
As we can see, the gain obtained by the discriminator guidance depends on (a) The sign, and (b) The magnitude of \(\nabla_{\bmX} D_t^*(\bmX_t)\). To quantify this gain, first in the setting considered in Section~\ref{Sec:DiscInfusion}, consider the expression for the discriminator gradient:
\begin{align*}
    \nabla_{\bmX} D_t^*(\bmX_t) &= \mfrakC_{\kappa} \nabla_{\bmX} ((p_{t-1} - p_d) * \kappa)(\bmX_t) \\
    &= \mfrakC_{\kappa} \int_{\bmy} \left(p_{t-1}(\bmy) - p_d(\bmy)\right) \nabla_{\bmX}\kappa(\bmX - \bmy)\bigg|_{\bmX = \bmX_t} \rmd\bmy \\
    &= \mfrakC_{\kappa} \int_{\bmy} \nabla_{\bmX}\left(p_{t-1}(\bmX - \bmy) - \nabla_{\bmX}p_d(\bmX - \bmy)\right)\bigg|_{\bmX = \bmX_t} \kappa(\bmy) \rmd\bmy 
\end{align*}
where \(\mfrakC_{\kappa}\) is a kernel-dependent positive-valued constant. To analyze the above  for \(0\leq t \leq T\), noting that \(p_{0} = p_m \approx p_d \) and \(p_{T} = \mcalN(\bm{0},\mbbI) \), we make the following observations
\begin{itemize}
    \item \textbf{Gradient of \(\bm{\kappa}\)}: The kernel \(kappa\) are derived as solutions to Fokker-Plank equations that govern the optimality of GAN discriminator, and as shown in Table~\ref{Table_KenrelGradients}, are all radially symmetric functions. Consequently the gradients of the kernel are anti-symmetric in nature. 
    \item \textbf{Magnitude of \(\bm{\kappa}\)}: Considering either the popular \(n\)-dimensional Gaussian kernel, or the polyharmonic family of kernels for order \(m\leq \frac{n}{2}\), we observe that the kernels peak at the origin (or alternatively, \(\kappa(\cdot - \bmX_t)\) peaks at \(\bmX_t\)), and decay rapidly.
    \item \textbf{Sign and Magnitude of} \(\left(p_{t-1}(\bmy) - \pd(\bmy) * \kappa \right)\). Given that \(\pd\) is the data distribution, which is known to be drawn from a low-dimensional manifold in a high-dimensional space, and that \(p_{t-1}\) is \textit{closer} to Gaussian noise (or noise-convolved version of \(p_d\)) in early iterations, the density difference (\(p_{t-1} - \pd\)) These results are also in alignment with the observations made by~\citep{SpiderGAN23,SIDuse24}, in the context of the signed Inception distance, which leveraged the kernel-based discriminator to evaluate GANs. 
\end{itemize}
    From the above argument, we see that the gain in KL-divergence, when \(\bmX_t \sim p_t\) far from \(p_d\), the discriminator improves the performance of the standard score-based sampler. \par

    For the special case where \(h(t) = 1\), and adding and subtracting \(\nabla D_{SGAN,t}^*\) to Equation~\ref{Eqn_Gain1}, \(\varepsilon_{D^*}\) can be simplified as:
    \begin{align*}
         &\frac{1}{2} \mbbE_{p^*_t}\left[\int g^2(t)\Big\|\nabla \ln p_t^*(\bmX_t) - {\epsilon}_{\theta}(\bmX_t) - \nabla D_{SGAN,t}^*(\bmX_t) + \nabla D_{SGAN,t}^*(\bmX_t)  -  \nabla_{\bmX} D_t^*(\bmX_t)\Big\|^2\rmd t\right]\\
        &= \frac{1}{2} \mbbE_{p^*_t}\left[\int g^2(t)\Big\|\nabla \ln p_t^*(\bmX_t) - \nabla \ln p_t(\bmX_t)  - \nabla \ln \frac{p_t^*(\bmX_t)}{p_t(\bmX_t)} + \nabla D_{SGAN,t}^*(\bmX_t) -  \nabla_{\bmX} D_t^*(\bmX_t)\Big\|^2\rmd t\right] \\
        &= \frac{1}{2} \mbbE_{p^*_t}\left[\int g^2(t)\Big\|\nabla D_{SGAN,t}^*(\bmX_t) -  \nabla_{\bmX} D_t^*(\bmX_t)\Big\|^2\rmd t\right]
    \end{align*}
\end{proof}
The above result suggests that, in moving from the standard score-based sampler to the closed-form discriminator-guided sampler, the bound on the KL divergence between the true and learnt distributions is transformed from the error in estimating the score, to the error between the optimal SGAN and IPM-GAN discriminators. 

\textit{\textbf{Application of the Lemma to WANDA}}: The result does not make any assumption on $h(t)$, which is the coefficient of the discriminator gradient. In the WANDA setting, wherein the discriminator is *turned off* after $T_D$, setting $h(t) = h_1(t)H_{T_D}(t)$, where $H_{T_D}(t)$ is the Heaviside/unit-step function with the step at $T_D$. Furthermore, to understand the convergence result in WANDA, we can simplify the gain derived in the preceding lemma as follows:
\begin{align*}
\Big\Vert E_{S^*} - h(t) \nabla_{\bmX} D_t^*(\bmX_t)\Big\Vert^2 
&= \Big\Vert E_{S^*} - h(t)\nabla_{\bmX} D_t(\bmX_t) + h(t)\nabla_{\bmX} D_t(\bmX_t) -  h(t) \nabla_{\bmX} D_t^*(\bmX_t)\Big\Vert^2\\
&= \Big\Vert E_{S^*} - h_1(t)H_{T_D}(t)\nabla_{\bmX} D_t(\bmX_t) + h_1(t)H_{T_D}(t)\varepsilon_{\nabla D}\Big\Vert^2
\end{align*}
where $D_t(\bmX_t)$ denotes the sample estimate of the optimal discriminator defined in Equation 8 (L346) of the submission, and $\varepsilon_{\nabla D}$ denotes the error in estimating the true closed-form discriminator via the sample estimate. The discriminator guidance phase can now be defined as a choice of $T_D$ such that the gains obtained by the closed-form discriminator remain positive (\textit{i.e.,} select $T_D$ such that $[ \nabla_{\bmX} D_t(\bmX_t) - \varepsilon_{\nabla D}]_i > 0~\forall~i$ (element-wise inequality)). However, computing $T_D$ in closed-form via this approach is impractical as we do not have access to the form or characteristics of $p_d$ or $p_{t-1}$ in practice. As discussed in the ablations, this value was found empirically to be around 10\% of the total number of iterations, $T$.

However, we remark that this analysis in not entirely aligned with the derived optimal discriminator, as DG$^*$ is optimal in the sense of the Wasserstein-2 metric, and the convergence of score-based diffusion is in the \(f\)-divergence sense, and in particular, the KL divergence. A more in-depth analysis of the proposed SDE, in terms of the Wasserstein metric, is a promising direction for future research. 

\subsection{Accelerated Convergence of the WANDA Framework}\label{App_WANDAConv}

To build intuition, we show that the proposed guidance framework can be viewed as effectively resulting in a second-order update scheme, owing to the form of the discriminator kernel graident. The second-order update resembles Polyak heavy-ball momentum update found in the literature~\citep{StatNotes18,RechtBook22,FastDiff23}, and can be attributed to being the source for the observed acceleration. Two key contributing factor in this analysis are (a) The explicit dependence of the discriminator gradient at time $t$, on the generated distribution at time $t-1$ (appearing in the form of the convolution with $p_{t-1}$); and (b) the radial symmetry of the kernel ($\kappa(\Vert \x\Vert)$), which always yields a gradient of the form $\mathfrak{c} \x \kappa^{\prime}(\Vert\x\Vert)$. In particular, consider a setting wherein the kernel is a polyharmonic spline kernel of order $k=1$ (cf. Table 3):
$$
\nabla D_t(\bmX_t) =   \mathfrak{C}^{\prime}_k\sum_{\bmg^j \sim \{\bmX_{t-1}\}} \nabla_{\bmX}\kappa(\bmX_t-\bmg^j)- \mathfrak{C}^{\prime}_k\sum_{\bmd^i \sim p_d} \nabla_{\bmX}\kappa(\bmX_t-\bmd^i).
$$

A simplified single-sample approximation gives

$$
\nabla D_t(\bmX_t) =  \mathfrak{C}^{2}_k \frac{\bmX_{t} - \bmX_{t-1}}{\Vert \bmX_{t} - \bmX_{t-1} \Vert} - \mathfrak{C}^{2}_k  \frac{\bmX_{t} - \bmd}{\Vert \bmX_{t} - \bmd \Vert}, 
$$
where $\bmd$ is a random sample drawn from the target data distribution. Consider the standard closed-form discriminator guided Diffusion update:
$$
\bmX_{t+1} = \alpha_{1,t} \bmX_t - \alpha_{2,t} \epsilon_{\theta}(\bmX_t) - \alpha_{3,t} \nabla D_t(\bmX_t) + \alpha_{4,t} \mathbf{Z}_t
$$

Substituting in for the above discriminator gradient and simplifying results in an update of the form:
$$
\bmX_{t+1} = \beta_{1,t} \bmX_t - \alpha_{2,t} \epsilon_{\theta}(\bmX_t) - \beta_{3,t} \bmX_{t-1} + \alpha_{4,t} \mathbf{Z}_t + \bm{\beta}_{5,t},
$$
where $\beta_{1,t} = \alpha_{1,t} - \frac{\alpha_{3,t}\mathfrak{C}^{2}_k}{\Vert \bmX_{t} - \bmX_{t-1} \Vert} + \frac{\alpha_{3,t}\mathfrak{C}^{2}_k}{\Vert \bmX_{t} - \bmd \Vert}$ and $\beta_{3,t} = \frac{\alpha_{3,t}\mathfrak{C}^{2}_k}{\Vert \bmX_{t} - \bmX_{t-1} \Vert}$. The above equation defines a second-order update, which resembles the update schemes encountered in momentum-based diffusion models~\citep{FastDiff23} --- we hypothesize that this is one of the sources of \textit{acceleration} in the proposed technique.

\subsection{Convergence Analysis of Discriminator Guidance}
The baseline analysis follows the analysis presented in (\url{https://fa.bianp.net/blog/2023/ulaq/}), which covers the unadjusted Langevin algorithm. Consider the baseline setting:
\[
p_d(\mathbf{x}) = \frac{1}{Z} \exp\{-f(\mathbf{x})\}~\text{where}~Z=\int_{\mathbb{R}^d} \exp\{-f(\mathbf{x})\}\rmd \mathbf{x}.
\]
where \(f:\mbbR^d\rightarrow\mbbR\) with access to \(\nabla f(\x)\), \textit{i.e.,} in our setting, $f$ is the log-probability of the data. The unadjusted Langevin algorithm (ULA) is:
\begin{align}
\mathbf{x}_{t+1} = \mathbf{x}_t - \gamma \nabla f(\x_t) + \sqrt{2\gamma} \epsilon_t\,\,\text{where}\,\, \epsilon_t \sim \mathcal{N}(\mathbf{0},\mathbb{I}) \label{eqn_BaseIteration}
\end{align}
Convergence is measured in the distribution sense between the desired target distribution \(p_d\) and the iterate distribution \(p_t\) in terms of the Wasserstein 2 metric, \textit{i.e.,}
\begin{align*}
\mathcal{W}_2^2(p_q,p_t) = \inf_{\pi \in \Pi(p_d,p_t)} \mbbE_{(\mathbf{x},\mathbf{y})\sim\pi} \left[ \|\x-\y \|_2^2\right].
\end{align*}
 \textbf{Assume} that \(p_d \sim \mathcal{N}(\mu,H^{-1})\). Then, if both \(p_d\) and \(p_t\) are Gaussians, with commuting covariances, we have
 \begin{align*}
\mathcal{W}_2^2(p_d,p_t) &= \| \mu - \mu_t\|_2^2 + \mathrm{Tr}\left( H^{-1} + \Sigma_t - 2\sqrt{H^{-1}\Sigma_t}\right) \\
&= \| \mu - \mu_t\|_2^2 + \|H^{-\frac{1}{2}} - \Sigma^{\frac12}_t\|_F^2,
\end{align*}
To analyze the proposed setting, first, considering the Gaussian (or locally Gaussian) model on the data, we have 
\begin{align}
f(\mathbf{x}) = \frac12 (\mathbf{x}-\mu)^{\mathrm{T}}H(\mathbf{x}- \mu)\,\,\Rightarrow\,\, \nabla f(\mathbf{x}) = H(\mathbf{x} - \mu).
\end{align}
The discriminator guidance can be introduced into the model as follows: 
\begin{align*}
 \mathbf{x}_{t+1} &=\mathbf{x}_t-\gamma\nabla f(\mathbf{x}_t) - \alpha_3 \nabla D_t^*(\mathbf{x}_t)+\sqrt{2\gamma}\,\epsilon_t \quad \epsilon_t\sim\mathcal{N} \\
 \mathbf{x}_{t+1} &=\mathbf{x}_t-\gamma\nabla f(\mathbf{x}_t) + \beta \left( \mathbf{x}_t - \mathbf{x}_{t-1} \right) - \eta \left( \mathbf{x}_t - \mathbf{d} \right) + \sqrt{2\gamma}\,\epsilon_t \quad \epsilon_t\sim\mathcal{N},
 \end{align*}
 where we expand the discriminator about a single real centre \(\mathbf{d}\) and a single fake centre, which is the sample from the previous iteration \(\mathbf{x}_{t-1}\). For simplicity, we \textbf{Assume that} the coefficients \(\beta\) and \(\eta\) are constant. (In practice, \(\beta_t = \frac{C_{k}}{\|\mathbf{x}_{t} - \mathbf{x}_{t-1}\|}\) and \(\eta_t = \frac{C_{k}}{\|\mathbf{x}_{t} - \mathbf{d}\|}\). This modified update scene can be analyzed under the standard second-order dynamics setting, if not for the \( \left( \mathbf{x}_t - \mathbf{d} \right)\) term. To account for this, we must reformulate the function \(f\) as follows. Let \(\lambda = \frac{\eta}{\gamma}\). Then, let
 \begin{align*}
 \tilde{f}(\mathbf{x}) = f(\mathbf{x}) + \frac{\lambda}{2}\|\mathbf{x} - \mathbf{d}\|_2^2.
 \end{align*}
Then we have the following modified definitions:
\begin{align*}
\tilde{H} &= H + \lambda\mathbb{I} \\
\tilde{\mathbf{x}}^* &= \arg\min\tilde{f(\mathbf{x}}) = \left(H + \lambda\mathbb{I}\right)^{-1}\left(H\mu + \lambda \mathbf{d}\right) = \tilde{H}^{-1}\left(H\mu + \lambda \mathbf{d}\right)~\text{and}, \\
\nabla \tilde{f} &= \nabla f(\mathbf{x}) + \lambda(\mathbf{x} - \mathbf{d}) \\
&= H(\mathbf{x} - \mu) + \lambda(\mathbf{x} - \mathbf{d}) \\
&= (H + \lambda\mathbb{I})\mathbf{x} - ( H\mu + \lambda \mathbf{d}) \\
&= \tilde{H}\mathbf{x} - \tilde{H} \tilde{H}^{-1}( H\mu + \lambda \mathbf{d})\\
&=\tilde{H}(\mathbf{x} - \tilde{\mathbf{x}}^* ).
\end{align*}
 Let \(\rho(H) \in [\ell^\prime,L^\prime]\) be the bounds on the singular values of H. We can analyze the shifted system \(\mathbf{y}_t = \mathbf{x}_t -  \tilde{\mathbf{x}}^* \). Then, the iterates become: 
\begin{align*}
\mathbf{y}_{t+1}=\underbrace{\big((1+\beta)\mathbb{I}-\gamma\tilde{H}\big)}_{A}\mathbf{y}_t- \beta \mathbf{y}_{t-1}+\sqrt{2\gamma}\,\epsilon_t.
\end{align*}
Rearranging to form the state \(\mathbf{s}_t=\begin{bmatrix}\mathbf{y}_t\\ \mathbf{y}_{t-1}\end{bmatrix}\), with \(\zeta_t = \begin{bmatrix} \epsilon_t \\ 0_{n \times 1}\end{bmatrix}\), we have the update equation:
$$
\mathbf{s}_{t+1}=M_\gamma \mathbf{s}_t+\sqrt{2\eta}\,B\zeta_t,\quad
M_{\gamma,\beta}=
\begin{bmatrix}A&-\beta I\\ I&0\end{bmatrix},\;
B=\begin{bmatrix}\mathbb{I}\\0_{n\times n}\end{bmatrix}.
$$
For a given initial condition $\mathbf{s}_0$, The mean and covariance are given by: 
\begin{align*}
\mu_t^s = M_{\gamma,\beta}^t \mathbf{s}_0~\text{and}~\Sigma_{t}^s = M_{\gamma,\beta}\Sigma_{t-1}^sM_{\gamma,\beta}^{\mathrm{T}} + 2\gamma BB^{\mathrm{T}},
\end{align*}
where \(\rho(M_{\gamma,\beta}) \in [\ell,L]\). We can now analyze the stability of this system by leveraging results from the optimization literature. The optimal Polyak step size is given by:
\begin{align*}
\gamma^* = \frac{4}{\sqrt{L} + \sqrt{\ell}}~~\text{and}~~\beta^* = \left(\frac{\sqrt{\kappa} - 1}{\sqrt{\kappa} + 1}\right)^2,~\text{where}~\kappa = \frac{L}{\ell},
\end{align*}
which gives us the rate \(\frac{\sqrt{\kappa} - 1}{\sqrt{\kappa} + 1}\), which is \(1-\mathcal{O}\left(\frac{1}{\sqrt{\kappa}}\right)\). We can also extend this analysis to derive a bound on the Wasserstein distance, which gives us:
\begin{align*}
\mathcal{W}_2^2(p_d,p_t) & \leq \rho(M_{\gamma})^{2t} \mathcal{W}_2^2(p_0,p_t) + C_{\text{bias}}(\gamma, H^{-1}) \\
& \leq \rho(M_{\gamma})^{2t} \left( \|\mathbf{s}_0\| + C_{\Sigma_0} \right) + \mathcal{O}(\gamma + \|H\|) \\
& \leq \rho(M_{\gamma})^{2t} + \mathcal{O}(\gamma).
\end{align*}
\textbf{Insights:} Besides the Polyak Heavy-ball equivalence, we see the following additional insights. First, we have a bound on \(\beta\) in order to be able to achieve the desired acceleration. However, in practice, \(\beta_t\) grows as iterations progress \(\left(\beta_t = \frac{C_{k}}{\|\mathbf{x}_t - \mathbf{x}_{t-1}\|}\right)\), which can be viewed as the reason why the discriminator guidance is beneficial only in the initial iterations, where \(\beta\) is sufficiently small. Second, we observe that, in addition to the acceleration introduced by the ``push'' term, the ``pull'' terms in the discriminator add a regularization to the score function, centered about the true samples. 
\newpage

\section{Additional Experimental Results on Discriminator-guided Langevin Sampling} \label{App_DiscInfusion}
We present additional experimental results on generating 2-D shapes, and images using the discriminator-guided Langevin sampler. 
\subsection{Additional Results on Synthetic Data Learning} \label{App_InfusionGaussians}

On the 2-D learning task, we present additional combinations on the {\it shape morphing experiment}. \par
\begin{figure*}[!b]
  \begin{center}
    \begin{tabular}[b]{P{.17\linewidth}P{.17\linewidth}P{.17\linewidth}P{.17\linewidth}P{.17\linewidth}}
\includegraphics[width=1.15\linewidth]{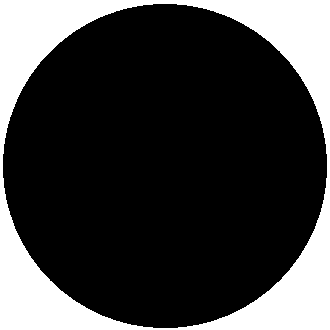} & 
\includegraphics[width=1.15\linewidth]{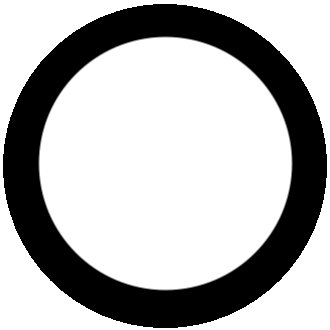} & 
\includegraphics[width=1.15\linewidth]{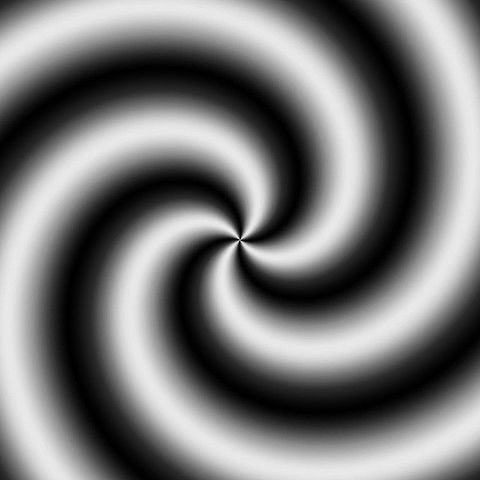} & 
\includegraphics[width=1.15\linewidth]{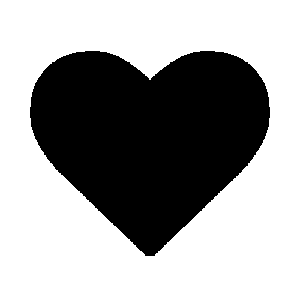} & 
\includegraphics[width=1.15\linewidth]{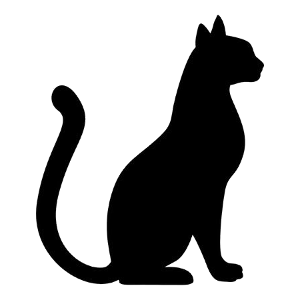}  \\[1pt] 
    \end{tabular} 
  \caption[]{(\includegraphics[height=0.009\textheight]{Rgb.png} Color online)  Images considered in generating the source and target in the {\it Shape morphing} experiment.} 
    \label{Plot_SrcMorphing}
    \end{center}
    \vskip-1.25em
  \end{figure*}

{\it {\bfseries Training Parameters}}: All samplers are implemented using TensorFlow~\citep{TF} library. The discriminator gradient is built as a custom radial basis function network, whose weights and centers are assigned at each iteration. At \(t=0\), the centers \(\bm{g}^j\sim p_{t-1}\) are sampled from the unit Gaussian, {\it i.e.,} \(p_{-1} = \mcalN(\bm0,\mbbI)\). In subsequent iterations, the batch of samples from time instant \(t-1\) serve as the centers for \(D_t^*\). Based on experiments presented in Appendix~\ref{App_InfusionImages}, we set \(\gamma_t = 0\) and \(\alpha_t = 1~\forall~t\). The input and target distributions are created following the approach presented by~\citep{USobolevDescent20}. Figure~\ref{Plot_SrcMorphing} shows the supports of the input/output distributions (black denotes the support). For grayscale images, the support corresponds to regions with pixel intensities below the threshold of 128.\par
\begin{figure*}[t!]
  \begin{center}
    \begin{tabular}[b]{P{.01\linewidth}|P{.12\linewidth}P{.12\linewidth}P{.12\linewidth}P{.12\linewidth}P{.12\linewidth}P{.12\linewidth}l}
      \rotatebox{90}{ \quad Circular} &
      \includegraphics[width=1.15\linewidth]{G2_Diffusion/Disk_1.pdf} & 
      \includegraphics[width=1.15\linewidth]{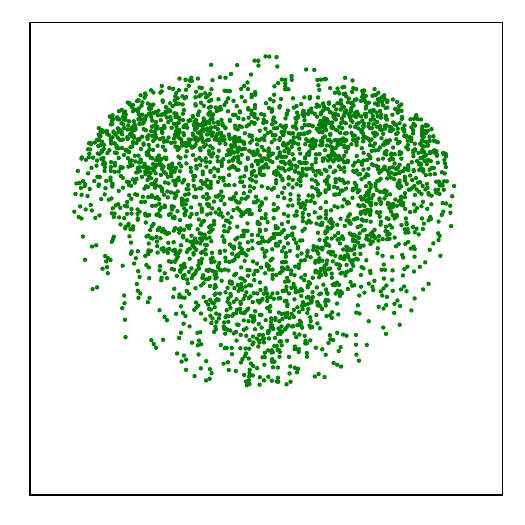} & 
       \includegraphics[width=1.15\linewidth]{G2_Diffusion/Disk_50.pdf} & 
       \includegraphics[width=1.15\linewidth]{G2_Diffusion/Disk_100.pdf} & 
      \includegraphics[width=1.15\linewidth]{G2_Diffusion/Disk_250.pdf} &
      \includegraphics[width=1.15\linewidth]{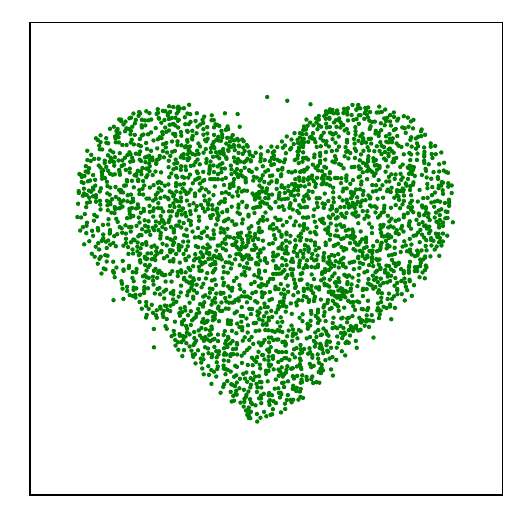} & \multirow{5}*{
      {\rotatebox{90}{Heart\quad\qquad\qquad\quad}}} \\[1pt] 

      \rotatebox{90}{\enskip \quad Ring} &
      \includegraphics[width=1.15\linewidth]{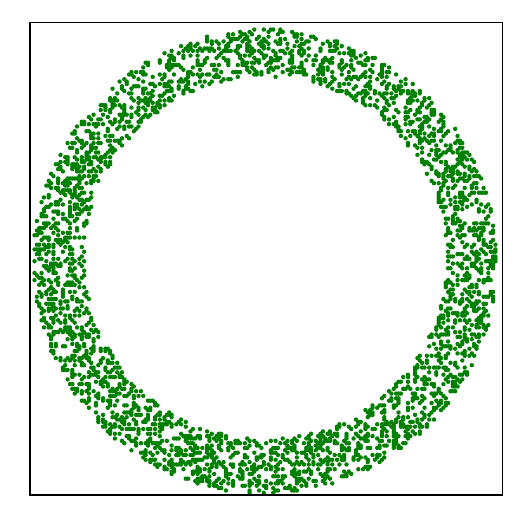} & 
      \includegraphics[width=1.15\linewidth]{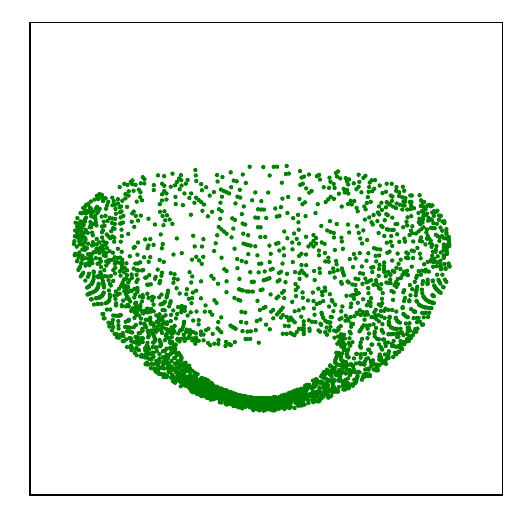} & 
       \includegraphics[width=1.15\linewidth]{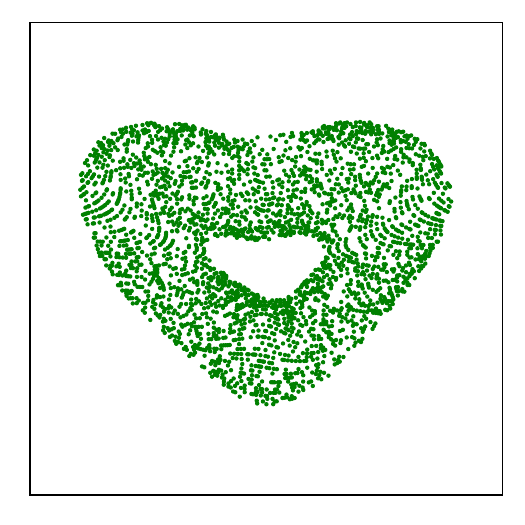} & 
       \includegraphics[width=1.15\linewidth]{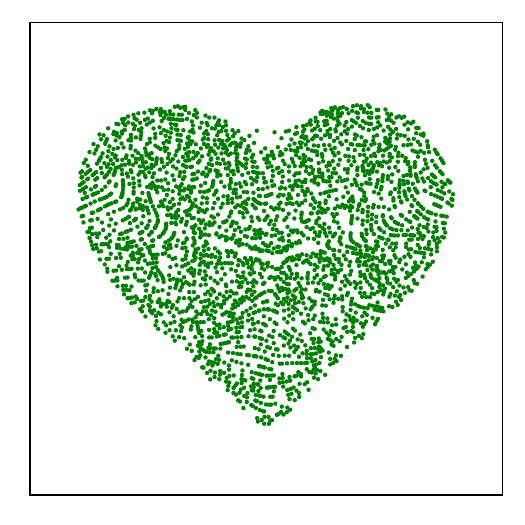} & 
      \includegraphics[width=1.15\linewidth]{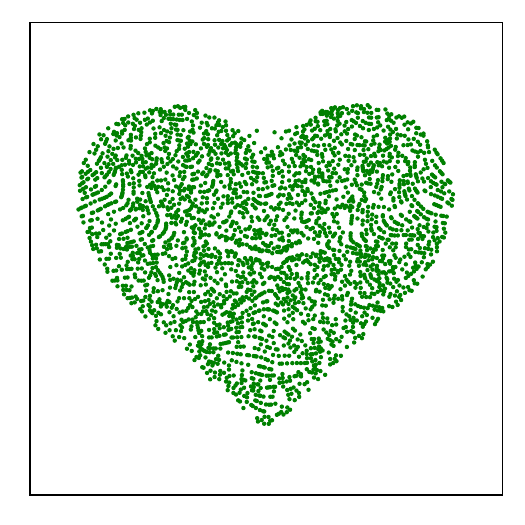} &
      \includegraphics[width=1.15\linewidth]{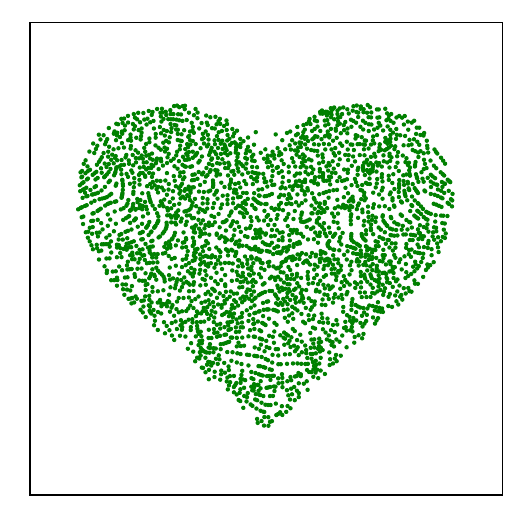} & \\[1pt] 

       \rotatebox{90}{\enskip \quad Spiral} &
       \includegraphics[width=1.15\linewidth]{G2_Diffusion/Spiral_1.pdf} & 
       \includegraphics[width=1.15\linewidth]{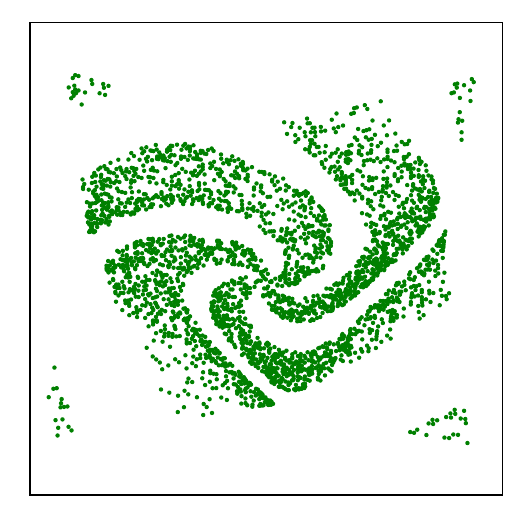} & 
        \includegraphics[width=1.15\linewidth]{G2_Diffusion/Spiral_50.pdf} & 
        \includegraphics[width=1.15\linewidth]{G2_Diffusion/Spiral_100.pdf} & 
       \includegraphics[width=1.15\linewidth]{G2_Diffusion/Spiral_250.pdf} &
       \includegraphics[width=1.15\linewidth]{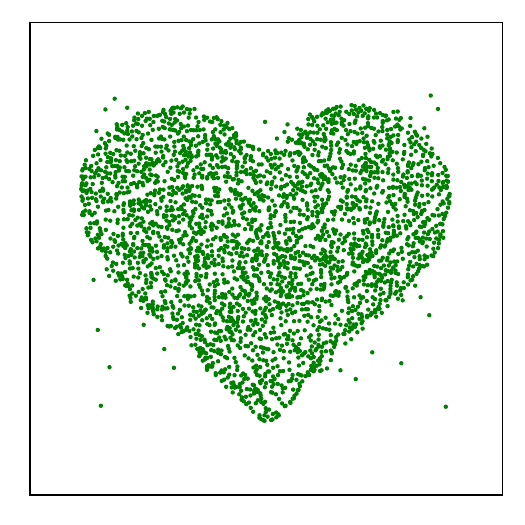} & \\[1pt]

      \rotatebox{90}{\enskip \qquad Cat} &
      \includegraphics[width=1.15\linewidth]{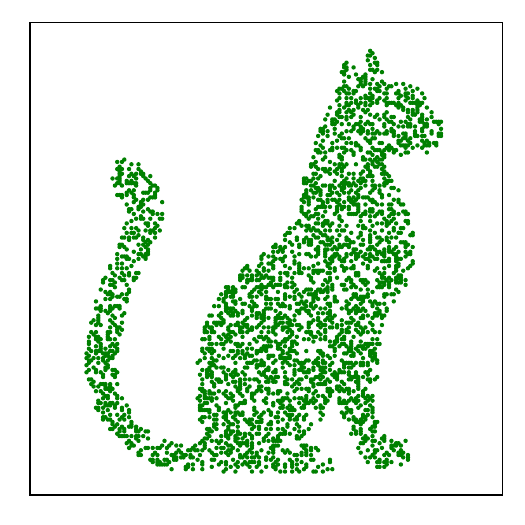} & 
      \includegraphics[width=1.15\linewidth]{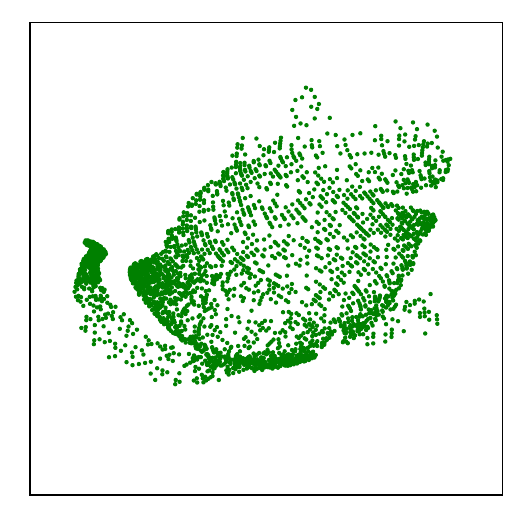} & 
       \includegraphics[width=1.15\linewidth]{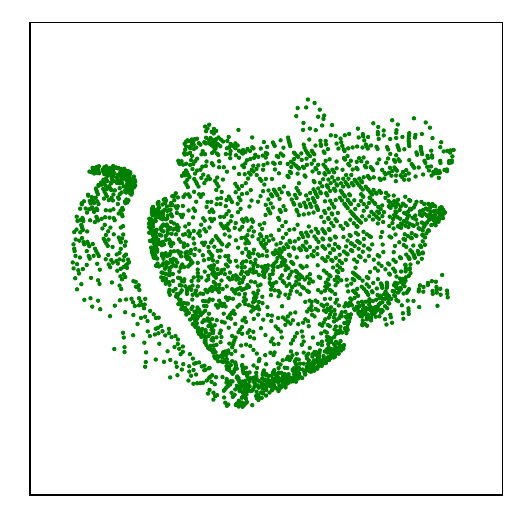} & 
       \includegraphics[width=1.15\linewidth]{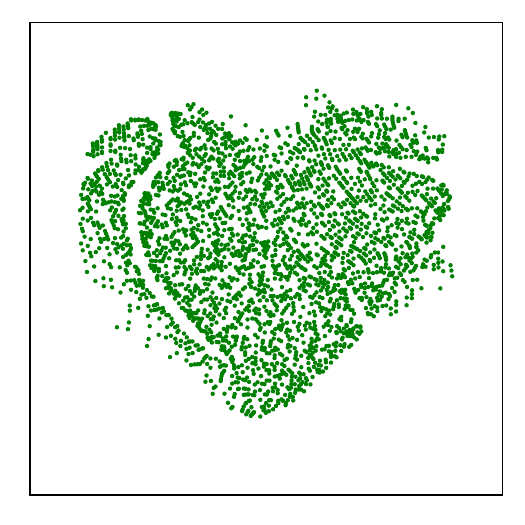} & 
      \includegraphics[width=1.15\linewidth]{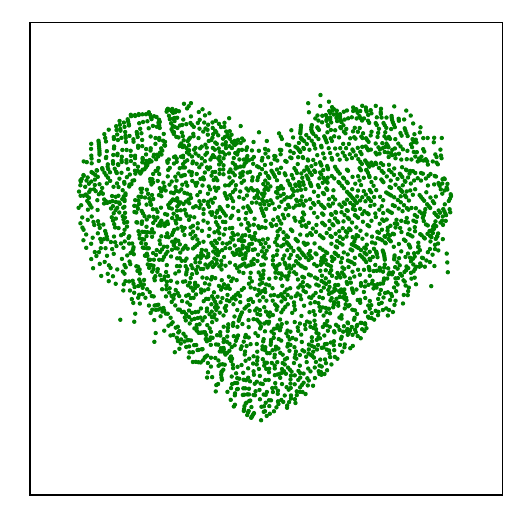} &
      \includegraphics[width=1.15\linewidth]{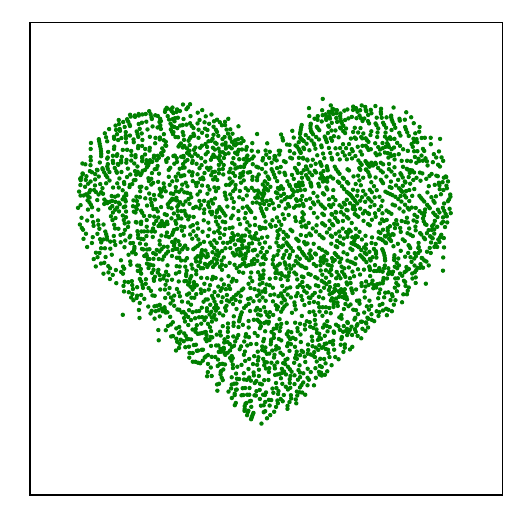} & \\[1pt] 
       
  \midrule
  \rotatebox{90}{ \quad Circular} &
  \includegraphics[width=1.15\linewidth]{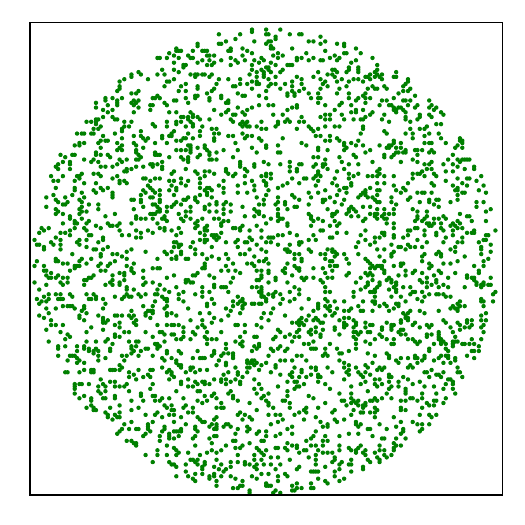} & 
  \includegraphics[width=1.15\linewidth]{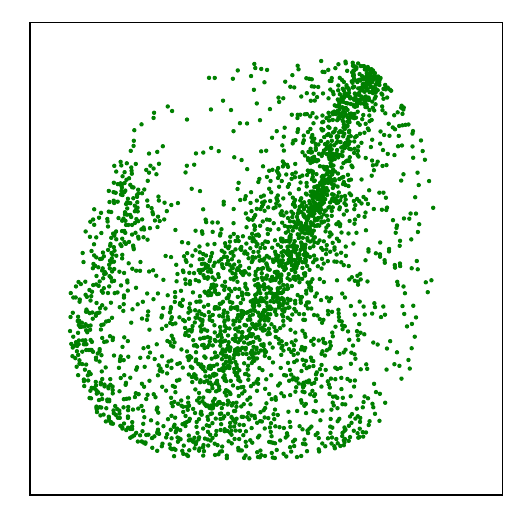} & 
   \includegraphics[width=1.15\linewidth]{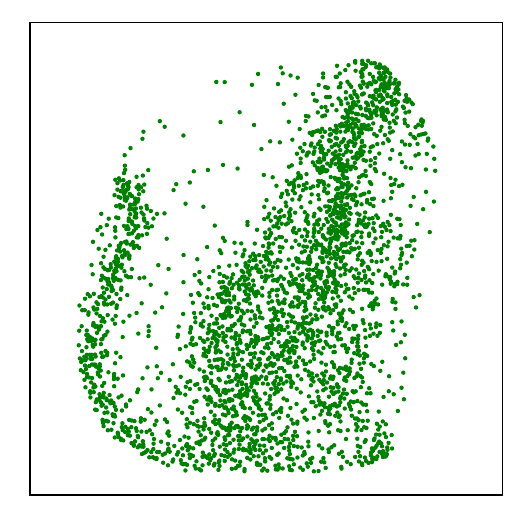} & 
   \includegraphics[width=1.15\linewidth]{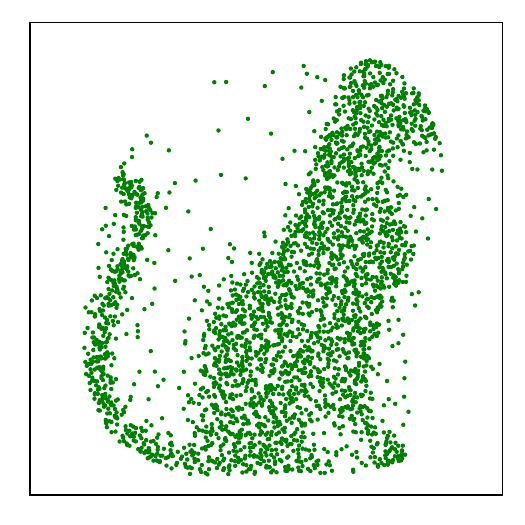} & 
  \includegraphics[width=1.15\linewidth]{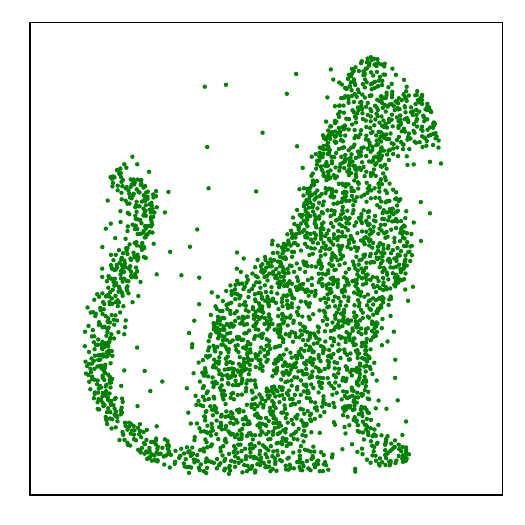} &
  \includegraphics[width=1.15\linewidth]{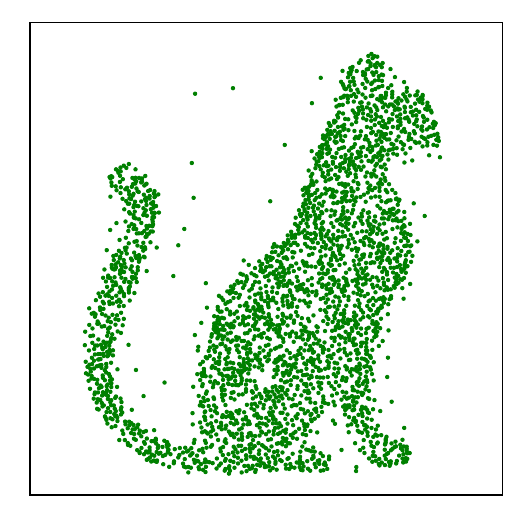} & \multirow{5}*{
  {\rotatebox{90}{Cat\quad\qquad\qquad\quad\enskip}}} \\[1pt] 

  \rotatebox{90}{\enskip \quad Ring} &
  \includegraphics[width=1.15\linewidth]{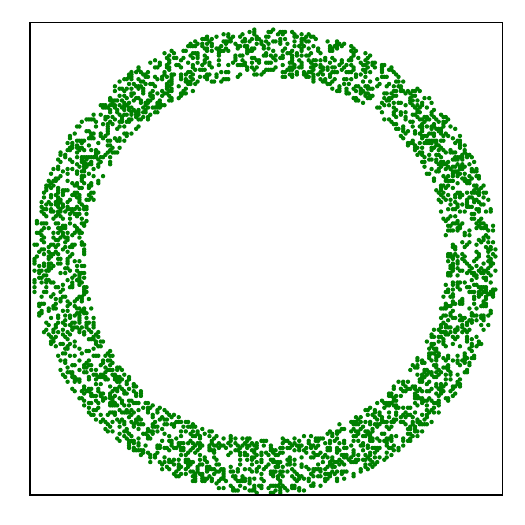} & 
  \includegraphics[width=1.15\linewidth]{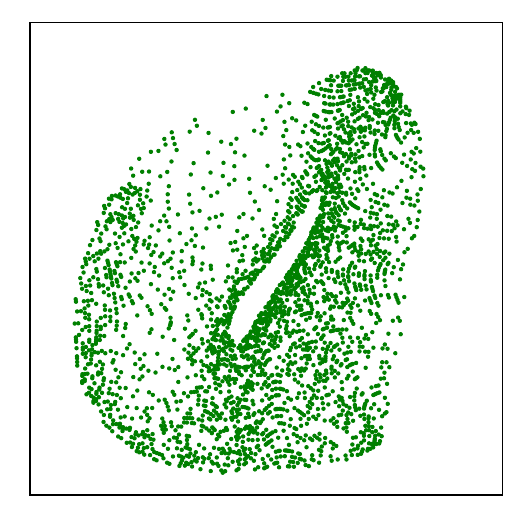} & 
   \includegraphics[width=1.15\linewidth]{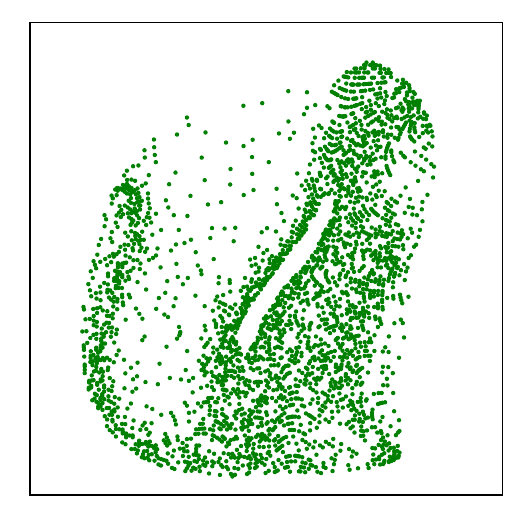} & 
   \includegraphics[width=1.15\linewidth]{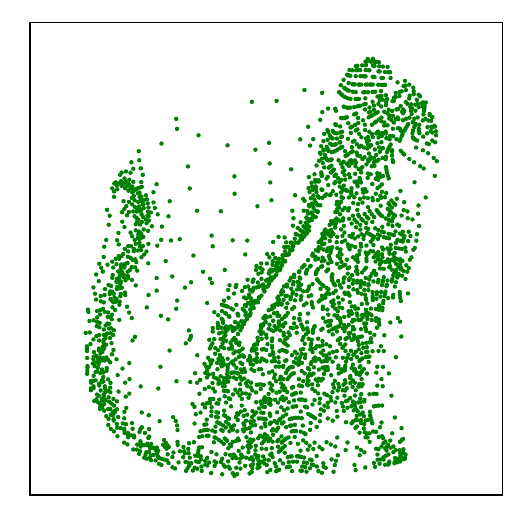} & 
  \includegraphics[width=1.15\linewidth]{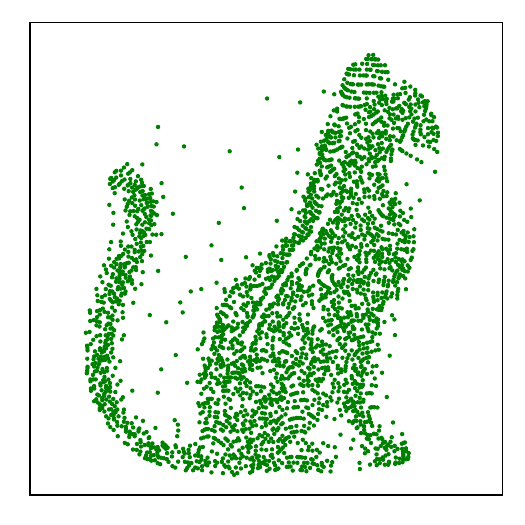} &
  \includegraphics[width=1.15\linewidth]{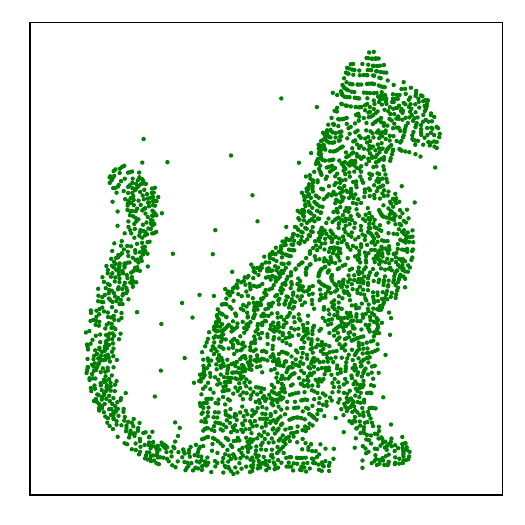} & \\[1pt] 

   \rotatebox{90}{\enskip \quad Spiral} &
   \includegraphics[width=1.15\linewidth]{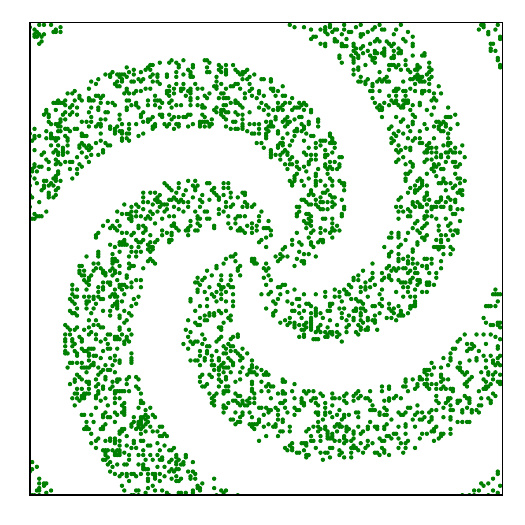} & 
   \includegraphics[width=1.15\linewidth]{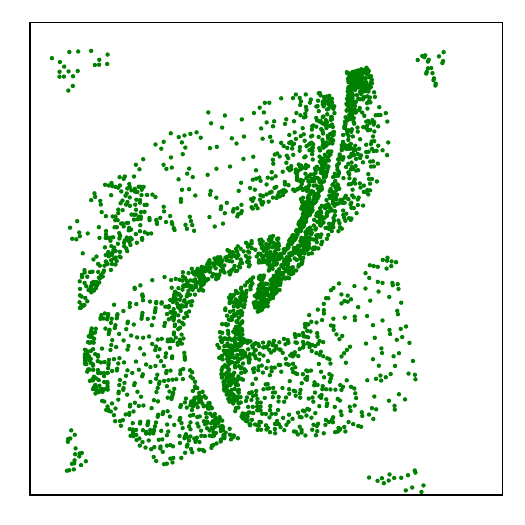} & 
    \includegraphics[width=1.15\linewidth]{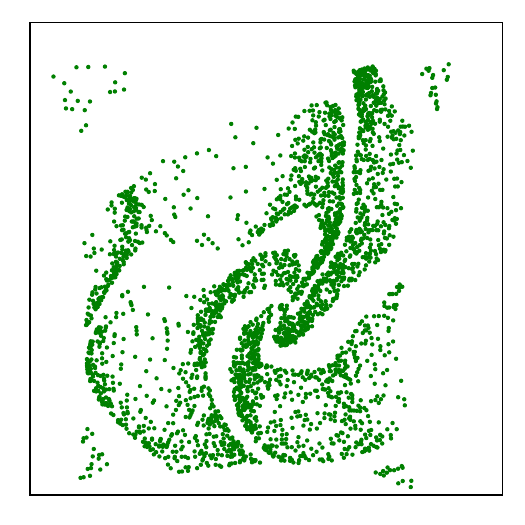} & 
    \includegraphics[width=1.15\linewidth]{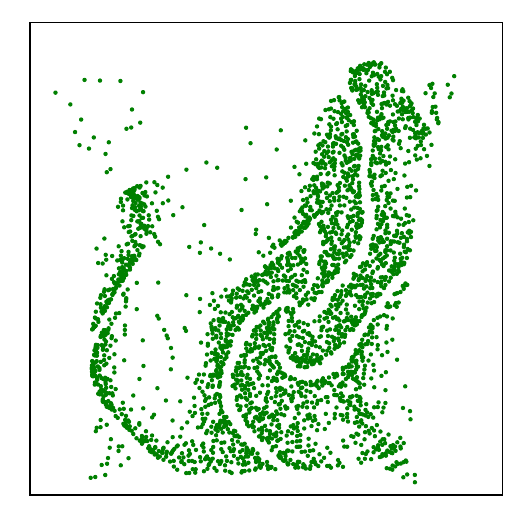} & 
   \includegraphics[width=1.15\linewidth]{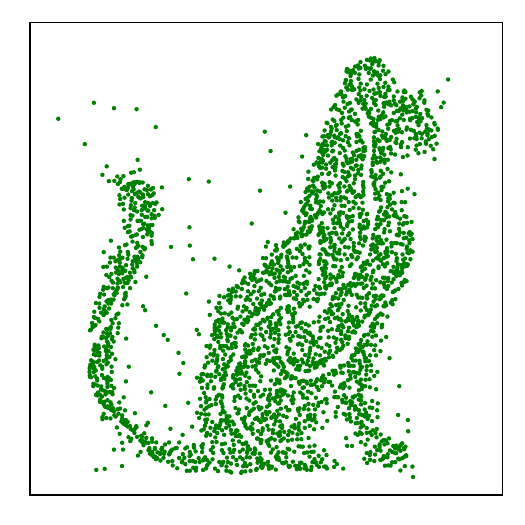} &
   \includegraphics[width=1.15\linewidth]{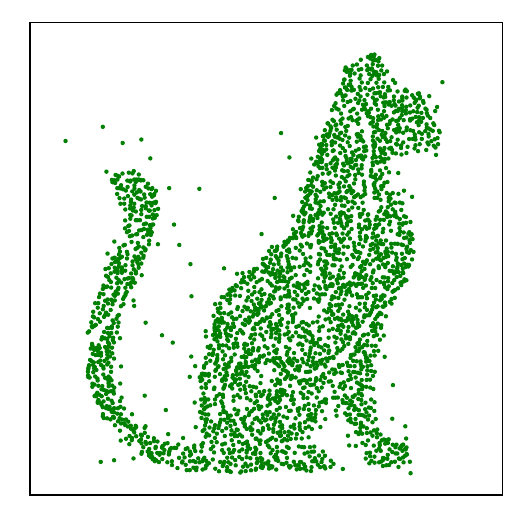} & \\[1pt]

  \rotatebox{90}{ \enskip \quad Heart} &
  \includegraphics[width=1.15\linewidth]{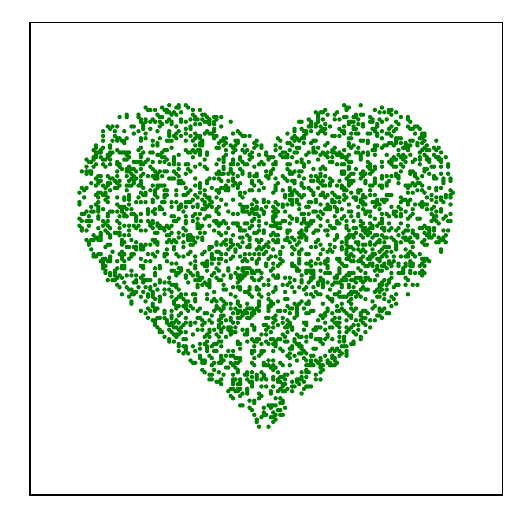} & 
  \includegraphics[width=1.15\linewidth]{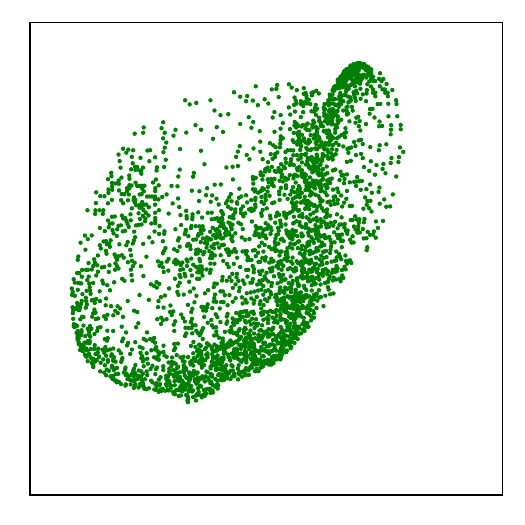} & 
   \includegraphics[width=1.15\linewidth]{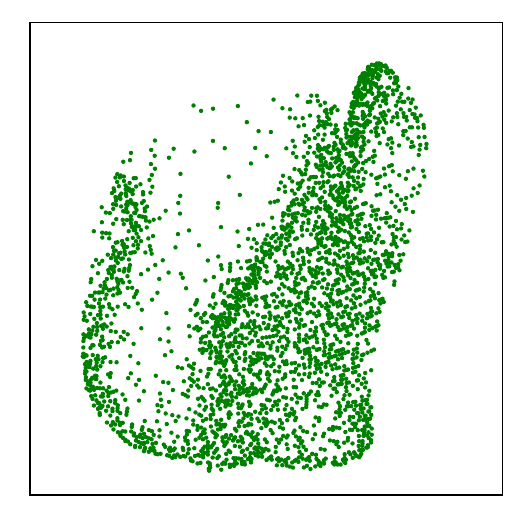} & 
   \includegraphics[width=1.15\linewidth]{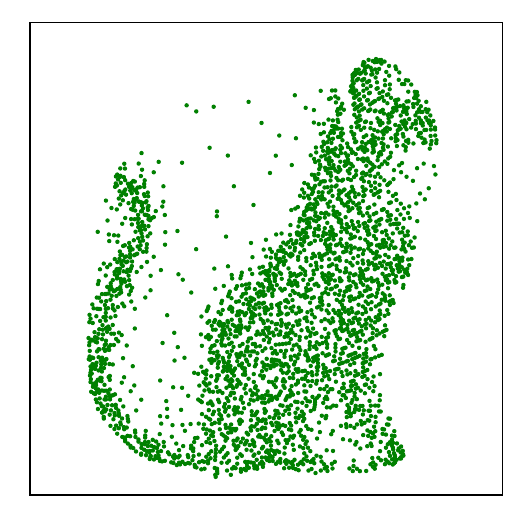} & 
  \includegraphics[width=1.15\linewidth]{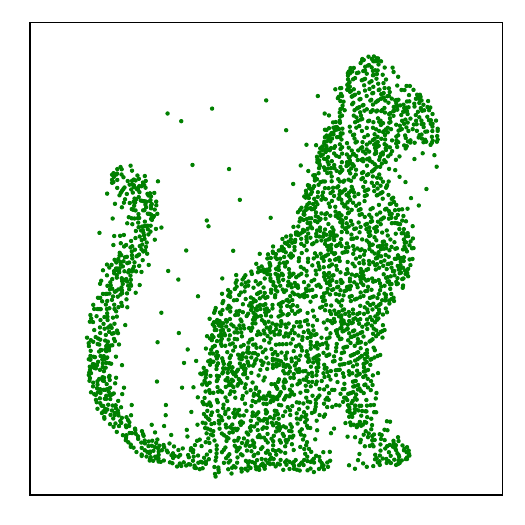} &
  \includegraphics[width=1.15\linewidth]{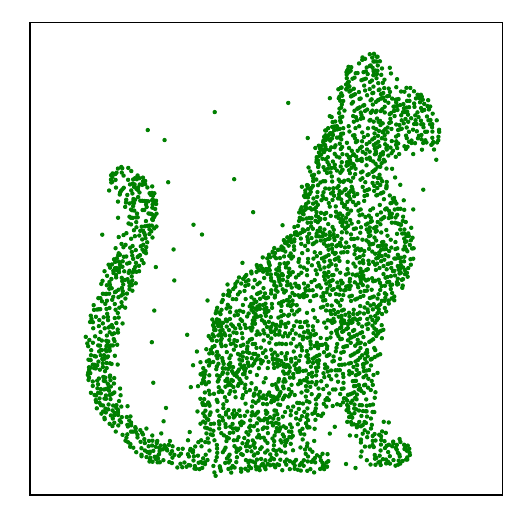} & \\[1pt]


      & \scriptsize{1 iteration} & \scriptsize{20 iterations} &\scriptsize{50 iterations} & \scriptsize{100 iterations} & \scriptsize{250 iterations} &\scriptsize{500 iterations} & \\[1pt] 

    \end{tabular} 
  \caption[]{(\includegraphics[height=0.009\textheight]{Rgb.png} Color online) Samples evolving with iterations for the discriminator-guided Langevin sampler, considering various shapes of the initial uniform distributions, given a target uniform distribution shaped like a {\it Heart}, or a {\it Cat} as indicated. For relatively simpler input shapes, such as the circular pattern, the sampler converges in about 100 iterations, while in the  spiral case, the sampler converges in about 250 steps.} 
    \label{PlotApp_Morphing}
    \end{center}
    \vskip-1.25em
  \end{figure*}

{\it {\bfseries Experimental Results}}: We consider the {\it Heart} and {\it Cat} shapes as the target, while considering various input shapes, corresponding to varying levels  of difficulty in matching the target distribution. In the case of learning the {\it Heart} shape, for input shapes that do not contain {\it gaps/holes}, the convergence is relatively fast, and shape matching occurs in about 100 to 250 iterations. For more challenging input shapes, such as the {\it Cat} logo, the discriminator-guided Langevin sampler converges in about 500 iterations. This is superior to the reported 800 iterations in the Unbalanced Sobolev descent formulation. The results are similar in the case where the {\it Cat} image is the target (cf. Figure~\ref{PlotApp_Morphing}).

\subsection{Additional Results on Image Learning} \label{App_InfusionImages}
We present ablation experiments on generating images with the discriminator-guided Langevin sampler to determine the choice of \(\alpha_t\) and \(\gamma_t\) in the update regime. We also provide additional images pertaining to the experiments presented in the {\it Main Manuscript}.

{\it {\bfseries Choice of coefficients \(\alpha_t\) and \(\gamma_t\)}}: For the ablation experiments, we consider MNIST, SVHN, and 64-dimensional CelebA images. Based on the analysis presented in~\citet{PolyGAN23}, we consider the kernel-based discriminator with the polyharmonic spline kernel in all subsequent experiments. Recall the update scheme:
\begin{align*}
\x_{t} = \x_{t-1} - \alpha_t \nabla_{\x} D_t^*(\x_t;\,p_{t-1},\pd) + \gamma_t \z_t,\quad\text{where}\quad \z_t\sim\mcalN(\bm0,\mbbI).
\end{align*} 
Based on the observations made by~\citet{EDM22}, to ascertain the optimal choice of the coefficients, we consider the following scenarios:
\begin{itemize}
  \item {\bf The ordinary differential equation (ODE) formulation}, wherein the noise perturbations are ignored, giving rise to an ODE that the samples are evolved through. Here \(\gamma_t = 0,~\forall~t\).
  \item {\bf The stochastic differential equation (SDE) formulation}, wherein we retain the noise perturbations. Based on the links between score-based approaches and the GANs, we consider the approach presented in noise-conditioned score networks (NCSNv1)~\citep{NCSN19}, with \(\gamma_t = \sqrt{2\alpha_t}\). 
\end{itemize}
Within these two scenarios, we further consider the following cases:
\begin{itemize}
  \item {\bf Unadjusted Langevin dynamics (ULD)}, wherein \(\alpha_t\) is fixed, {\it i.e.,} \(\alpha_t = \alpha_0,~\forall~t\).
  \item {\bf Annealed Langevin dynamics (ALD)}, wherein \(\alpha_t\) decays according to a schedule. While various approaches have been proposed for scaling~\citep{NCSN19, NCSNv220, NCSNPP21, GGF21,EDM22}, we consider the geometric decay considered in NCSNv1~\citep{NCSN19}. 
\end{itemize}
For either case, we present results considering \(\alpha_0\in\{100,10,1\}\). 
\par

\begin{figure*}[!t]
  \begin{center}
    \begin{tabular}[b]{P{.02\linewidth}|P{.02\linewidth}|P{.4\linewidth}P{.4\linewidth}}
      && Constant \(\alpha_t = \alpha_0~\forall~t\)  & Geometrically decaying \(\alpha_t\)  \\[1pt]
      \toprule
      \multirow{6}*{\rotatebox{90}{ \(\gamma_t = 0;~\forall~t\)\qquad\enskip}} & \rotatebox{90}{ \enskip\quad\(\alpha_0 = 1\)} &
      \includegraphics[width=1.0\linewidth]{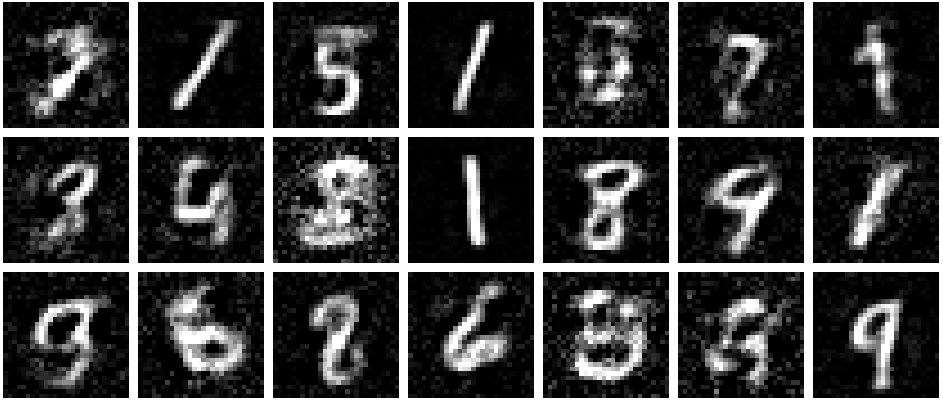} & 
     \includegraphics[width=1.0\linewidth]{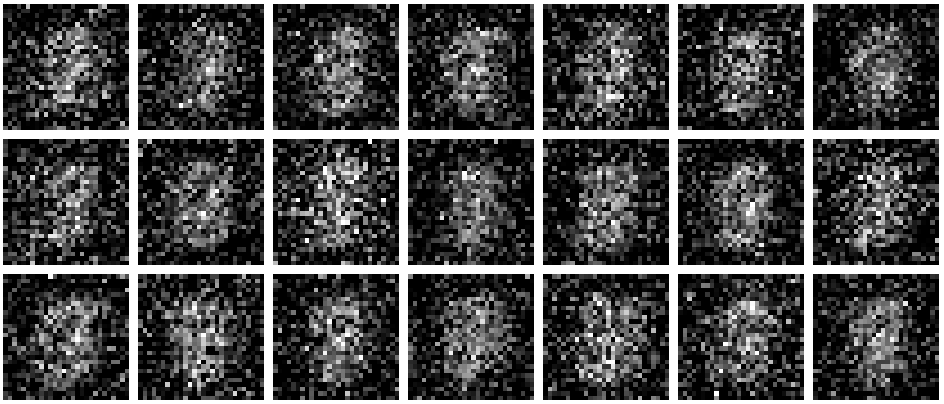}  \\[-1pt] \cmidrule{2-4}
      &
      \rotatebox{90}{ \quad\(\alpha_0 = 10\)} &\includegraphics[width=1.0\linewidth]{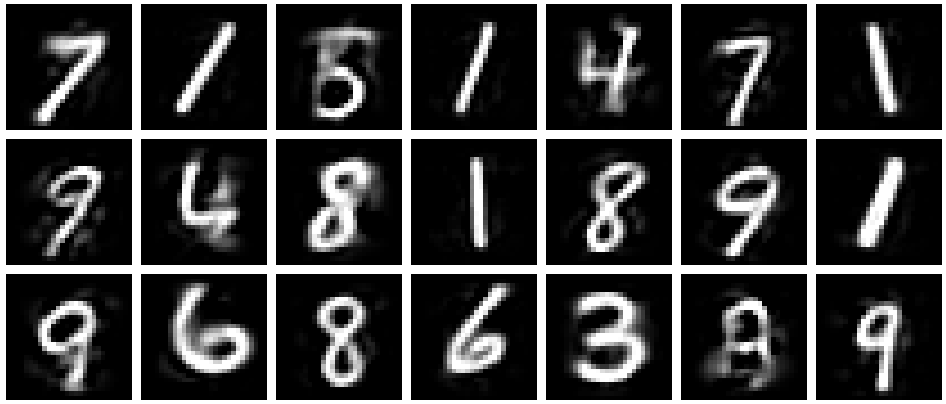} & 
     \includegraphics[width=1.0\linewidth]{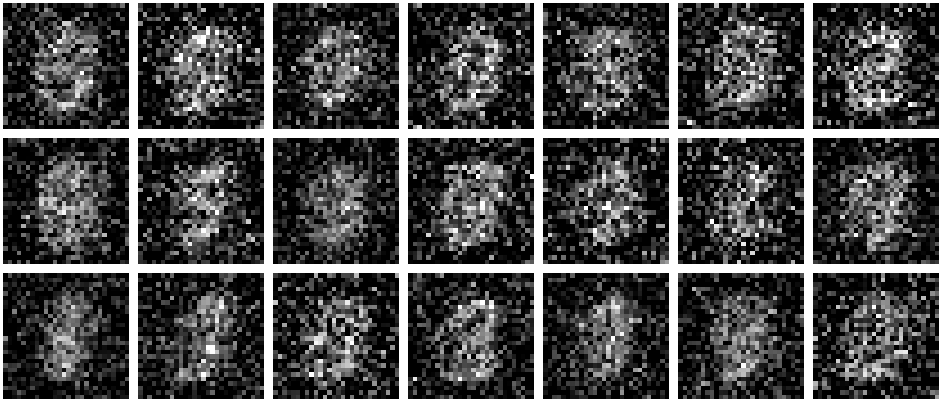}  \\[-1pt] \cmidrule{2-4}
     &
     \rotatebox{90}{ \quad\(\alpha_0 = 100\)} &\includegraphics[width=1.0\linewidth]{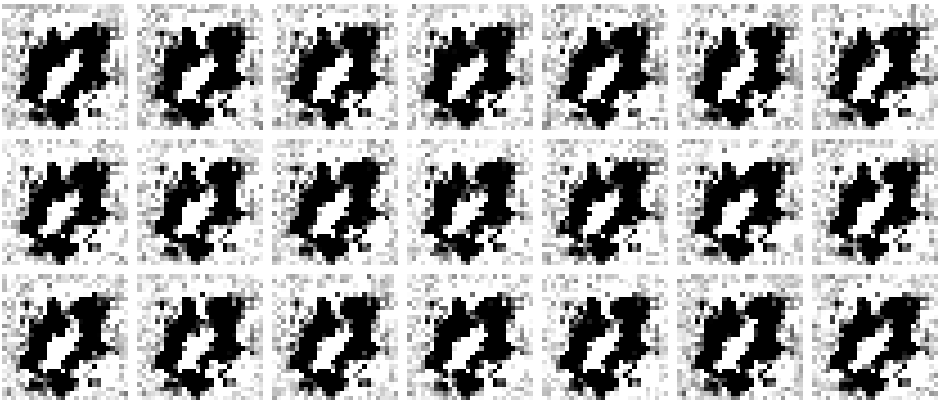} & 
     \includegraphics[width=1.0\linewidth]{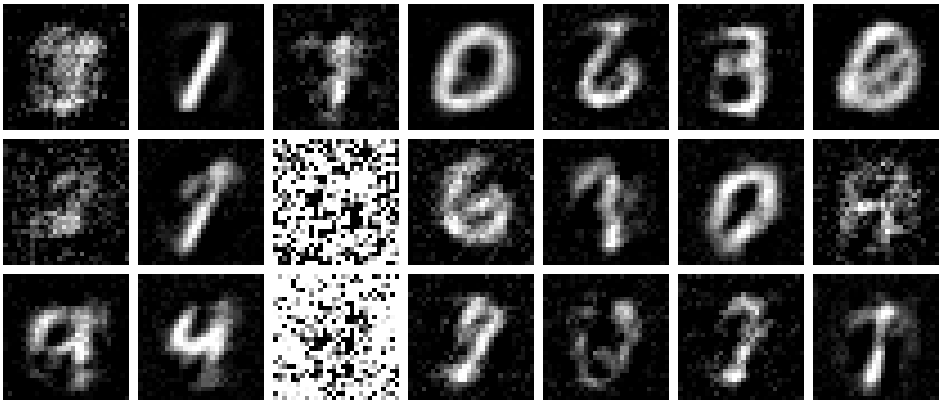}\\[-1pt] \midrule\midrule
     \multirow{9}*{\rotatebox{90}{ \(\gamma_t = \sqrt{2\alpha_t};~\forall~t\)\qquad\enskip}} &
     \rotatebox{90}{ \enskip\quad\(\alpha_0 = 1\)} &\includegraphics[width=1.0\linewidth]{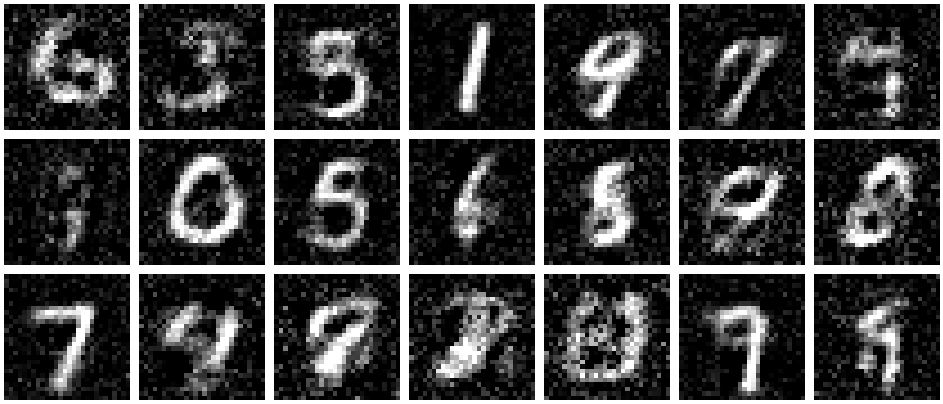} & 
    \includegraphics[width=1.0\linewidth]{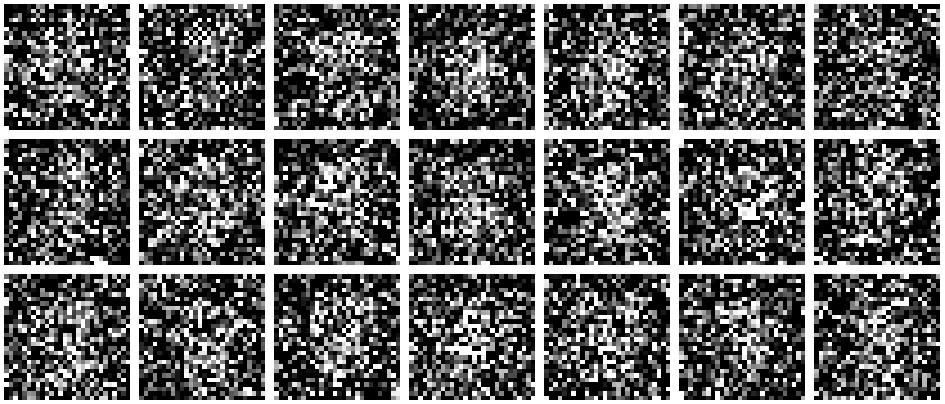} \\[-1pt] \cmidrule{2-4}
    &
    \rotatebox{90}{ \quad\(\alpha_0 = 10\)} &\includegraphics[width=1.0\linewidth]{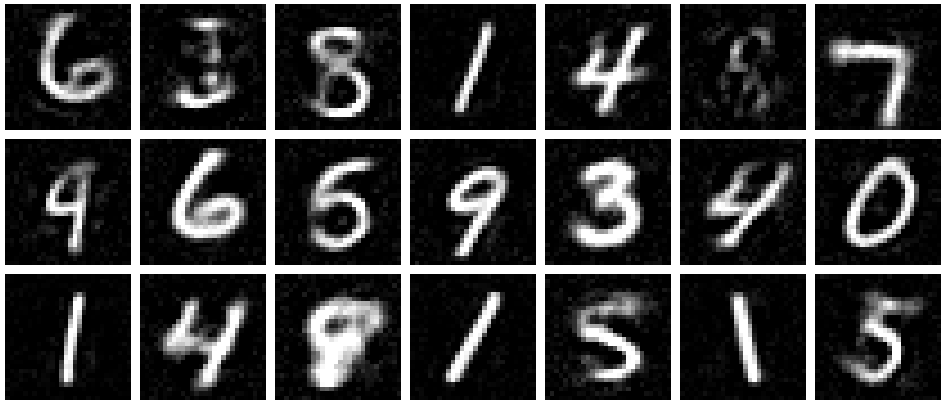} & 
    \includegraphics[width=1.0\linewidth]{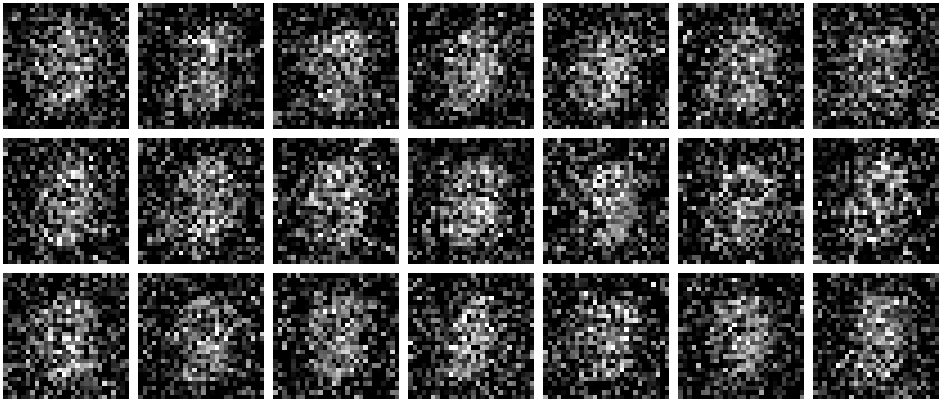} \\[-1pt] \cmidrule{2-4}
    &
    \rotatebox{90}{ \quad\(\alpha_0 = 100\)} &\includegraphics[width=1.0\linewidth]{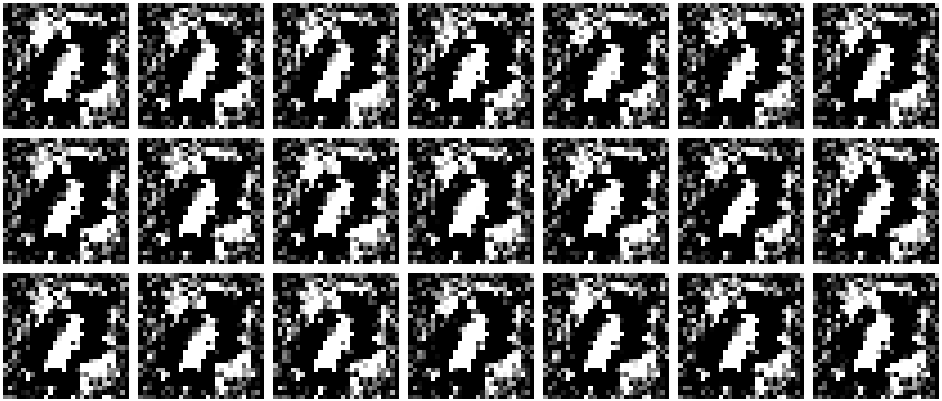} & 
    \includegraphics[width=1.0\linewidth]{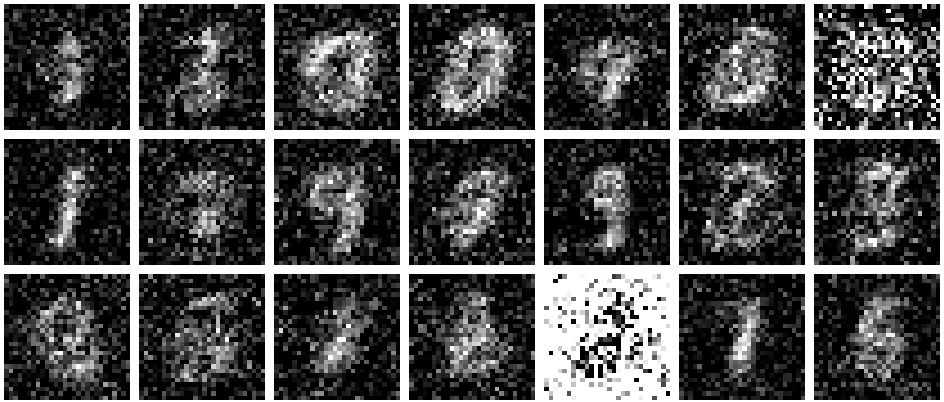}\\[-1pt] 
    \bottomrule
    \end{tabular} 
  \caption[]{(\includegraphics[height=0.012\textheight]{Rgb.png} Color online)~Images generated using the discriminator-guided Langevin sampler with MNIST as the target. The model fails to converge when \(\alpha_t\) decays, for small \(\alpha_0 \leq 10\). When \(\alpha_0=100\), some samples diverge due to gradient explosion. We observe that \(\alpha_0=10\), with \(\z_t = \bm{0} \) yields the best performance. }
  \label{Fig_BetaCompares_MNSIT}  
  \end{center}
  \vskip1cm
\end{figure*}

\begin{figure*}[!t]
  \begin{center}
    \begin{tabular}[b]{P{.02\linewidth}|P{.02\linewidth}|P{.4\linewidth}P{.4\linewidth}}
      && Constant \(\alpha_t = \alpha_0~\forall~t\)  & Geometrically decaying \(\alpha_t\)  \\[1pt]
      \toprule
      \multirow{6}*{\rotatebox{90}{ \(\gamma_t = 0;~\forall~t\)\qquad\enskip}} & \rotatebox{90}{ \enskip\quad\(\alpha_0 = 1\)} &
      \includegraphics[width=1.0\linewidth]{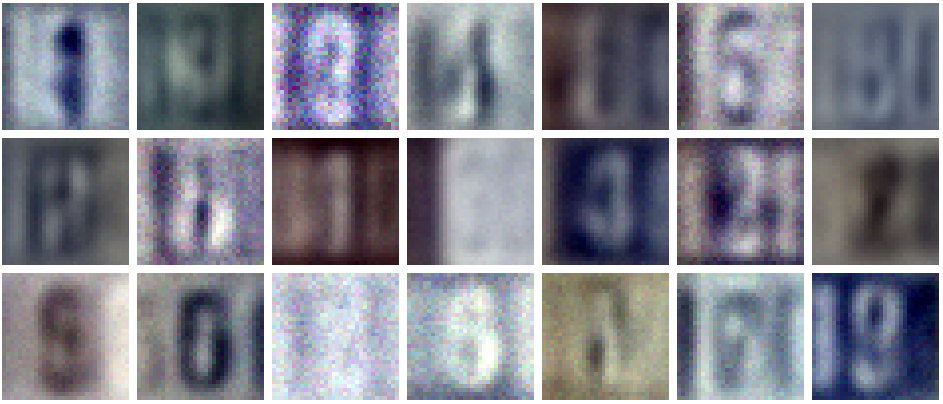} & 
     \includegraphics[width=1.0\linewidth]{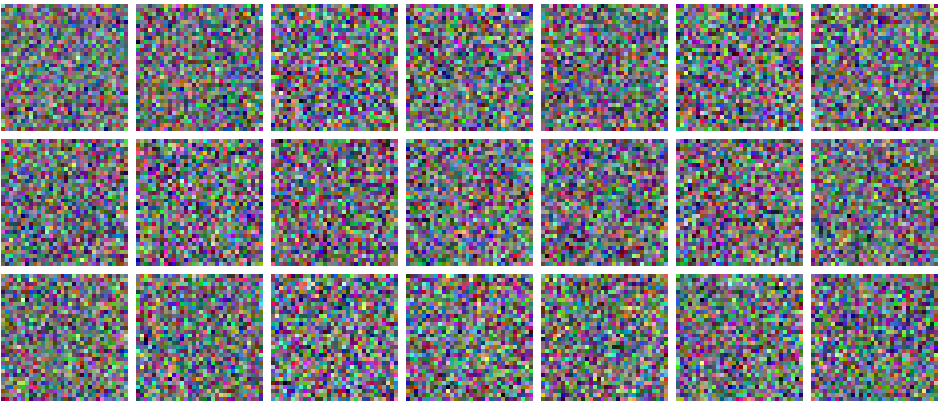}  \\[-1pt] \cmidrule{2-4}
      &
      \rotatebox{90}{ \quad\(\alpha_0 = 10\)} &\includegraphics[width=1.0\linewidth]{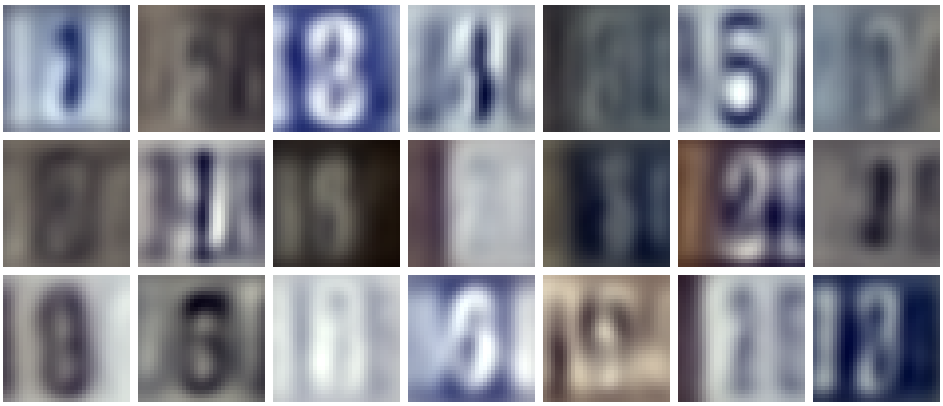} & 
     \includegraphics[width=1.0\linewidth]{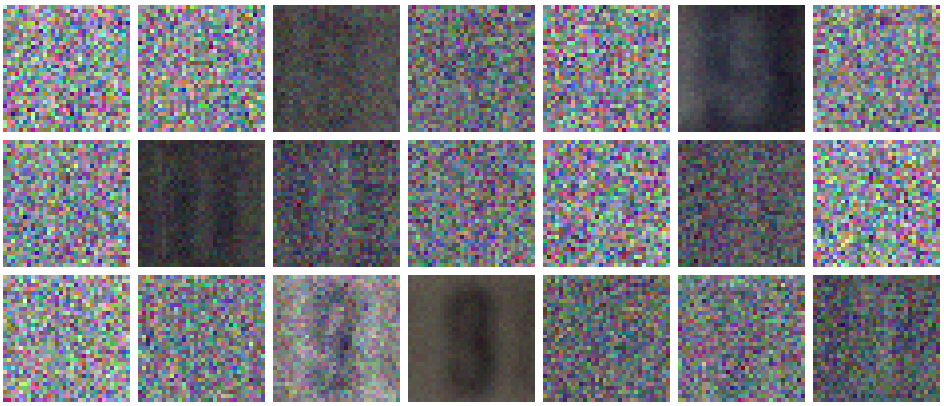}  \\[-1pt] \cmidrule{2-4}
     &
     \rotatebox{90}{ \quad\(\alpha_0 = 100\)} &\includegraphics[width=1.0\linewidth]{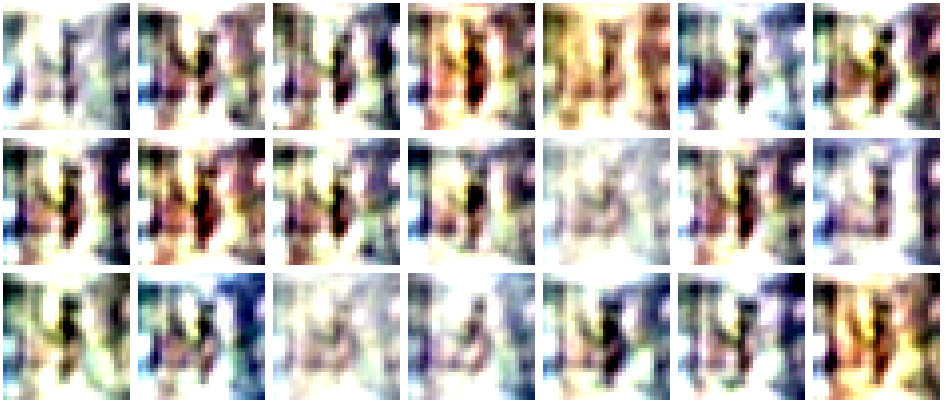} & 
     \includegraphics[width=1.0\linewidth]{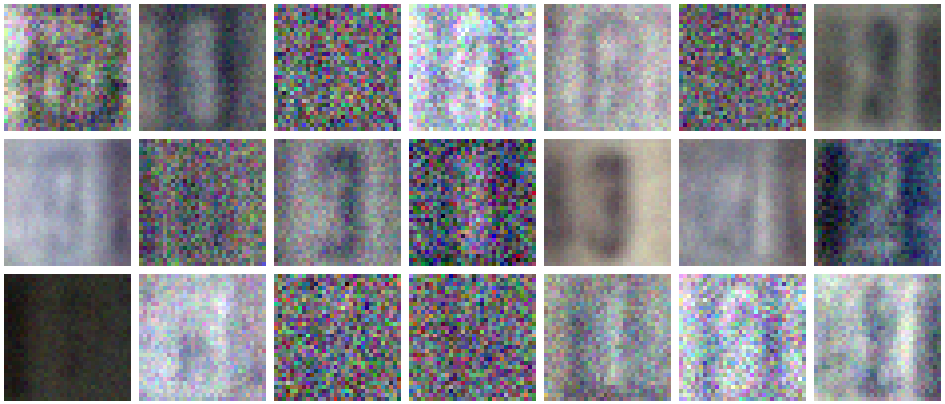}\\[-1pt] \midrule\midrule
     \multirow{9}*{\rotatebox{90}{ \(\gamma_t = \sqrt{2\alpha_t};~\forall~t\)\qquad\enskip}} &
     \rotatebox{90}{ \enskip\quad\(\alpha_0 = 1\)} &\includegraphics[width=1.0\linewidth]{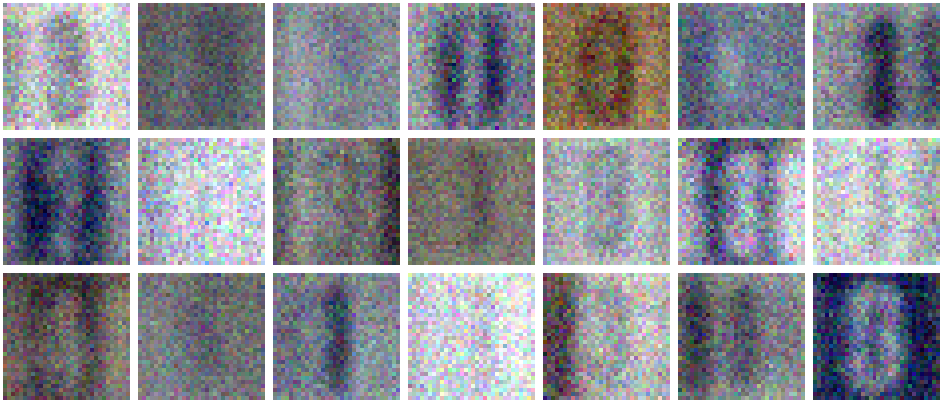} & 
    \includegraphics[width=1.0\linewidth]{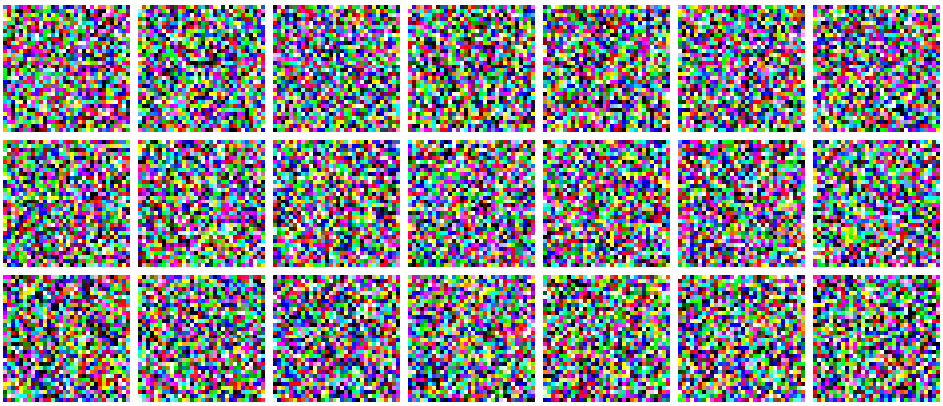} \\[-1pt] \cmidrule{2-4}
    &
    \rotatebox{90}{ \quad\(\alpha_0 = 10\)} &\includegraphics[width=1.0\linewidth]{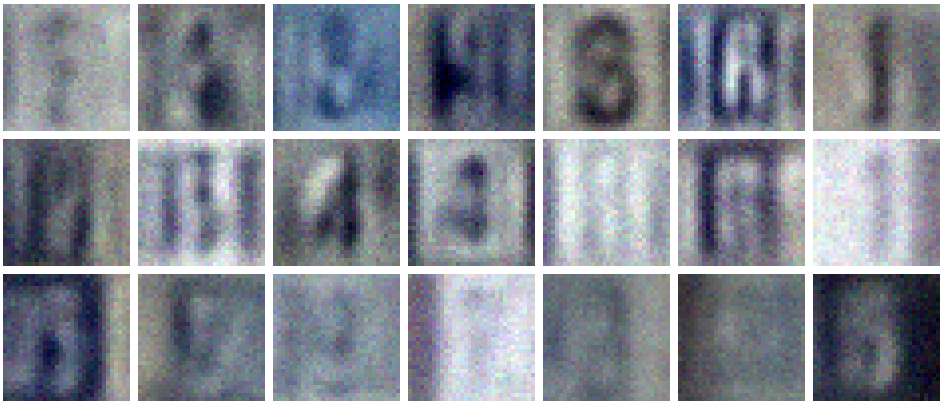} & 
    \includegraphics[width=1.0\linewidth]{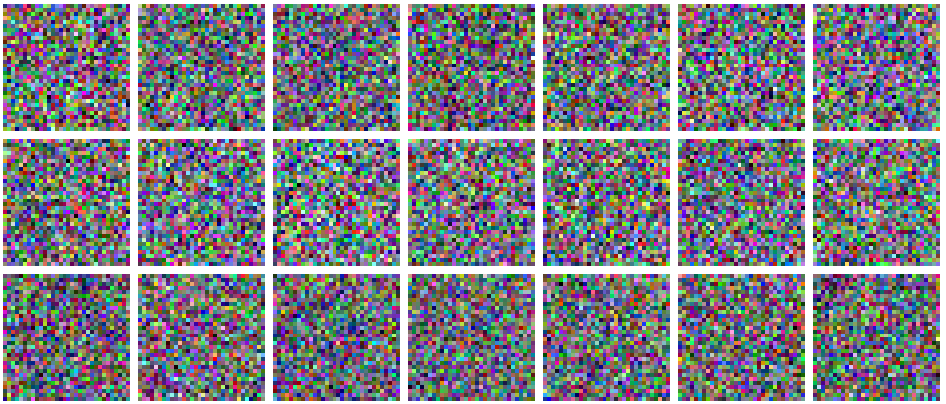} \\[-1pt] \cmidrule{2-4}
    &
    \rotatebox{90}{ \quad\(\alpha_0 = 100\)} &\includegraphics[width=1.0\linewidth]{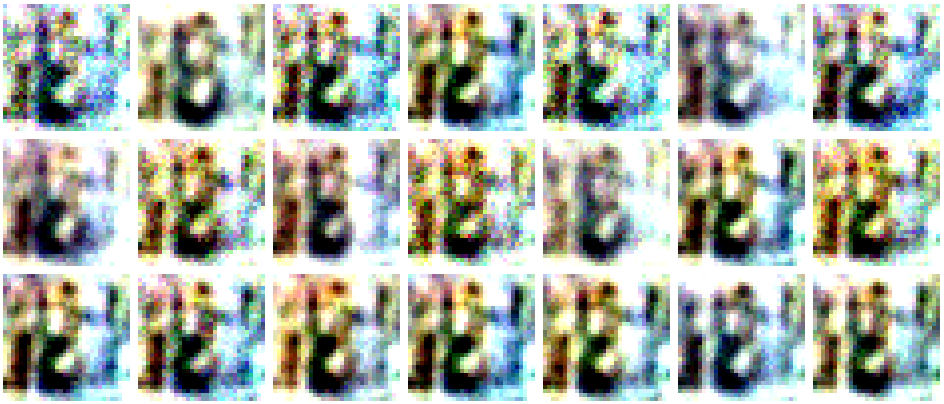} & 
    \includegraphics[width=1.0\linewidth]{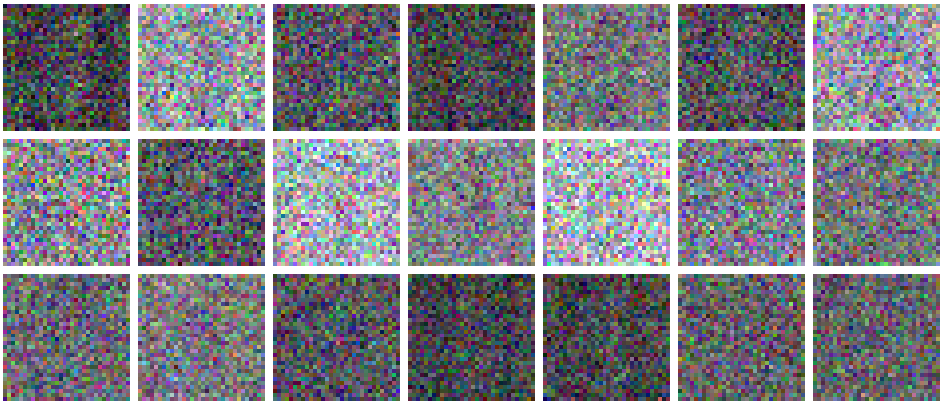}\\[-1pt] 
    \bottomrule
    \end{tabular} 
    \caption[Images generated using the discriminator-guided Langevin sampler.]{(\includegraphics[height=0.012\textheight]{Rgb.png} Color online)~Images generated using the discriminator-guided Langevin sampler with SVHN as the target. The model fails to converge with geometrically decaying \(\alpha_t\), or when \(\z_t\) is not the zero vector. As in the case of MNIST, observe that \(\alpha_0 = 10\), with \(\z_t = 0 \) yields the best performance. Setting \(\alpha_0 =1\) with \(\z_t = 0\) results in slow convergence. }
    \label{Fig_BetaCompares_SVHN}  
  \end{center}
  \vskip1cm
\end{figure*}

\begin{figure*}[!t]
  \begin{center}
    \begin{tabular}[b]{P{.02\linewidth}|P{.02\linewidth}|P{.4\linewidth}P{.4\linewidth}}
      && Constant \(\alpha_t = \alpha_0~\forall~t\)  & Geometrically decaying \(\alpha_t\)  \\[1pt]
      \toprule
      \multirow{6}*{\rotatebox{90}{ \(\gamma_t = 0;~\forall~t\)\qquad\enskip}} & \rotatebox{90}{ \enskip\quad\(\alpha_0 = 1\)} &
      \includegraphics[width=1.0\linewidth]{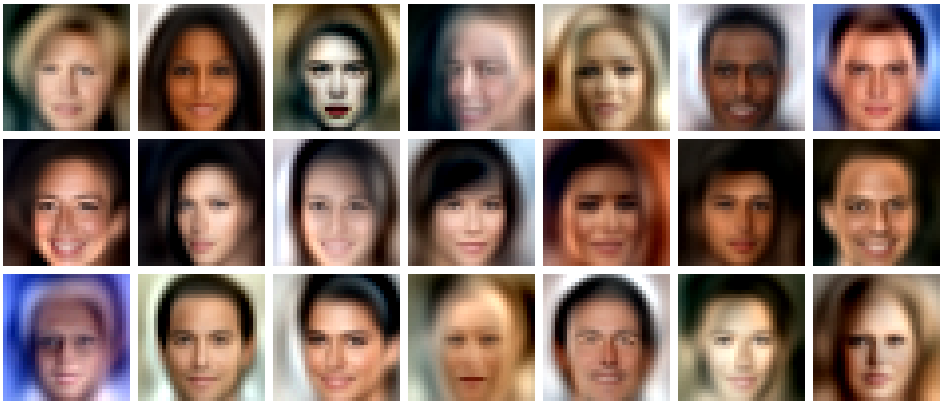} & 
     \includegraphics[width=1.0\linewidth]{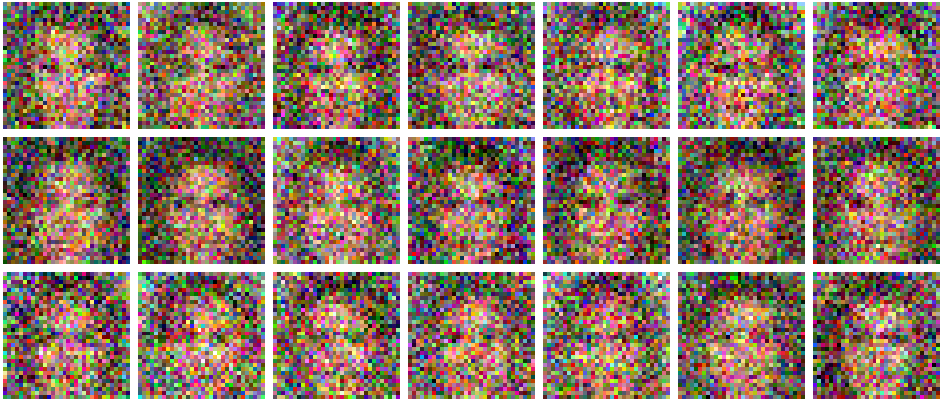}  \\[-1pt] \cmidrule{2-4}
      &
      \rotatebox{90}{ \quad\(\alpha_0 = 10\)} &\includegraphics[width=1.0\linewidth]{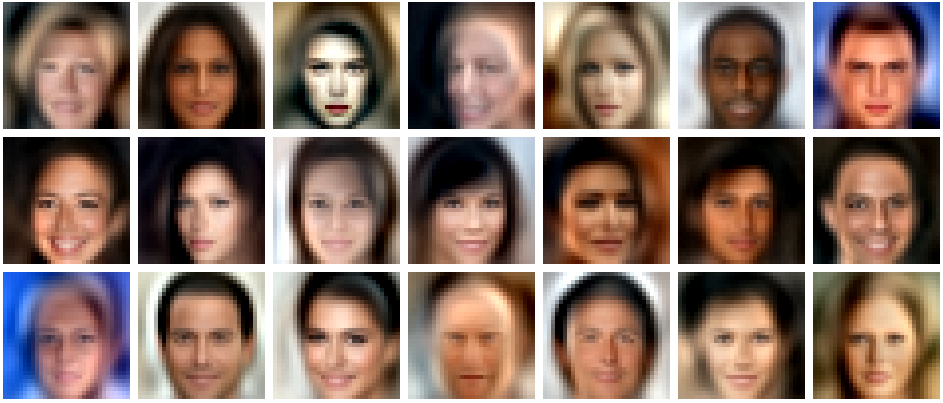} & 
     \includegraphics[width=1.0\linewidth]{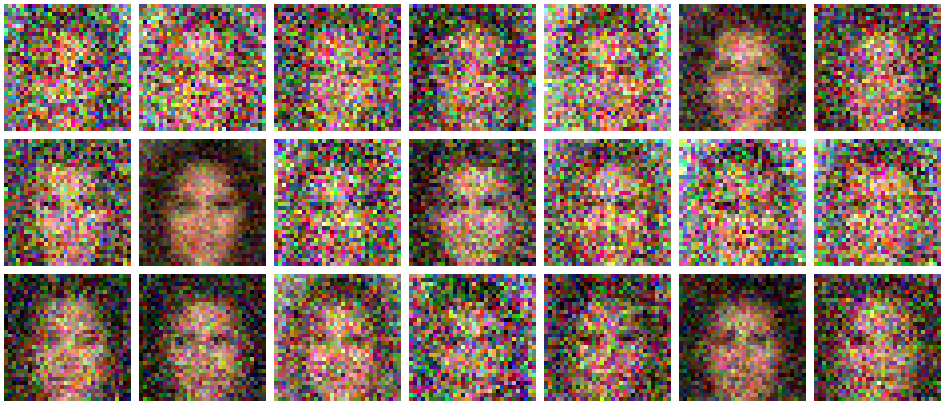}  \\[-1pt] \cmidrule{2-4}
     &
     \rotatebox{90}{ \quad\(\alpha_0 = 100\)} &\includegraphics[width=1.0\linewidth]{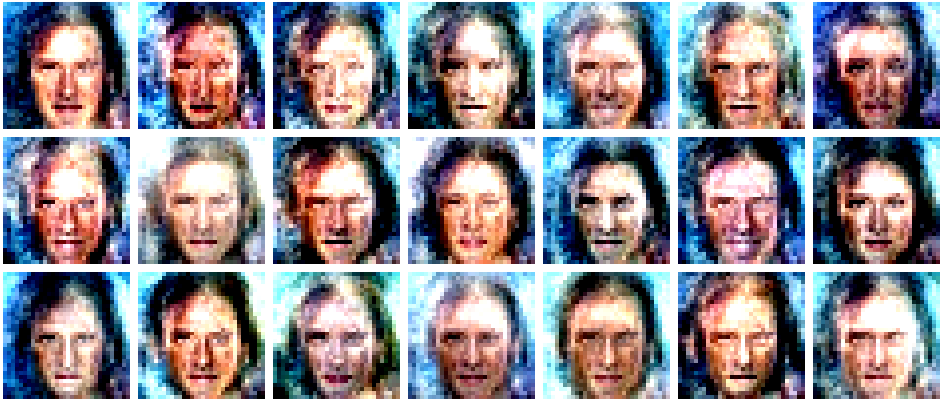} & 
     \includegraphics[width=1.0\linewidth]{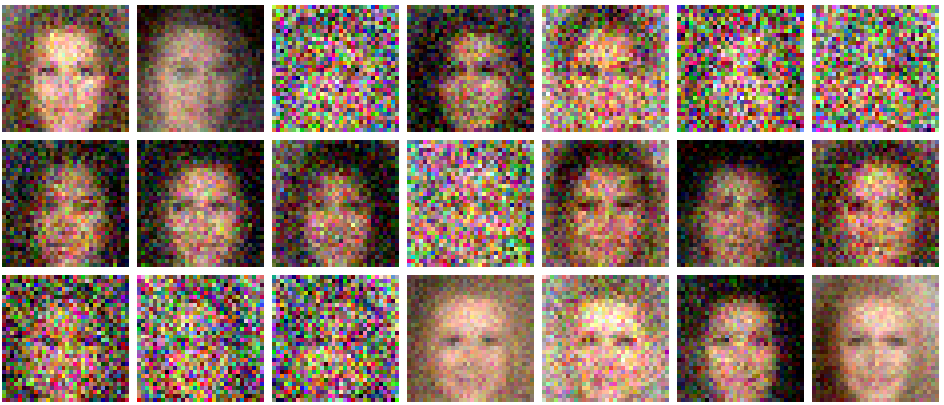}\\[-1pt] \midrule\midrule
     \multirow{9}*{\rotatebox{90}{ \(\gamma_t = \sqrt{2\alpha_t};~\forall~t\)\qquad\enskip}} &
     \rotatebox{90}{ \enskip\quad\(\alpha_0 = 1\)} &\includegraphics[width=1.0\linewidth]{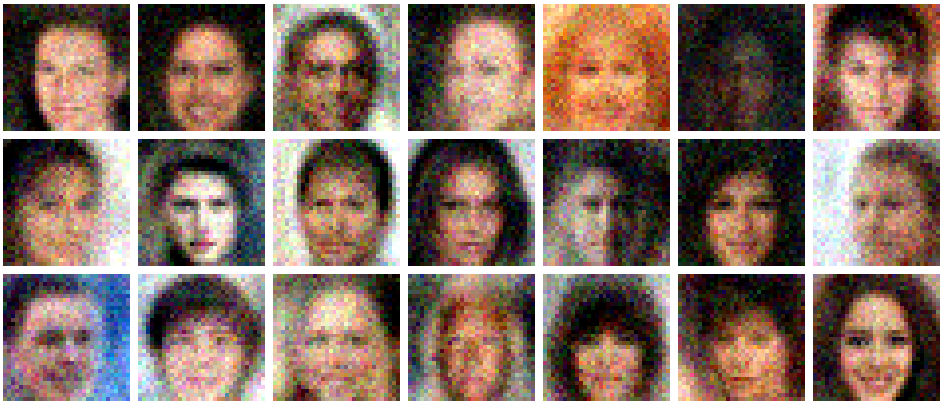} & 
    \includegraphics[width=1.0\linewidth]{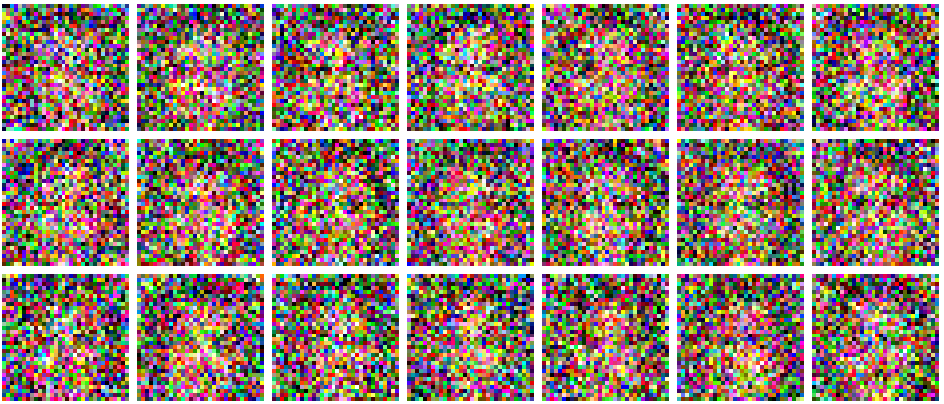} \\[-1pt] \cmidrule{2-4}
    &
    \rotatebox{90}{ \quad\(\alpha_0 = 10\)} &\includegraphics[width=1.0\linewidth]{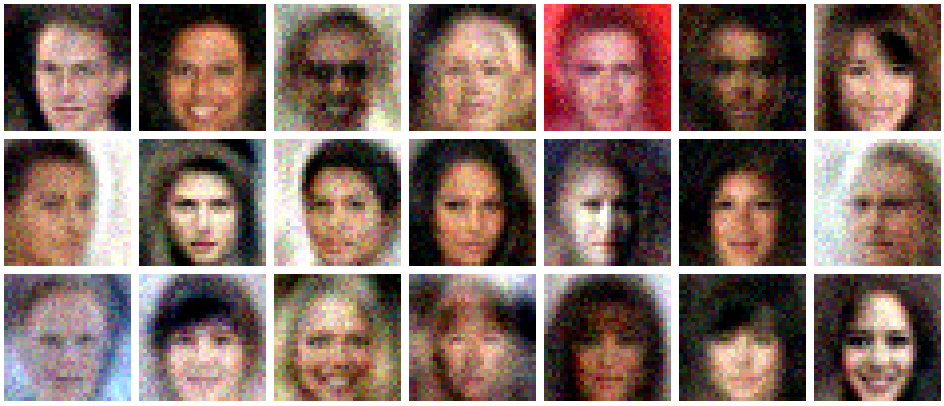} & 
    \includegraphics[width=1.0\linewidth]{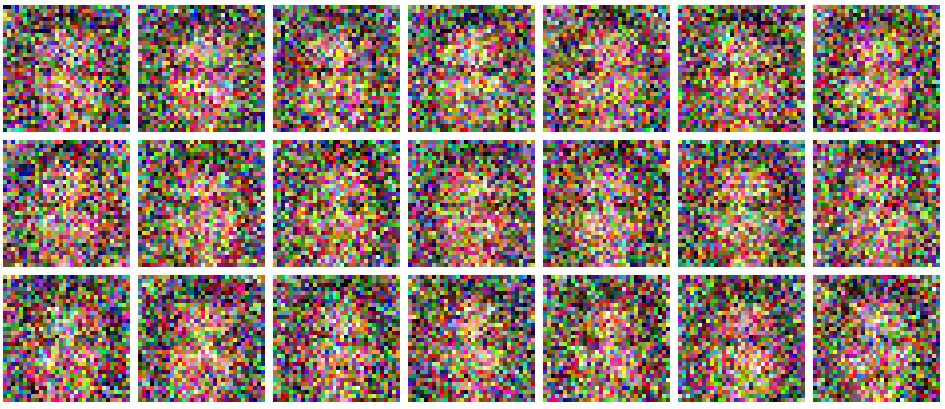} \\[-1pt] \cmidrule{2-4}
    &
    \rotatebox{90}{ \quad\(\alpha_0 = 100\)} &\includegraphics[width=1.0\linewidth]{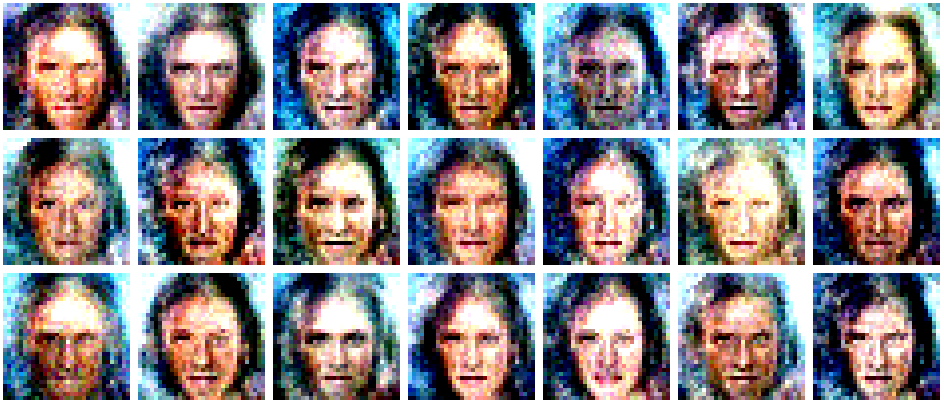} & 
    \includegraphics[width=1.0\linewidth]{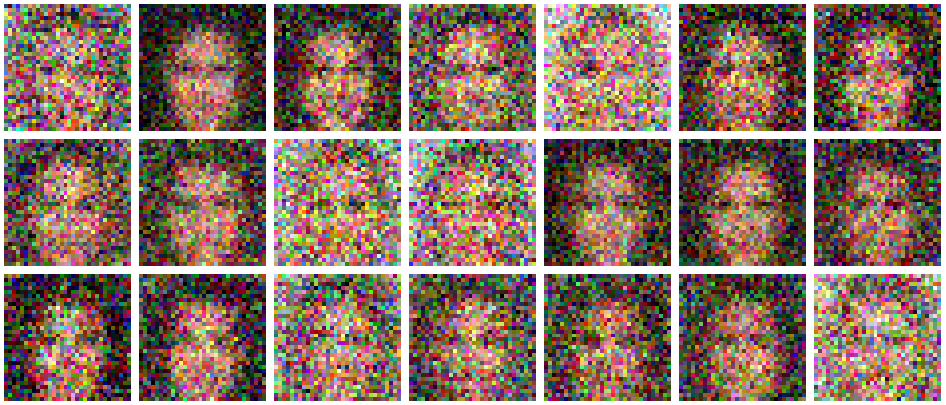}\\[-1pt] 
    \bottomrule
    \end{tabular} 
    \caption[Images generated using the discriminator-guided Langevin sampler.]{(\includegraphics[height=0.012\textheight]{Rgb.png} Color online)~Images generated using the discriminator-guided Langevin sampler with CelebA as the target. The model fails to converge when \(\alpha_t\) decays geometrically, or when \(\z_t \neq \bm{0}\). Setting \(\alpha_0 \in [1,10]\), with \(\z_t = \bm{0} \) results in the sampler generating realistic images. For these choices of \(\alpha_0\), when \(\z_t \neq \bm{0}\), the generated images are noisy.}
    \label{Fig_BetaCompares_CelebA}  
  \end{center}
  \vskip1cm
\end{figure*}

Figures~\ref{Fig_BetaCompares_MNSIT}--\ref{Fig_BetaCompares_CelebA} show the images generated by the discriminator-guided Langevin sampler on MNIST, SVHN and CelebA, respectively, for the various scenarios considered. Across all datasets, we observe that annealing the coefficients results in poor convergence. We attribute this to the fact that the polyharmonic kernel, being a distance function, decays {\it automatically} as the iterates converge, {\it i.e.,} as \(p_t\) approaches \(\pd\). Consequently, the magnitude of the discriminator gradient, in the case when \(\alpha_t\) is decays, is too small to significantly move the particles along the discriminator gradient field. Next, we observe that for relatively small \(\alpha_0 \leq 10\), the samplers converge to realistic images. When \(\alpha_0\) is large, the resulting {\it gradient explosion} during the initial steps of the sampler results in {\it mode-collapse} in all scenarios. Thirdly, in choosing \(\z_t\), the experimental results indicate that the model converges to visually superior images when \(\z_t = 0\). For the scenarios where \(\alpha_t\), the coefficient of \(\nabla_{\x}D_t^*\), is kept constant, but the coefficient \(\gamma_t\) decays with $t$ as in the baseline setting. When \(\z_t\) is non-zero, the generated images are noisy. We attribute the convergence of the discriminator-guided Langevin sampler to unique samples even in scenarios when \(\z_t\) is zero, to the implicit randomness of the centers of the radial basis function kernels introduced by the sample estimates in the discriminator \(D_t^*\). \par

\begin{figure*}[!th]
  \begin{center}
    \begin{tabular}[b]{P{.45\linewidth}P{.5\linewidth}}
            \includegraphics[width=1\linewidth]{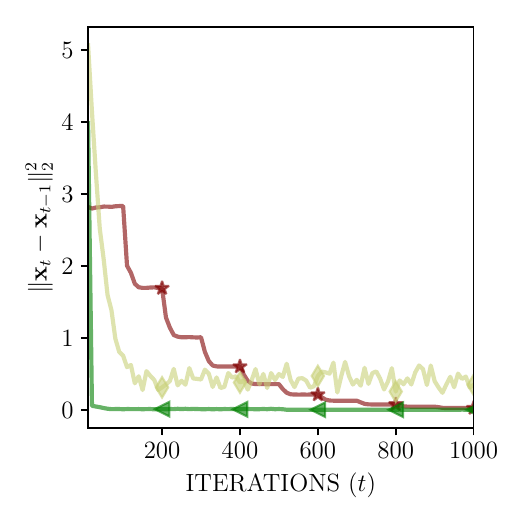}   &
            \includegraphics[width=1\linewidth]{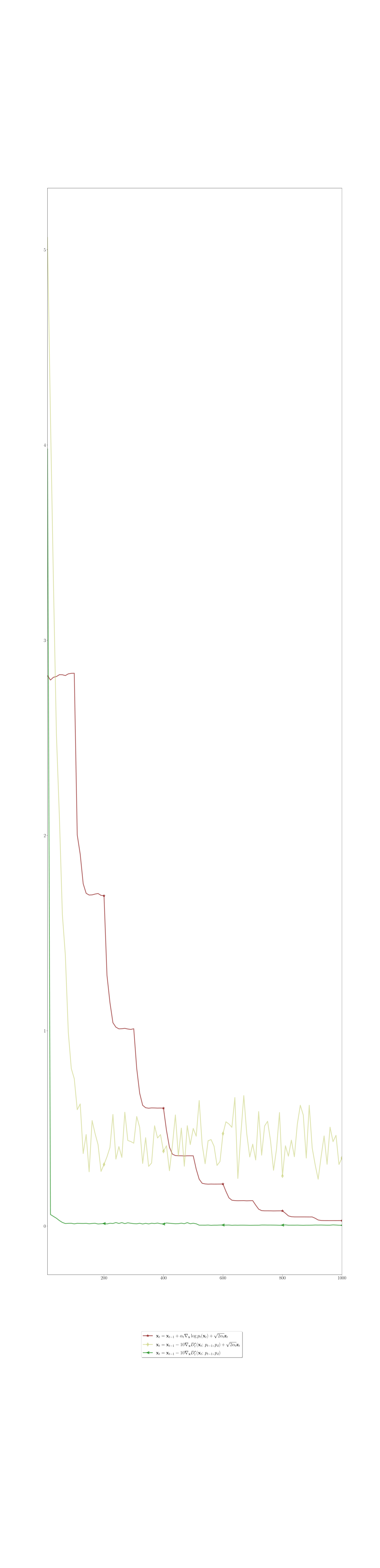}    \\
    \end{tabular} 
    \caption[]{(\includegraphics[height=0.012\textheight]{Rgb.png} Color online)~Plot comparing the {\it iterate convergence} of the discriminator-guided Langevin diffusion model, compared against the baseline NCSNv1~\citep{NCSN19} model. The score in NCSN is replaced with the output of a score network \(S_{\theta}\). The norm of the iterate-differences decays as the noise-scale in the case of NCSN. This is consistent with the observations made by~\citet{NCSNv220}, who showed that the score network \(S_{\theta}\) implicitly scales its output by the noise variance \(\sigma\). In discriminator-guided Langevin diffusion, adding noise results in poorer performance, while the unadjusted Langevin sampler performs the best. } 
    \vspace{-1.2em}
    \label{Fig_NormDiff}  
    \end{center}
  \end{figure*}

The superior convergence of the proposed approach is further validated by the {\it iterate convergence} presented in Figure~\ref{Fig_NormDiff}. We compare discriminator-guided Langevin sampler, with \(\alpha_t = \alpha_0 = 10\), with and without noise perturbations \(\z_t\), against the base NCSN model, owing to the links to the score-based results derived. We plot \(\|\x_t - \x_{t-1}\|_2^2\) as a function of iteration \(t\) for the MNIST learning task. In NCSN, the iterates converge at each noise level, and subsequently, when the noise level drops, the sample quality improved. This is consistent with the observations made by~\citet{NCSNv220}, who showed that the score network \(S_{\theta}\) implicitly scales its output by the noise variance \(\sigma\). The proposed approach, with \(\z_t = \bm{0}\), performs the best. \par

\begin{figure*}[!th]
  \begin{center}
    \begin{tabular}[b]{c|c}
    \(\x_T\) & \(k\)-nearest neighbors of \(x_T\) (\(k = 9\))  \\[5pt]
            \includegraphics[height=0.95\linewidth]{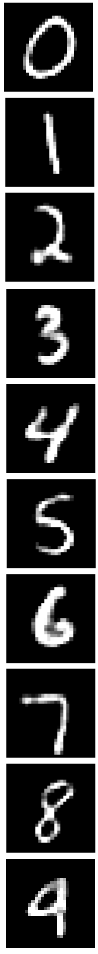}   &
            \includegraphics[height=0.95\linewidth]{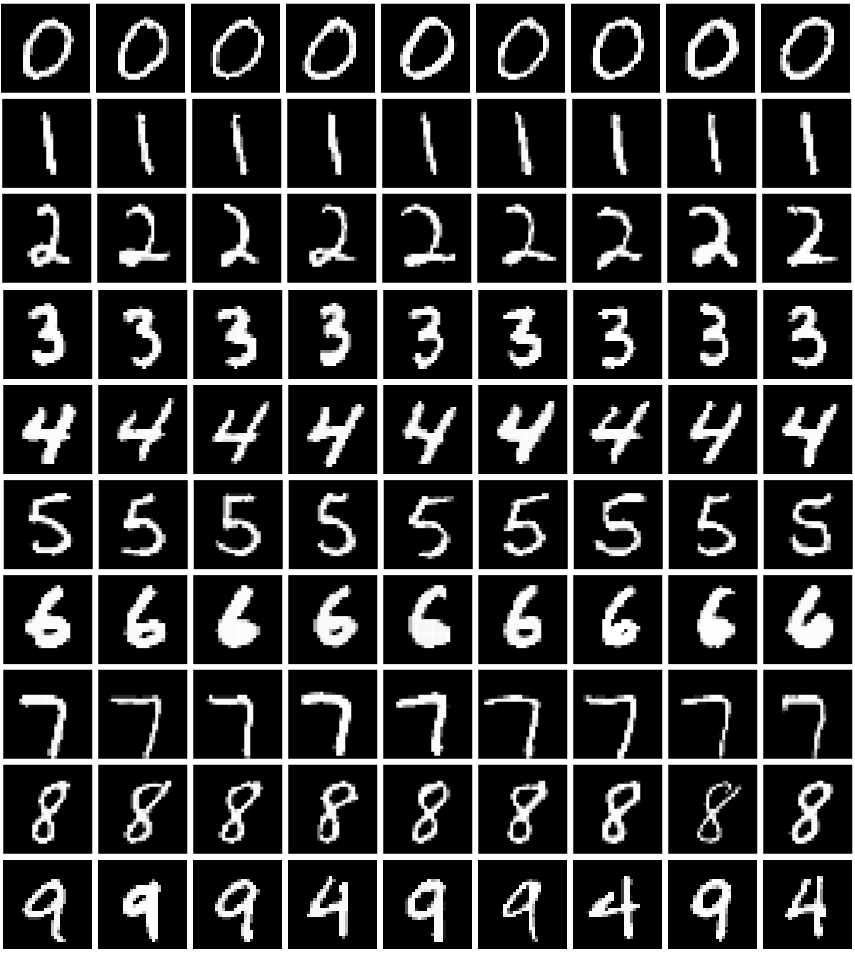}   \\
    \end{tabular} 
    \caption[]{(\includegraphics[height=0.012\textheight]{Rgb.png} Color online)~The \(k\)-nearest neighbor (\(k\)-NN) test performed on images generated by the discriminator-guided Langevin sampler, when \(\alpha_t = \alpha_0 = 10\) and \(\z_t = 0\), on the MNIST dataset. We observe that the generated images are unique and distinct from the top-9 neighbors drawn from the target dataset, indicating that the sampler {\bf does not memorize} the images seen as part of the interpolating RBF discriminator's centers.} 
    \vspace{-1.2em}
    \label{Fig_MNIST_KNN}  
    \end{center}
  \end{figure*}

\begin{figure*}[!th]
  \begin{center}
    \begin{tabular}[b]{c|c}
    \(\x_T\) & \(k\)-nearest neighbors of \(x_T\) (\(k = 9\))  \\[5pt]
            \includegraphics[height=0.95\linewidth]{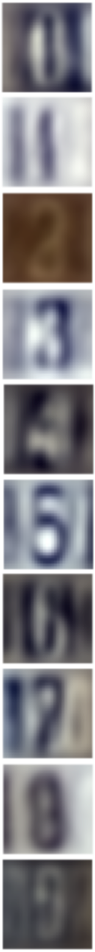}   &
            \includegraphics[height=0.95\linewidth]{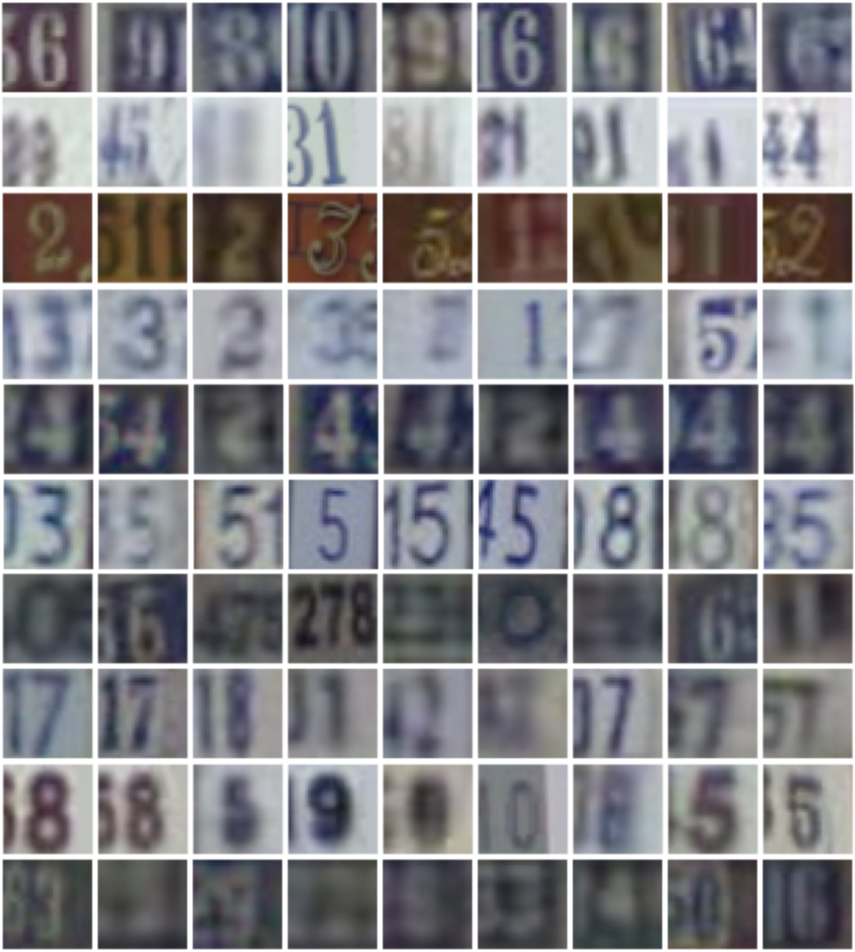}   \\
    \end{tabular} 
    \caption[]{(\includegraphics[height=0.012\textheight]{Rgb.png} Color online)~The \(k\)-nearest neighbor (kNN) test performed on images generated by the discriminator-guided Langevin sampler, when \(\alpha_t = \alpha_0 = 10\) and \(\z_t = \bm{0}\), on the SVHN dataset. We observe that the generated images are unique, compared to the top-9 neighbors drawn from the target dataset. For generated samples such as the {\it digit 9} or {\it digit 5}, we observe that the top \(k\)-NN images are from classes different from that of the generated image, indicative of the model's ability to interpolate between the classes seen as part of discriminator centers during sampling. } 
    \vspace{-1.2em}
    \label{Fig_SVHN_KNN}  
    \end{center}
  \end{figure*}

\begin{figure*}[!th]
  \begin{center}
    \begin{tabular}[b]{c|c}
    \(\x_T\) & \(k\)-nearest neighbors of \(x_T\) (\(k = 9\))  \\[5pt]
            \includegraphics[height=0.95\linewidth]{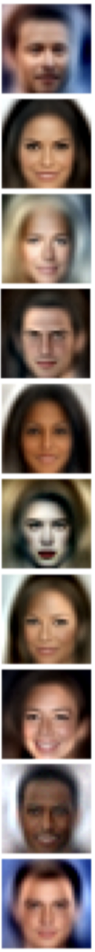}   &
            \includegraphics[height=0.95\linewidth]{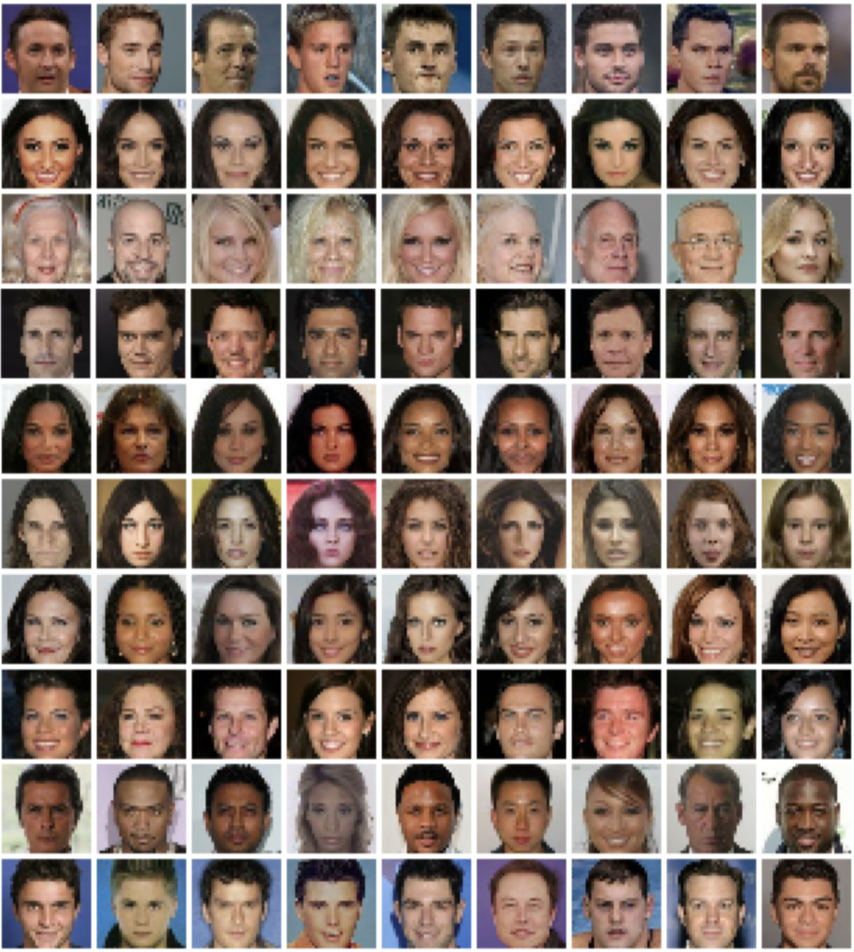}   \\
    \end{tabular} 
    \caption[]{(\includegraphics[height=0.012\textheight]{Rgb.png} Color online)~The \(k\)-nearest neighbor (kNN) test performed on images generated by the discriminator-guided Langevin sampler, when \(\alpha_t = \alpha_0 = 10\) and \(\z_t = \bm{0}\), on the CelebA dataset. The generated images are unique and distinct from the top-9 neighbors drawn from the target dataset, which suggests that the proposed approach does not memorize data. } 
    \vspace{-1.2em}
    \label{Fig_CelebA_KNN}  
    \end{center}
  \end{figure*}

{\it {\bfseries Uniqueness of generated images}}: As the kernel-based discriminator operates directly on the target data, drawing batches of samples as centers in the RBF interpolator, an obvious question to ask is whether the discriminator-guided Langevin iterations converge to unique samples \emph{not seen in the dataset}. To verify this, we perform a \(k\)-nearest neighbor analysis, considering \(k=9\) in the experiments. Figures~\ref{Fig_MNIST_KNN}--~\ref{Fig_CelebA_KNN} present the top-\(k\) neighbors of samples generated by the proposed images from each digit class of MNIST, SVHN, and CelebA datasets. The neighbors are found across all {\it digit} classes in the case of MNIST and SVHN. It is clear from these results that the proposed approach {\bf does not} memorize the dataset. In the case of SVHN, considering the samples generated from {\it digit class 5} of {\it digit class 9}, we observe that the nearest neighbor is from a different class, indicative of the sampler's ability to interpolate between the classes seen as part of discriminator centers during sampling. \par

\begin{figure*}[!t]
  \begin{center}
    \begin{tabular}[b]{P{.01\linewidth}|P{.245\linewidth}P{.245\linewidth}P{.245\linewidth}}
      &{\small MNIST \((\mbbR^{28\times28\times1})\)} & {\small SVHN \((\mbbR^{32\times32\times3})\)}  & {\small Ukiyo-E \((\mbbR^{256\times256\times3})\)}    \\[1pt]
      \rotatebox{90}{ ~~~\qquad\quad \(t = 10\)} &
      \includegraphics[width=0.987\linewidth]{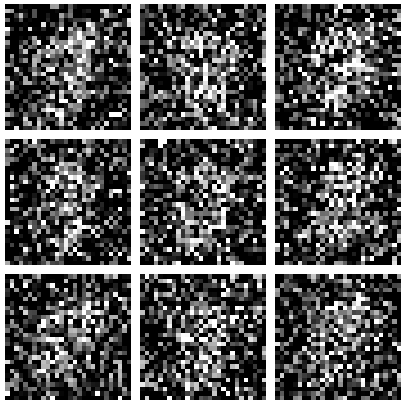} & 
     \includegraphics[width=0.987\linewidth]{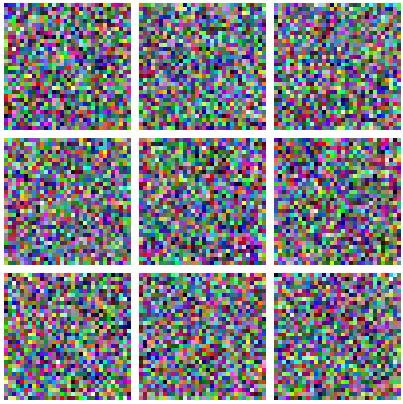}  &
      \includegraphics[width=0.987\linewidth]{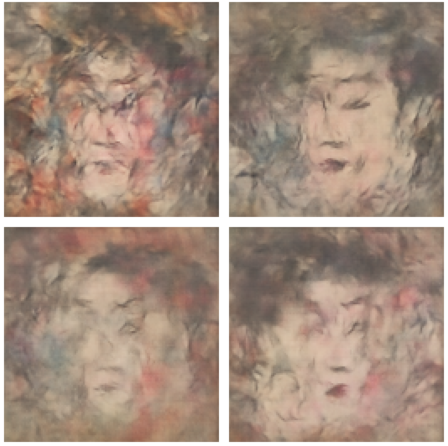}  \\[2pt]
      \rotatebox{90}{ ~~~\qquad\quad \(t = 100\)} &
      \includegraphics[width=0.987\linewidth]{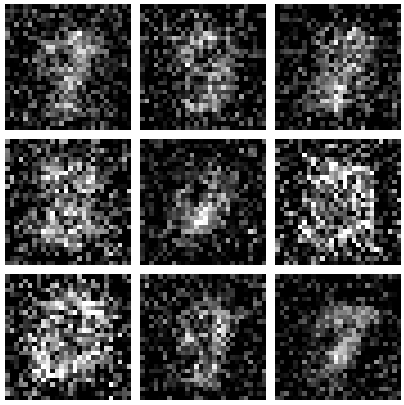} & 
     \includegraphics[width=0.987\linewidth]{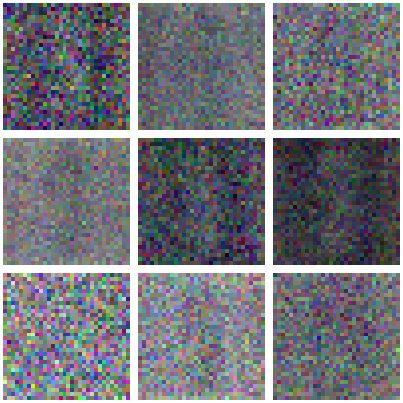}  &
      \includegraphics[width=0.987\linewidth]{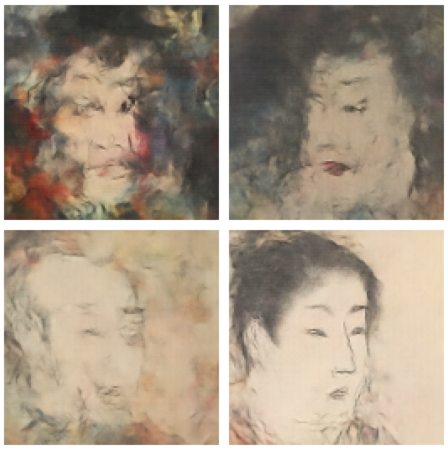}  \\[2pt]
      \rotatebox{90}{ ~~~\qquad\quad \(t = 500\)} &
      \includegraphics[width=0.987\linewidth]{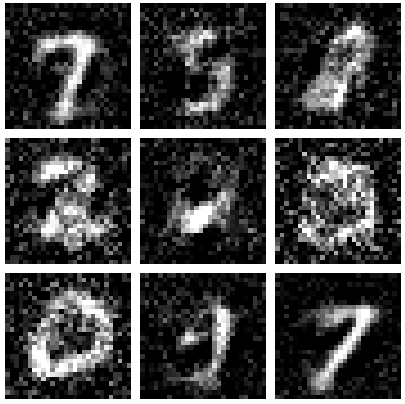} & 
     \includegraphics[width=0.987\linewidth]{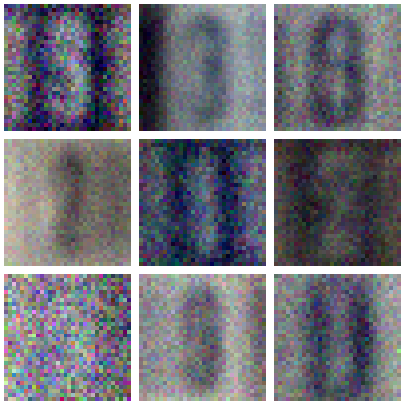}  &
      \includegraphics[width=0.987\linewidth]{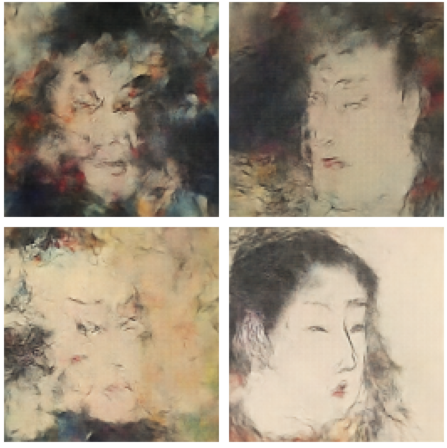}  \\[2pt]
      \rotatebox{90}{ \qquad\quad \(t = 1\times10^3\)} &
      \includegraphics[width=0.987\linewidth]{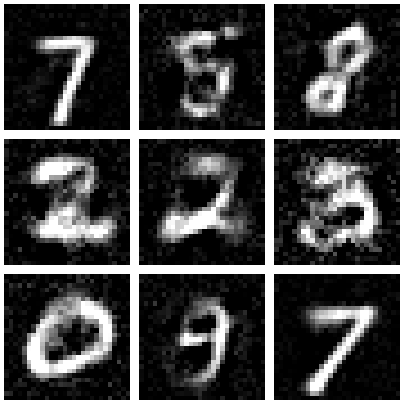} & 
     \includegraphics[width=0.987\linewidth]{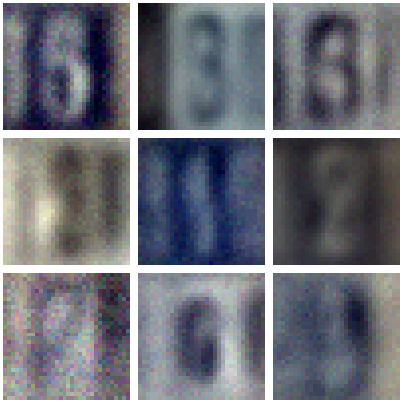}  &
      \includegraphics[width=0.987\linewidth]{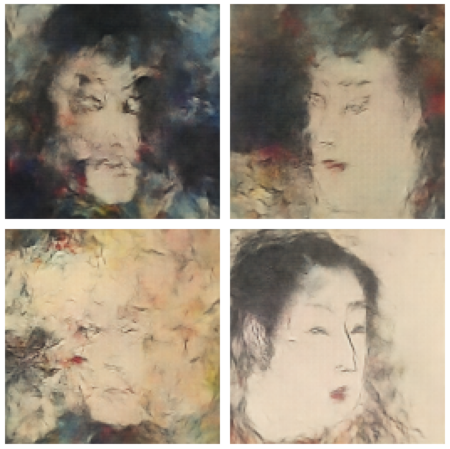}  \\[2pt]
      \rotatebox{90}{ \qquad\quad \(t = 2\times10^3\)} &
      \includegraphics[width=0.987\linewidth]{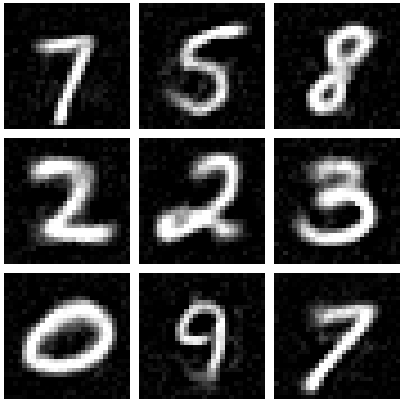} & 
     \includegraphics[width=0.987\linewidth]{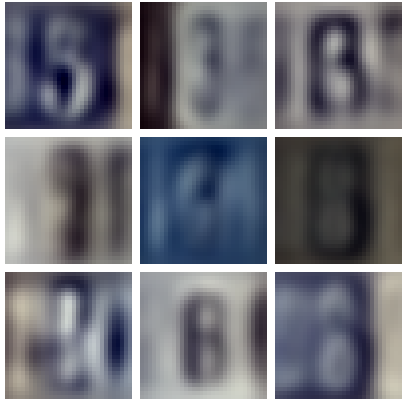}  &
      \includegraphics[width=0.987\linewidth]{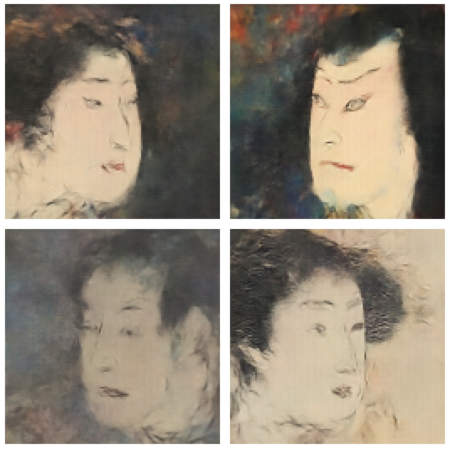}   \\[-5pt]
    \end{tabular} 
  \caption[Images generated using the discriminator-guided Langevin sampler.]{(\includegraphics[height=0.012\textheight]{Rgb.png} Color online)~Images generated using the discriminator-guided Langevin sampler. The score in standard diffusion models is replaced with the gradient field of the discriminator, obviating the need for any trainable neural network, while generating realistic samples.}
  \label{Fig_DiscDiffusion}  
  \end{center}
  \vskip1cm
\end{figure*}

{\it {\bfseries Details on the experiment presented in Section~\ref{Sec:DiscInfusion} of the Main Manuscript}}: Figure~\ref{Fig_DiscDiffusion} presents the images, considering the Langevin sampler with \(\alpha_t = \alpha_0 = 10\) with \(\z_t = 0\). Across all three datasets, we observe that the models converge to nearly realists samples in about \(t = 500\) iterations, while subsequent iterations serve to {\it denoise} the images. Animations pertaining to these iterations are provided as part of the Supplementary Material.

\begin{figure*}[!t]
  \begin{center}
    \begin{tabular}[b]{P{.45\linewidth}P{.45\linewidth}}
    EDM + Heun Sampler   (128 steps)&  {\bf Ours} + Heun Sampler  (40 steps)\\[1pt]
       \includegraphics[width=1.\linewidth]{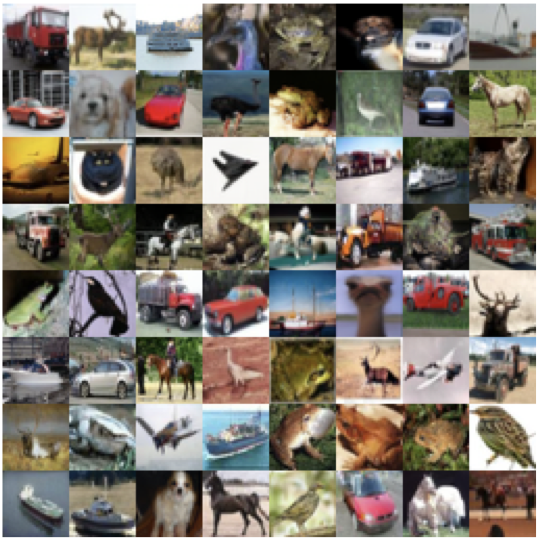} & 
       \includegraphics[width=1.\linewidth]{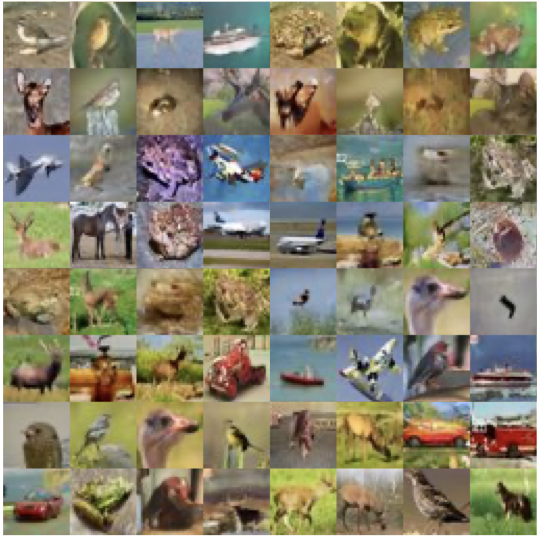} 
        \end{tabular} 
    \caption[]{(\includegraphics[height=0.009\textheight]{Rgb.png} Color online)~ Samples generated by the proposed discriminator-guided Langevin diffusion, compared against the baseline EDM~\citep{EDM22}, on the CIFAR-10 dataset. Both approaches are sampled using the Heun second-order sampler, with sampling parameters as described by~\citet{EDM22}. While the baseline model requires 128 iterations, the proposed sampler generates realistic images in about 40 iterations. }
    \label{Fig_EDMC10}
    \end{center}
    \vskip-1.5em
  \end{figure*}

\begin{figure*}[!t]
  \begin{center}
    \begin{tabular}[b]{P{.45\linewidth}P{.45\linewidth}}
    EDM + EDM Sampler (256 steps) &  {\bf Ours} + EDM Sampler (80 steps) \\[1pt]
       \includegraphics[width=1.\linewidth]{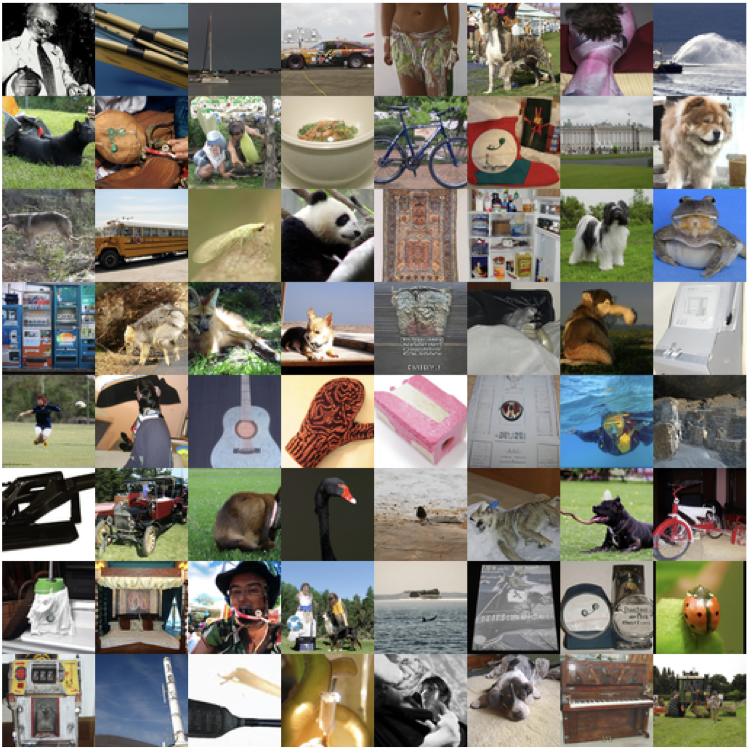} & 
       \includegraphics[width=1.\linewidth]{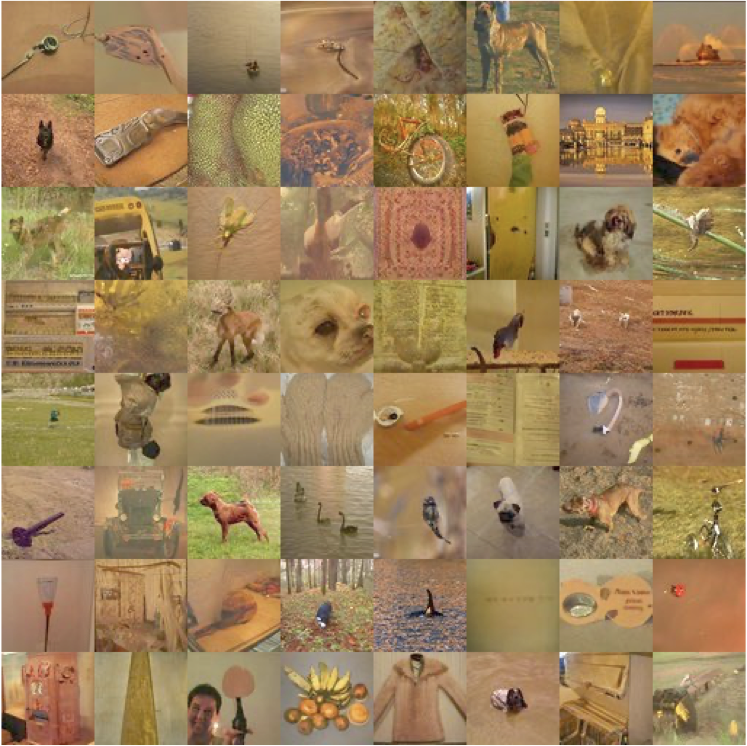} 
        \end{tabular} 
    \caption[]{(\includegraphics[height=0.009\textheight]{Rgb.png} Color online)~Samples generated by the proposed discriminator-guided Langevin diffusion, compared against the baseline EDM approach proposed by~\citet{EDM22}, on the ImageNet-64 dataset, using the EDM sampler, with sampling parameters as described by~\citet{EDM22} for the baseline. The baseline model requires 256 iterations, while the proposed discriminator-guided Langevin sampler converges in about 80 steps. The images generated by discriminator-guided Langevin diffusion lack significant color diversity, but were obtained entirely from kernel-guided sampling, without the need for training a score network. The issue of lack of sufficient color diversity on ImageNet-64 dataset requires further investigation.}
    \label{Fig_EDMImageNet}
    \end{center}
    \vskip-2.5em
  \end{figure*}

{\it {\bfseries Experiments with the EDM Sampler}}: Since the proposed approach suggests the interoperability of the score and the discriminator-kernel gradient in Langevin flow, we also consider discriminator-guided Langevin sampling on the CIFAR-10 and ImageNet-64 datasets, considering EDMs as the baseline~\citep{EDM22}. In both the scenarios, we also replace the sampler in discriminator-guided Langevin diffusion with the one used for the baseline considered by~\citet{EDM22}. We replace the score with the gradient of the polyharmonic kernel discriminator, with a constant coefficient, and ignore the exploratory noise term in our approaches. Images generated by the proposed method, with side-by-side comparisons with the baseline EDM are provided in  Figures~\ref{Fig_EDMC10}-\ref{Fig_EDMImageNet}). For CIFAR-10, we consider the second-order Heun sampler with 128 sampler steps in the baseline, while the proposed approach converges in 40 steps. For ImageNet-64, the baseline EDM sampler took 255 steps, while discriminator-guided Langevin diffusion took 80 steps to converge. \par

\begin{figure*}[!t]
\begin{center}
  \begin{tabular}[!t]{P{.01\linewidth}|P{.9\linewidth}}
   \rotatebox{90}{\qquad\qquad LDM} &\includegraphics[width=0.99\linewidth]{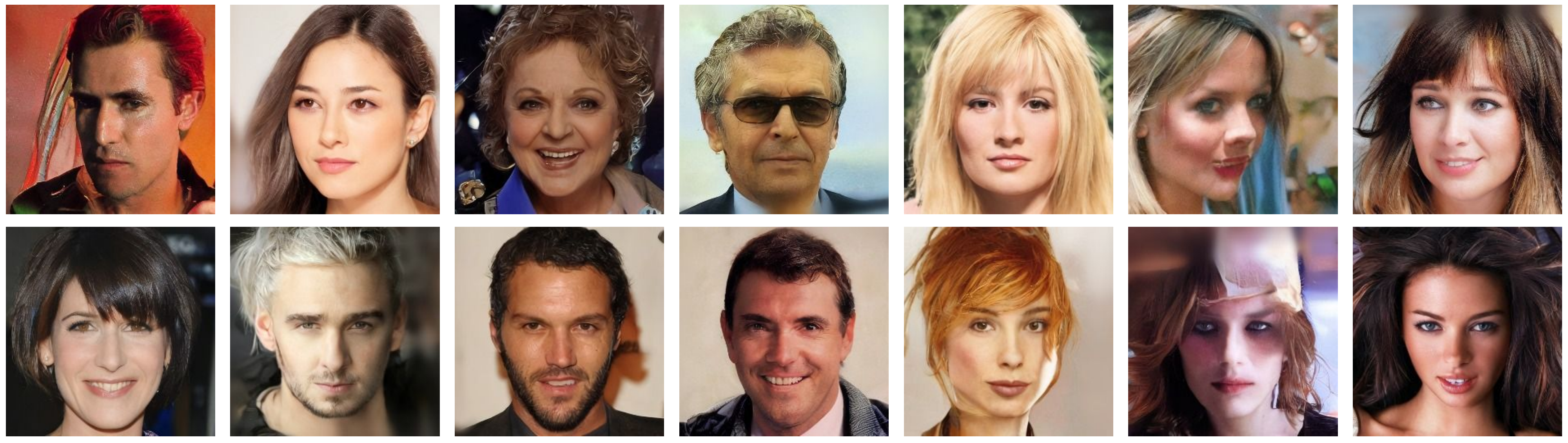}   \\[0.25pt] \midrule \\[-10pt]
   \rotatebox{90}{\qquad LDM+DG$_{\theta}$} &\includegraphics[width=0.99\linewidth]{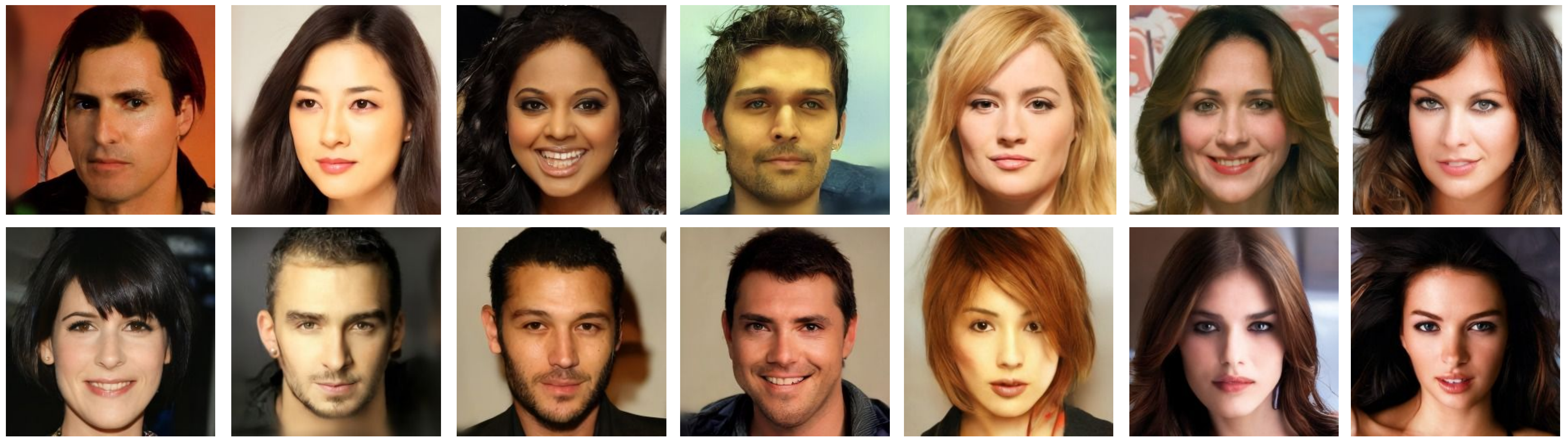}   \\[0.25pt] \midrule \\[-10pt]
    \rotatebox{90}{\quad LDM+DG$^*$ \textbf{(Ours)}} &\includegraphics[width=0.99\linewidth]{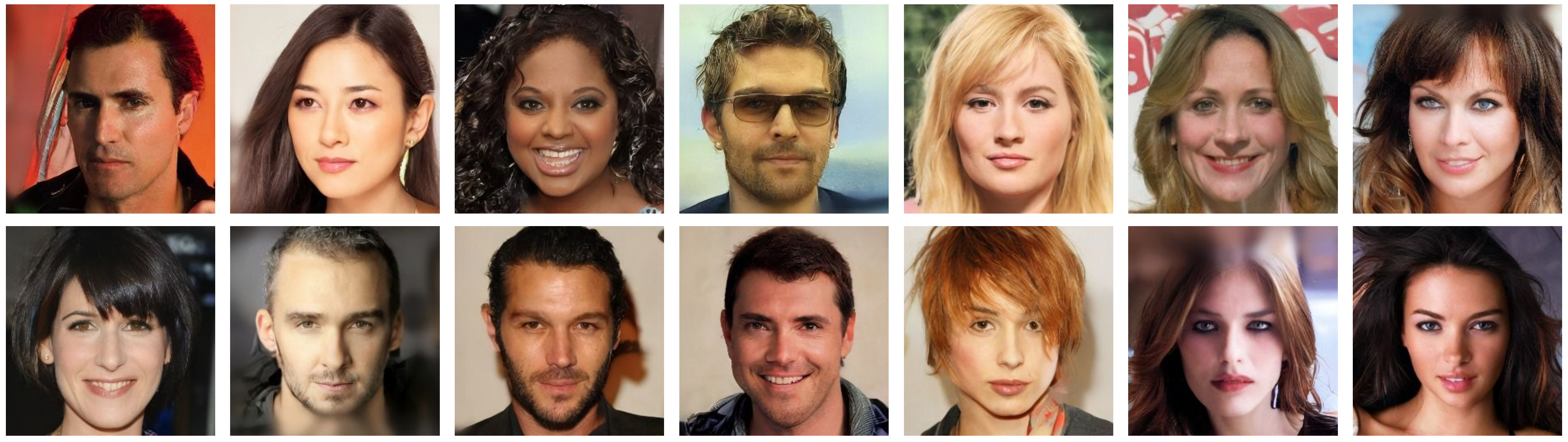}   \\[0.25pt] \midrule \\[-10pt]
    \rotatebox{90}{\qquad WANDA \textbf{(Ours)}} &\includegraphics[width=0.99\linewidth]{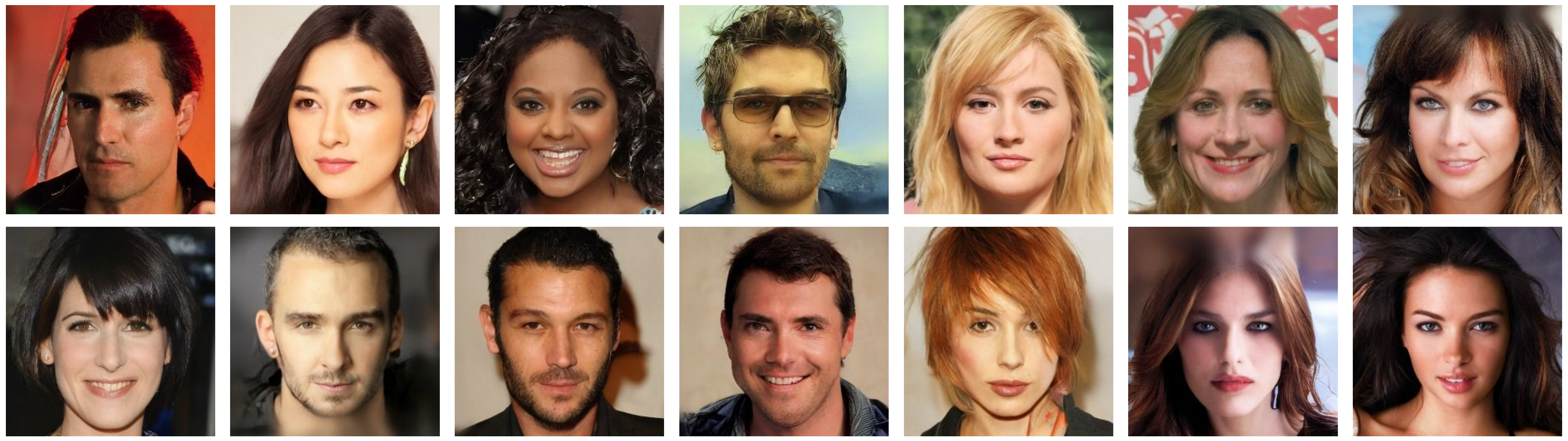} 
  \end{tabular} 
\caption[]{A comparison of the 256-dimensional CelebA-HQ images generated (given the same input) by the baseline latent diffusion model (LDM), and the proposed closed-form discriminator guidance models with and without time-step-shifted sampling (WANDA and LDM-DG$^*$, respectively). Images generated by LDM+DG$_{\theta}$ are oversmooth. The discriminator guidance in LDM-DG$^*$ significantly improves the quality of the images generated, by removing artifacts. WANDA is capable of generating images with a quality comparable to that of LDM-DG$^*$, with relatively fewer function evaluations.}
\label{Fig_CelebA_Full}  
\end{center}
\vskip-1em
\end{figure*}

\begin{figure*}[!t]
\begin{center}
  \begin{tabular}[!t]{P{.01\linewidth}|P{.9\linewidth}}
   \rotatebox{90}{\qquad\qquad LDM} &\includegraphics[width=0.99\linewidth]{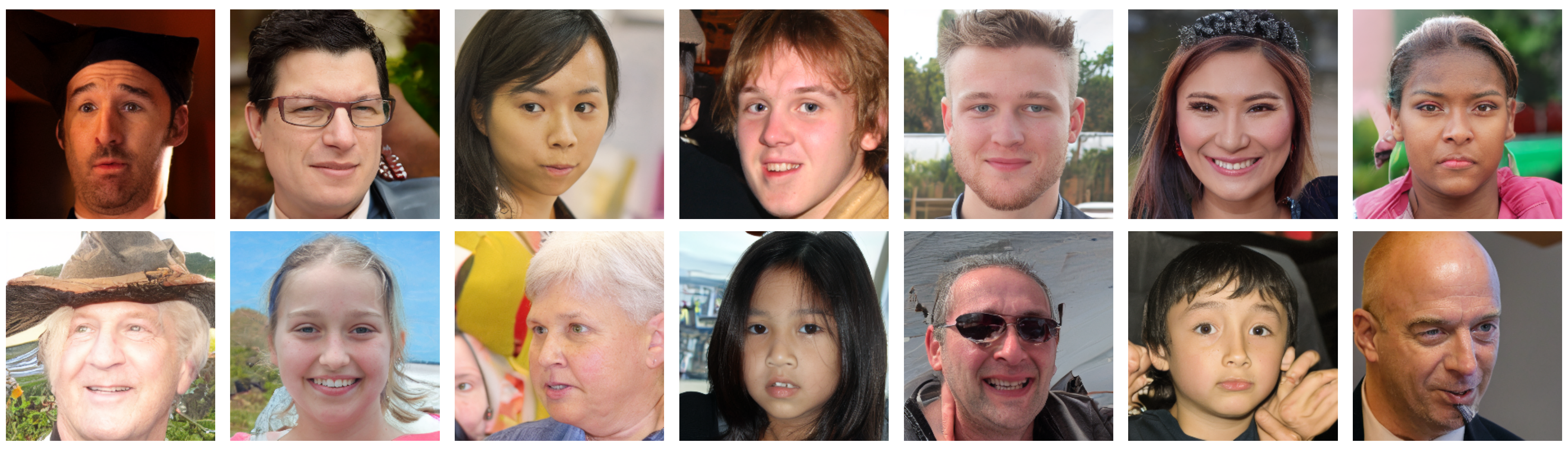}   \\[0.25pt] \midrule \\[-10pt]
   \rotatebox{90}{ LDM+DG$^*$ (Lin. Decay)} &\includegraphics[width=0.99\linewidth]{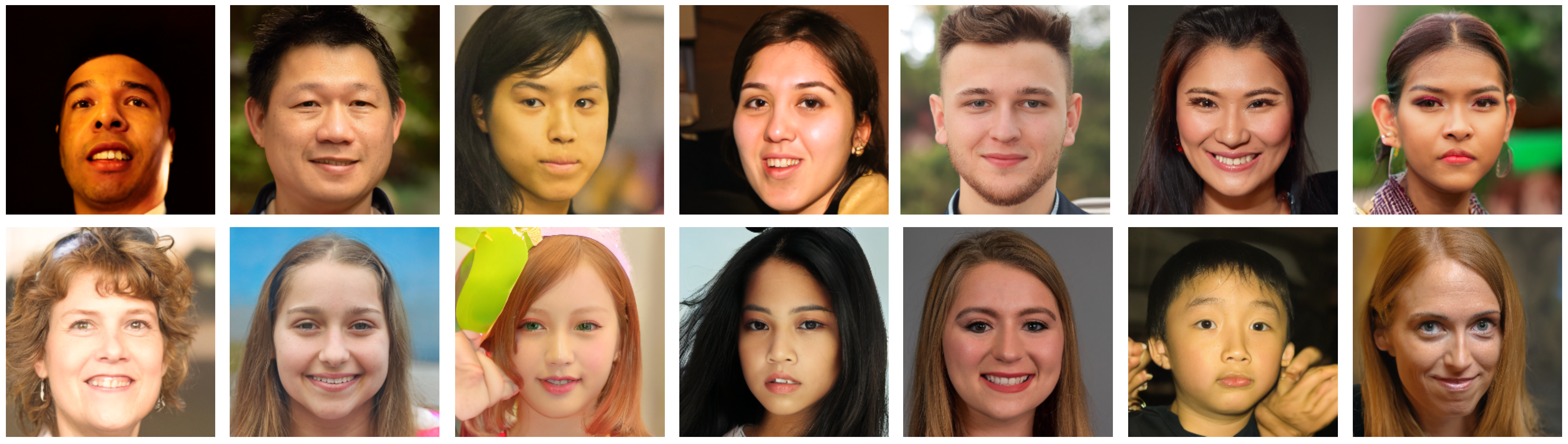}   \\[0.25pt] \midrule \\[-10pt]
    \rotatebox{90}{\quad LDM+DG$^*$ \textbf{(Ours)}} &\includegraphics[width=0.99\linewidth]{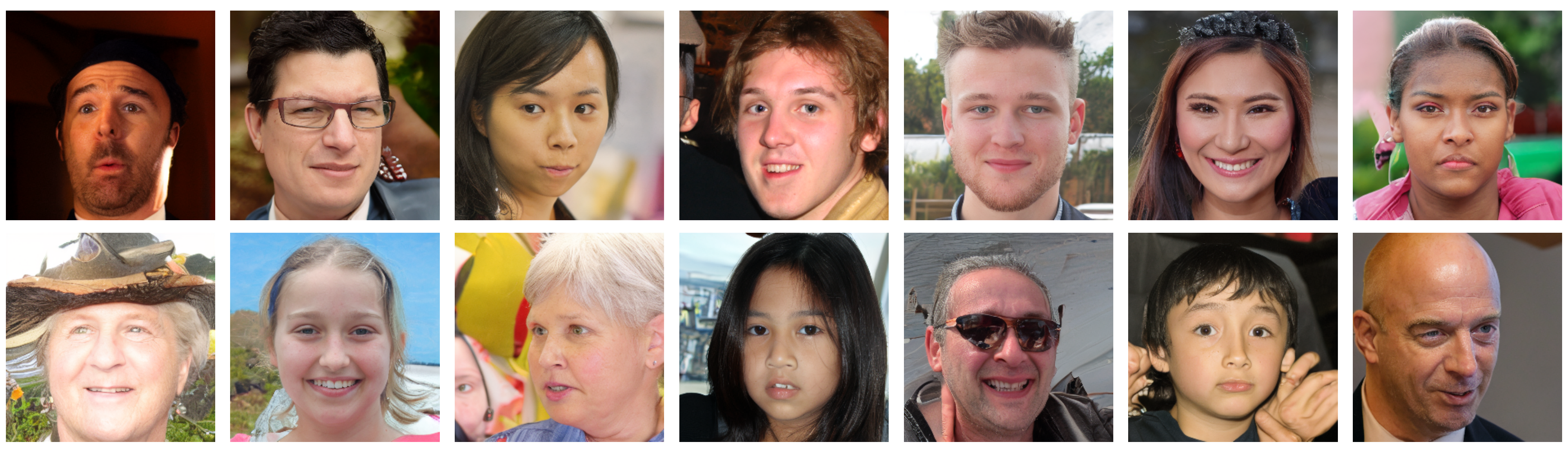}   \\[0.25pt] \midrule \\[-10pt]
    \rotatebox{90}{\qquad WANDA \textbf{(Ours)}} &\includegraphics[width=0.99\linewidth]{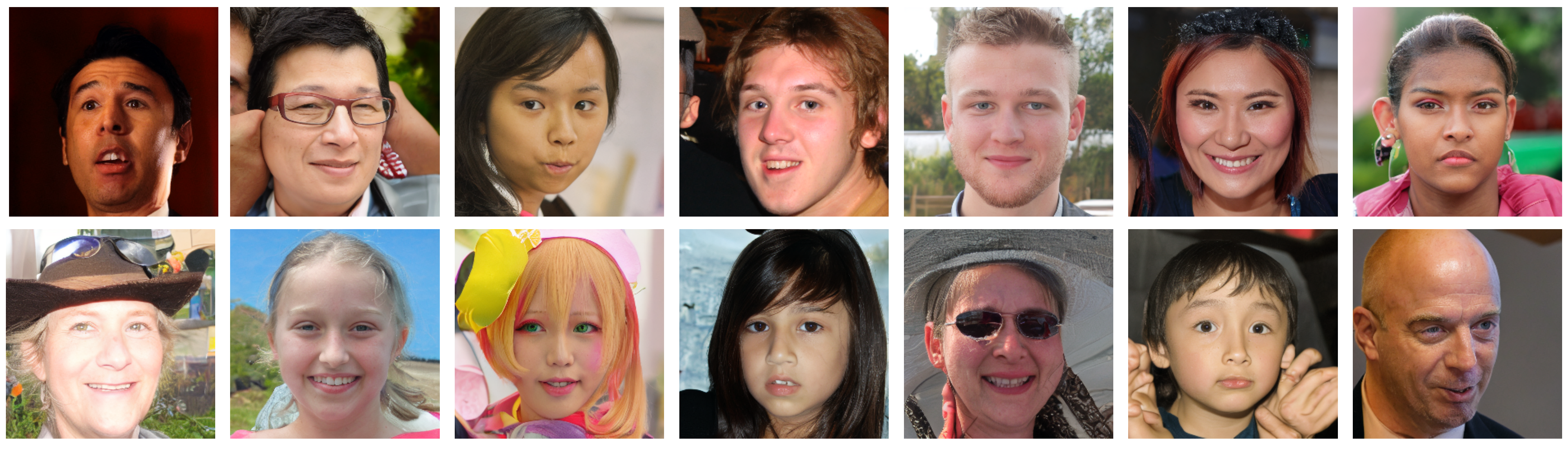} 
  \end{tabular} 
\caption[]{A comparison of the 256-dimensional FFHQ images generated (given the same input) by the baseline latent diffusion model (LDM), and the proposed closed-form discriminator guidance models with and without time-step-shifted sampling (WANDA and LDM-DG$^*$, respectively). Images generated by LDM+DG$^*$ with the linear decay (Lin. Decay) on \(w_{dg,t}\) are either oversmooth or have saturated colors, which we attribute to the discriminator guidance not decaying sufficiently fast. The discriminator guidance in LDM-DG$^*$ significantly improves the quality of the images generated, by removing artifacts. WANDA is capable of generating images with a quality comparable to that of LDM-DG$^*$, with relatively fewer function evaluations.}
\label{Fig_FFHQ_Full}  
\end{center}
\vskip-1em
\end{figure*}

\begin{figure*}[!t]
  \begin{center}
    \begin{tabular}[b]{P{.45\linewidth}P{.45\linewidth}}
    DPM+DG$^*$ (\textbf{Ours}) &  DPM \\[1pt]
       \includegraphics[width=1.\linewidth]{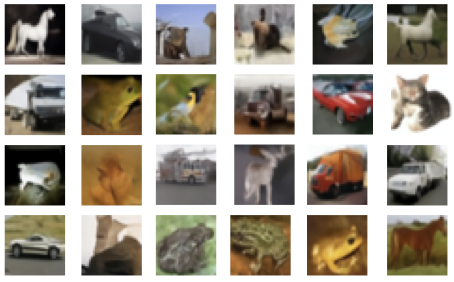} & 
       \includegraphics[width=1.\linewidth]{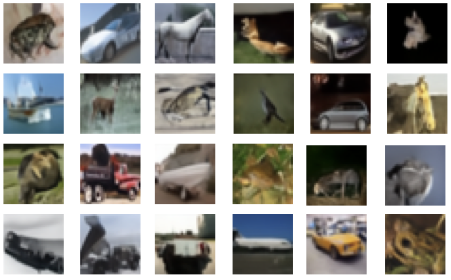} 
        \end{tabular} 
    \caption[]{(\includegraphics[height=0.009\textheight]{Rgb.png} Color online)~\revision{Samples generated by the proposed DPM+DG$^*$ sampler, compared against the DPM sampler on the CIFAR-10 dataset.}}
    \label{Fig_DPMC10}
    \end{center}
    \vskip-2.5em
  \end{figure*}

{\it {\bfseries Images for experiments presented in Section~\ref{Sec:LDM} of the Main Manuscript}}: Figures~\ref{Fig_CelebA_Full} and~\ref{Fig_FFHQ_Full} provide additional comparisons between the baseline and proposed LDM variants on the CelebA-HQ and FFHQ datasets, respectively. We also present images from CIFAR-10 in Figure~\ref{Fig_DPMC10}, when sampled using the DPM+DG$^*$ sampler.

\subsection{Additional Experimentation on LDM+DG$^*$}\label{App_Ablations}

{\it {\bfseries Ablations on discriminator weight \(\bm{w_{dg,t}}\)}}: To better understand the effect of the time-shifted diffusion, and the effect of the closed-form discriminator on generation performance, we perform ablations on the CelebA-HQ dataset. We ablate on the choice of the decay parameter, \(w_{dg,t}\) considering linear, exponential, and step-wise decay profiles. \revision{For the linear vs. exponential decay setting, considering LDM+DG$^*$, we found that exponential decay with  \(w_{dg,T} = 1.\) gave superior performance. Performance comparisons with a linear decay and \(w_{dg,T} = 0.1\), which leads to a comparable value for the weight as sampling completes (\textit{i.e.,} \(w_{dg,t}\) approach similar values in both cases, as \(t\rightarrow 0\)}). \par

{\it {\bfseries Comparisons against trainable discriminator guidance~\citep{DiscGuidance23}}}: We compare the performance of the LDM+DG$^*$ against a model wherein the discriminator is trained akin to the procedure described by~\citep{DiscGuidance23}. We employ a noise-embedded U-Net encoder with sigmoid activation as the discriminator that learns to classify the real and fake samples across all noise levels. The model is trained using the binary cross-entropy (BCE) loss. From Table~\ref{Table_Ablations}, we observe that the LDM model with the trained discriminator (LDM+D$_\theta$) either outperforms or is on par with the baselines. However, the trainable discriminator requires significantly more compute. On the contrary, the proposed LDM-DG$^*$ can be applied in a {\it plug-and-play} manner, with no additional training costs, and achieves a superior performance in terms of FID and KID metrics, compared to the LDM+D$_\theta$ sampler. \par

{\it {\bfseries Ablations on time step \(\bm{T_D}\)}}: We ablate on the time-step shifting algorithm with DG$^*$. We consider a sampling strategy wherein the discriminator is applied for the first \(T_D\) steps, and subsequently, transitioned to the base LDM sampler. We ablate over \(T_D\in\left\{50,100,200\right\}\). From the metrics shown in Table~\ref{Table_Ablations}, we observe that fewer discriminator steps lead to a superior performance. Empirically, this was found to be \(T_D^*\approx 50\). We observe that in the WANDA setting, there is a stark jump initially, of about 10 or so steps via the noise-variance-based time-step shifting. These observations show that DG$^*$ can be viewed as providing a quick high-quality transition at the initial iterations. \par

\begin{figure*}[!t]
\begin{center}
  \begin{tabular}[b]{P{.95\linewidth}}
     \includegraphics[width=0.95\linewidth]{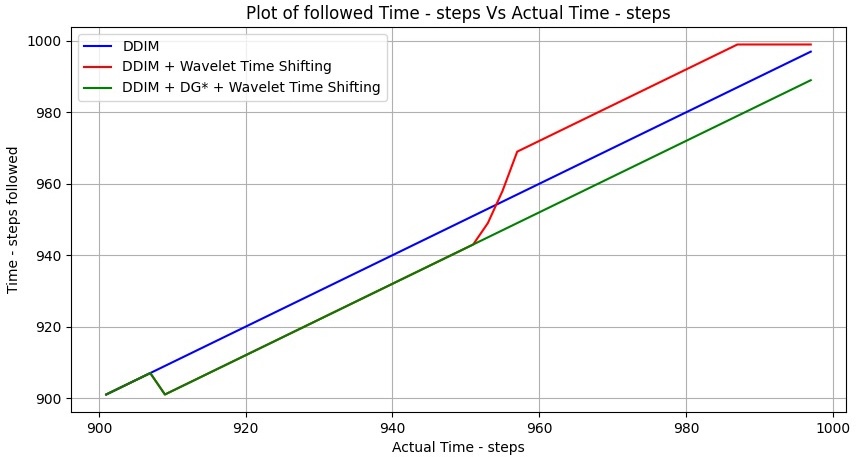} \\(a) \\\midrule
     \includegraphics[width=0.95\linewidth]{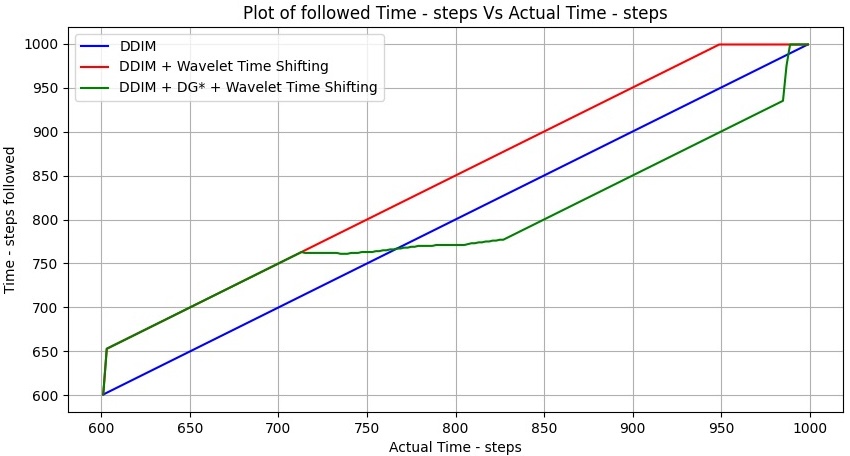}  \\
     (b) \\
      \end{tabular} 
  \caption[]{(\includegraphics[height=0.009\textheight]{Rgb.png} Color online)~\revision{A comparison of the predicted and actual time step $t$ in WANDA, and the baseline DDIM variants for (a) $T_D=900$ and (b) $T_D=600$, respectively, with $T=1000$. We observe that the the discriminator guidance term introduces a jump (a sharp drop in the \textit{time step followed} for the green curve) of 2-10\% of the steps is either setting.}}
  \label{Fig_Jumps}
  \end{center}
\end{figure*}

To analyze the choice of $T_D$, we perform additional ablations. First,  to further validate our choices, we perform an experiment wherein we plot the time-step jump predicted by the noise-variance-based time-shifted sampler at each step $t$. Since the step can occur at different $t$ for different images, we plot this curve. We performed the experiment over multiple images and observed that on average, the jump is about 2-10\% of the total steps. Illustrative plots of the predicted time vs the actual time $t$ of the iteration, wherein the discriminator guidance improves performance gains over the baseline time-shifting algorithm are provided in Figure~\ref{Fig_Jumps}. \par

{\it {\bfseries Choices of \(\bm{T, T_D}\) and \(\bm{w_{dg,t}}\) on FFHQ}}: We also perform additional ablations on DG$^*$, based on the choice of $T_D$ and $w_{dg,t}$ on the FFHQ dataset. The results are summarized in Table~\ref{Table_TDWAblations}. In summary, we observe that discriminator guidance performs best when run for less than 20\% of the overall iterations (*i.e.,* $T_D=5$ for $T=50$ or $T_D=5,10$ for $T=100$) and with the discriminator weight $w_{dg}\in(0.5,1)$. We observe similar trends when running discriminator guidance with the DPM solver, as seen in Tables~\ref{Table_DPMSolver}.

 \begin{table*}[!t]
  \fontsize{9}{12}\selectfont
 \begin{center}
\caption{\revision{Ablations of the proposed closed-form discriminator guidance for DPM Solver (DPM+DG$^*$) on the CelebA-HQ dataset, in terms of the Clean-FID, CLIP-FID and KID metrics. We observe that including discriminator guidance allows us to further accelerate the sample generation process, with the DPM+DG$^*$ sampler achieving comparable performance in $T=15$ (1 discriminator step with 14 DPM solver steps) steps, as the baseline DPM model with $T=20$. \(\ddagger\) denotes that the metric is computed via Clean-FID~\citep{CleanFID21}.}} 
 \label{Table_DPMSolver} 
 \begin{tabular}{P{0.05cm}|P{5.55cm}||P{1.6cm}|P{1.45cm}|P{1.8cm}}
 \toprule \toprule 
& Method &  Clean-FID\(\ddagger\) & CLIP-FID\(\ddagger\) & KID\(\ddagger\)  \\[2pt]
\midrule
\multirow{3}{*}{\rotatebox{90}{DPM}} & $T=20$ & 24.54 &  9.50 & \(0.0231\)  \\[2pt]
& $T=15$ & 26.63 &  10.07 & \(0.0262\)   \\[2pt]
\midrule
 \multirow{8}{*}{\rotatebox{90}{DPM+DG$^*$}}&\(T=20,~~T_D = 20,~~w_{dg} = 1.0\) & 24.10	 & 9.28  &  {\bfseries 0.0230} \\[2pt]
&\(T=20,~~T_D = 2,~~w_{dg} = 1.0\) & {\bfseries 24.07}	 & {\bfseries 9.22}  &  0.0235 \\[2pt]
&\(T=20,~~T_D = 2,~~w_{dg} = 0.5\) & 24.67	 & 9.28  & 0.0235 	\\[2pt]
&\(T=15,~~T_D = 1,~~w_{dg} = 1.0\) & 24.64 & 9.71 & 0.0233 \\[2pt]
&\(T=15,~~T_D = 1,~~w_{dg} = 0.5\) &  24.44 & 9.66 & 0.0232 \\[2pt]
&\(T=10,~~T_D = 1,~~w_{dg} = 1.0\) &  31.82 & 11.48 & 0.0320 \\[2pt]
&\(T=10,~~T_D = 1,~~w_{dg} = 0.5\) &  31.81 & 11.42 & 0.0328	\\[2pt]
\bottomrule\bottomrule
 \end{tabular}
 \end{center}
   \vskip-1em
 \end{table*}

 \begin{table*}[!t]
  \fontsize{9}{12}\selectfont
 \begin{center}
\caption{Ablations of the proposed closed-form discriminator guidance for LDM (LDM+DG$^*$) on the CelebA-HQ dataset. LDM+DG$^*$ with an exponential decay of the discriminator guidance weight performs the best, in terms of the Clean-FID, CLIP-FID and KID metrics. We also observe that fewer DG$^*$ steps leads to superior performance. Essentially, the DG$^*$ steps provide good initialization to the subsequent LDM sampling steps. \(\dagger\) denotes that the metric is computed via Torch Fidelity~\citep{TorchFid20}, and \(\ddagger\) denotes that the metric is  computed via Clean-FID~\citep{CleanFID21}.} 
 \label{Table_Ablations} 
 \begin{tabular}{P{3.75cm}||P{1.6cm}|P{1.45cm}|P{1.8cm}|P{1.25cm}|P{1.2cm}}
 \toprule \toprule 
Method &  Clean-FID\(\ddagger\) & CLIP-FID\(\ddagger\) & KID\(\ddagger\) & Precision\(\dagger\) & Recall\(\dagger\)  \\[2pt]
\midrule
LDM+DG$_{\theta}$~\citep{DiscGuidance23} & 21.44 &  7.08 & \(2.191\times10^{-2}\) & 0.5465 & 0.4420  \\[2pt]
 LDM+DG$^*$ (linear \(w_{dg,t}\)) & 31.68 & 10.99  &  \( 3.125\times10^{-2}\)	& 0.3602 & 0.5787  \\[2pt]
\midrule
 LDM+DG$^*$ \((T_D = 50)\) & {\bfseries20.49	}	 & {\bfseries6.48}  &  \({\bf 2.041\times10^{-2}}\)	& {\bf 0.4932} & 0.4806 \\[2pt]
WANDA \((T_D = 50)\)   & 22.76 & 7.98 & \(2.270\times10^{-2}\) & 0.4570 & 0.4990  \\[2pt]
WANDA \((T_D = 100)\) & 28.79 & 10.02	 & \(2.845\times10^{-2}\)	& 0.3574 & \textbf{0.5413} \\[2pt]
WANDA \((T_D = 200)\) & 37.83 & 12.64 & \(3.688\times10^{-2}\)	& 0.2030 & 0.5330 \\[2pt]
\bottomrule\bottomrule
 \end{tabular}
 \end{center}
   \vskip-1em
 \end{table*}

 \begin{table*}[!t]
  \fontsize{9}{12}\selectfont
 \begin{center}
\caption{\revision{Performance evaluation of WANDA, in terms of Clean-FID and CLIP-FID~\citep{CleanFID21} when ablations are carried out on the choice of the cut-off time \(T_D\) and guidance weight \(w_{dg}\). In general, we observe that, running discriminator guidance for about 10\% of the initial iterations, with the guidance weight \(w_{dg} \in (0.5,1)\) leads to the best performance. }} 
 \label{Table_TDWAblations} 
 \begin{tabular}{P{2cm}P{3.55cm}||P{1.6cm}|P{1.8cm}}
 \toprule \toprule 
\multicolumn{2}{c||}{Method} &  Clean-FID\(\ddagger\) & CLIP-FID\(\ddagger\)  \\[2pt]
\midrule
 \multirow{10}{*}{$T=50$} & Baseline & 12.95 &  3.78  \\[2pt]
 & $T_D = 50,~~w_{dg} = 25$ & 22.85 &  5.48  \\[2pt]
 & $T_D = 50,~~w_{dg} = 20$ & 19.92 &  5.01  \\[2pt]
 & $T_D = 50,~~w_{dg} = 10$ & 15.41 &  4.22  \\[2pt]
 & $T_D = 10,~~w_{dg} = 10$ & 15.37 &  4.18  \\[2pt]
 & $T_D = 5,~~w_{dg} = 10$ & 14.04 &  4.14  \\[2pt]
 & $T_D = 5,~~w_{dg} = 5$ & 12.79 &  3.90  \\[2pt]
 & $T_D = 5,~~w_{dg} = 2$ & 12.24 &  3.81  \\[2pt]
 & $T_D = 5,~~w_{dg} = 1$ & 12.13 &  3.79  \\[2pt]
 & $T_D = 5,~~w_{dg} = 0.5$ & {\bfseries 12.04} &  {\bfseries 3.72}  \\[2pt]
\midrule
 \multirow{8}{*}{$T=100$} & Baseline & 9.30 &  3.02  \\[2pt]
 & $T_D = 100,~~w_{dg} = 25$ & 15.37 &  4.16  \\[2pt]
 & $T_D = 100,~~w_{dg} = 15$ & 11.93 &  3.51  \\[2pt]
 & $T_D = 10,~~w_{dg} = 10$ & 10.70 &  3.26  \\[2pt]
 & $T_D = 10,~~w_{dg} = 5$ & 9.88 &  3.11  \\[2pt]
 & $T_D = 10,~~w_{dg} = 1$ & 9.39 &  3.06  \\[2pt]
 & $T_D = 5,~~w_{dg} = 5$ & 9.27 &  3.01  \\[2pt]
 & $T_D = 5,~~w_{dg} = 1$ & {\bfseries 9.07} &  {\bfseries 2.94} \\[2pt]
\bottomrule\bottomrule
 \end{tabular}
 \end{center}
   \vskip-1em
 \end{table*}

 \begin{table*}[!t]
  \fontsize{9}{12}\selectfont
 \begin{center}
\caption{\revision{Performance of LDM+DG$^*$ on the LSUN-Churches 256-dimensional dataset. \(\ddagger\) denotes that the metric is computed via Clean-FID~\citep{CleanFID21}.}} 
 \label{Table_LSUN} 
 \begin{tabular}{P{5.55cm}||P{1.6cm}|P{1.45cm}|P{1.8cm}}
 \toprule \toprule 
 Method &  Clean-FID\(\ddagger\) & CLIP-FID\(\ddagger\) & KID\(\ddagger\)  \\[2pt]
\midrule
 $T=200$ & 6.67 &  4.89 & \(0.0039\)  \\[2pt]
\midrule
\(T=200,~~T_D = 20,~~w_{dg} = 2.0\) & 6.99	 & 4.96  &  0.0044 \\[2pt]
\(T=200,~~T_D = 10,~~w_{dg} = 0.5\) & 6.43	 & 4.73 & 0.0037 	\\[2pt]
\(T=200,~~T_D = 10,~~w_{dg} = 0.1\) & {\bfseries 6.50} & {\bfseries 4.80} & {\bfseries 0.0032} \\[2pt]
\bottomrule\bottomrule
 \end{tabular}
 \end{center}
   \vskip-1em
 \end{table*}

\FloatBarrier

  \FloatBarrier

 \begin{figure*}[!t]
\begin{center}
  \begin{tabular}[b]{P{.35\linewidth}P{.35\linewidth}}
  {  Data Score: \(\DataScore{\x}\)} & {  IPM-GAN Discriminator Gradients } \\[1pt]
     \includegraphics[width=1.\linewidth]{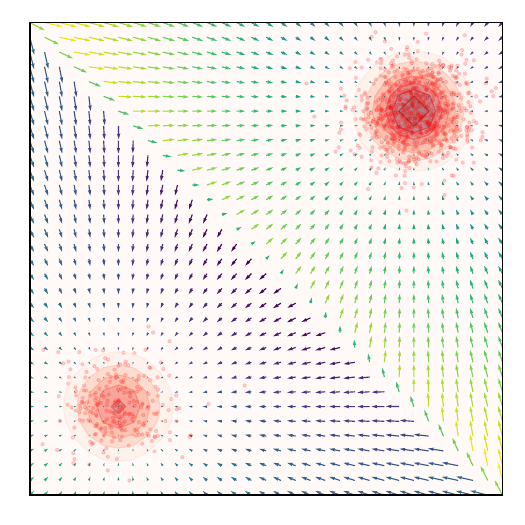} & 
     \includegraphics[width=1.\linewidth]{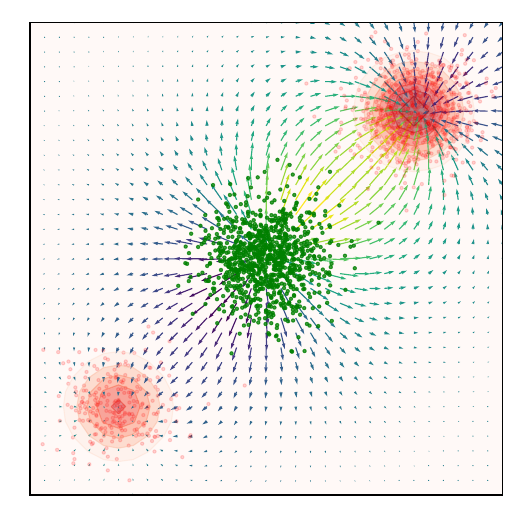}  \\
      \end{tabular} 
  \caption[]{(\includegraphics[height=0.009\textheight]{Rgb.png} Color online)~\revision{The loss landscape of the closed-form IPM-GAN discriminator, juxtaposed against the {\it (Stein) score} of the \textcolor{red}{target data}, for a Gaussian mixture \(\pd = \frac{1}{5}\mcalN(-5\bm{1}_2, \mbbI_2) + \frac{4}{5}\mcalN(5\bm{1}_2,\mbbI_2)\). The \textcolor{ForestGreen}{starting distribution}, \(p_T\) for the T-step diffusion process, is the standard normal Gaussian. All integral probability metric (IPM) minimizing GANs minimize the gradient field of the density difference \(\pd - \pg\) convolved with a kernel \(\kappa\), which corresponds to a kernel-convolved version of the score. The repulsive nature of the gradient field of the Discriminator improves stability and accelerated sampling in the proposed closed-form discriminator-guided diffusion.}}
  \label{Fig_ScoreDemo}
  \end{center}
\end{figure*}

\section{Discriminator Guidance with Time-Shifted Sampling} \label{App:NoiseVar}

\citet{TimeShift24} proposed the time-shifted sampler to mitigate \textit{exposure bias} in DPMs caused due to poor inference-time generalization, \textit{i.e.,} $\epsilon_{\theta}$ is trained on ground-truth samples $\x_{t}$, but inference is performed on $\hat{\x}_{t-1}$. Due to this discrepancy between training and generated samples, the exposure bias accumulates across the reverse process, causing it to divert from the intended trajectory. To mitigate this issue, given the sample $\hat{\x}_{t}$ an estimate of the noise variance in the image is used to evaluate a superior coupling time $t_{s}$ than the iteration's backward time $t$. Further, they also show that diffusion models basically contain \textit{two stages} -- The initial phase, wherein the input Gaussian distribution moves towards the image space, and the second phase, wherein patterns and structure emerge from latching onto a specific image to generate. Acceleration mechanisms such as time-step shifting (Li et al., 2024) and the proposed DG$^*$ operate in the first stage, which is why we focus the discriminator guidance to earlier iterations. 
Motivated by the above setting, and the observation in Section~\ref{Sec:DiscInfusion} that applying LDM+DG$^*$ for all time steps may be unnecessary, we adopt the time-shifted discriminator-guided diffusion strategy to ensure that the effect of discriminator guidance is restricted to the earlier, exploratory step. However, we observed that the noise-variance estimation technique proposed in the baseline was at a pixel-level sample estimate and could be improved. In particular,~\citet{MallatWavelets} and~\citet{Donoho95} showed that, in the context of image denoising, the noise variance can be estimated robustly using the Haar wavelet representation. The noise standard deviation is estimated as \(\Tilde{\sigma} = \frac{M_{\x}}{0.6745}\), wherein \(M_{\x}\) is the median of the absolute of the wavelet coefficients of the image \(\x\), and one level of decomposition suffices. The details are presented in  Appendix~\ref{App:NoiseVar}. We refer to the wavelet-based noise estimation for DG$^*$ guidance as WANDA.\par

To estimate the variance $\sigma^2$ of the noise $W[t]$ from the data $X[t] = W[t] + f[t]$ where $X[t]$ is $x_{t}$, we need to suppress the influence of $f[t]$. When $f$ is piecewise smooth, a robust estimator is calculated from the median of the finest-scale wavelet coefficients.  \\
A signal $X$ of size $N$ has $N/2$ wavelet coeffecients $\{\langle X,\psi_{l,m}\rangle\}_{0 \leq m < N/2}$ at the finest-scale $2^l = 2N^{-1}$. The coefficient $\mid \langle f,\psi_{l,m}\rangle \mid$ is small if $f$ is smooth over the support of $\psi_{l,m}$, in which case $ \langle X,\psi_{l,m}\rangle \approx \langle W,\psi_{l,m}\rangle$. In contrast, $\mid \langle f,\psi_{l,m}\rangle \mid$ is large if $f$ has sharp transitions in the support of $\psi_{l,m}$. A piece-wise regular signal has few sharp transitions, and thus produces a number of large coefficients that is small compared to $N/2$. At the finest scale, the signal $f$ thus influences the value of a small portion of large-amplitude coefficients $\langle X,\psi_{l,m}\rangle$ that are considered to be "outliers." All others are approximately equal to  $\langle W,\psi_{l,m}\rangle$, which are independent Gaussian random variables of variance $\sigma^2$. \par

A robust estimator of $\sigma^2$ is calculated from the median of $\langle X,\psi_{l,m}\rangle_{0 \leq m < N/2}$. The median of $P$ coefficients Med$(\alpha_{p})_{0 \leq p < P}$ is the value of the middle coefficient $\alpha_{n_0}$ of rank $P/2$. As opposed to an average, it does not depend on the specific values of coefficients $\alpha_{p} \geq \alpha_{n_0}$. If $M$ is the median of the absolute value of $P$ independent Gaussian random variables of zero mean and variance $\sigma^2_{0}$, then one can show that

\begin{align}
E\{X\} \approx 0.6745\sigma_{0}
\label{eqn: median1}
\end{align}

The variance $\sigma^2$ of the noise $W$ is estimated from the median $M_{X}$ of $\{\langle X,\psi_{l,m}\rangle\}_{0 \leq m < N/2}$, by neglecting the effect of $f$:

\begin{align}
    \Tilde{\sigma} &= \dfrac{M_{X}}{0.6745}
\end{align}

Indeed, $f$ is responsible for few large-amplitude outliers, and these have little impact on $M_{X}$.

\begin{figure*}[!th]
  \begin{center}
    \begin{tabular}[b]{c|c}
    \(\x_T\) & \(k\)-nearest neighbors of \(x_T\) (\(k = 9\))  \\[5pt]
            \includegraphics[height=0.95\linewidth]{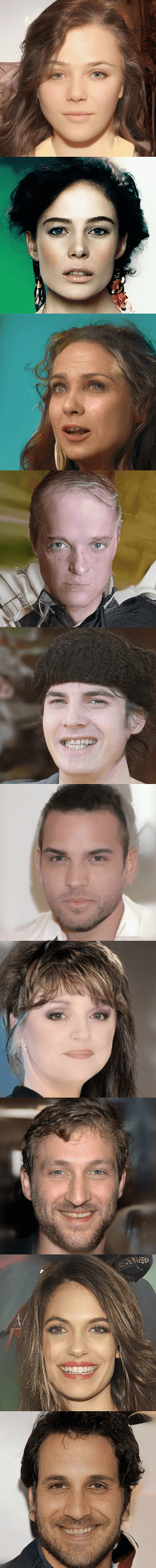}   &
            \includegraphics[height=0.95\linewidth]{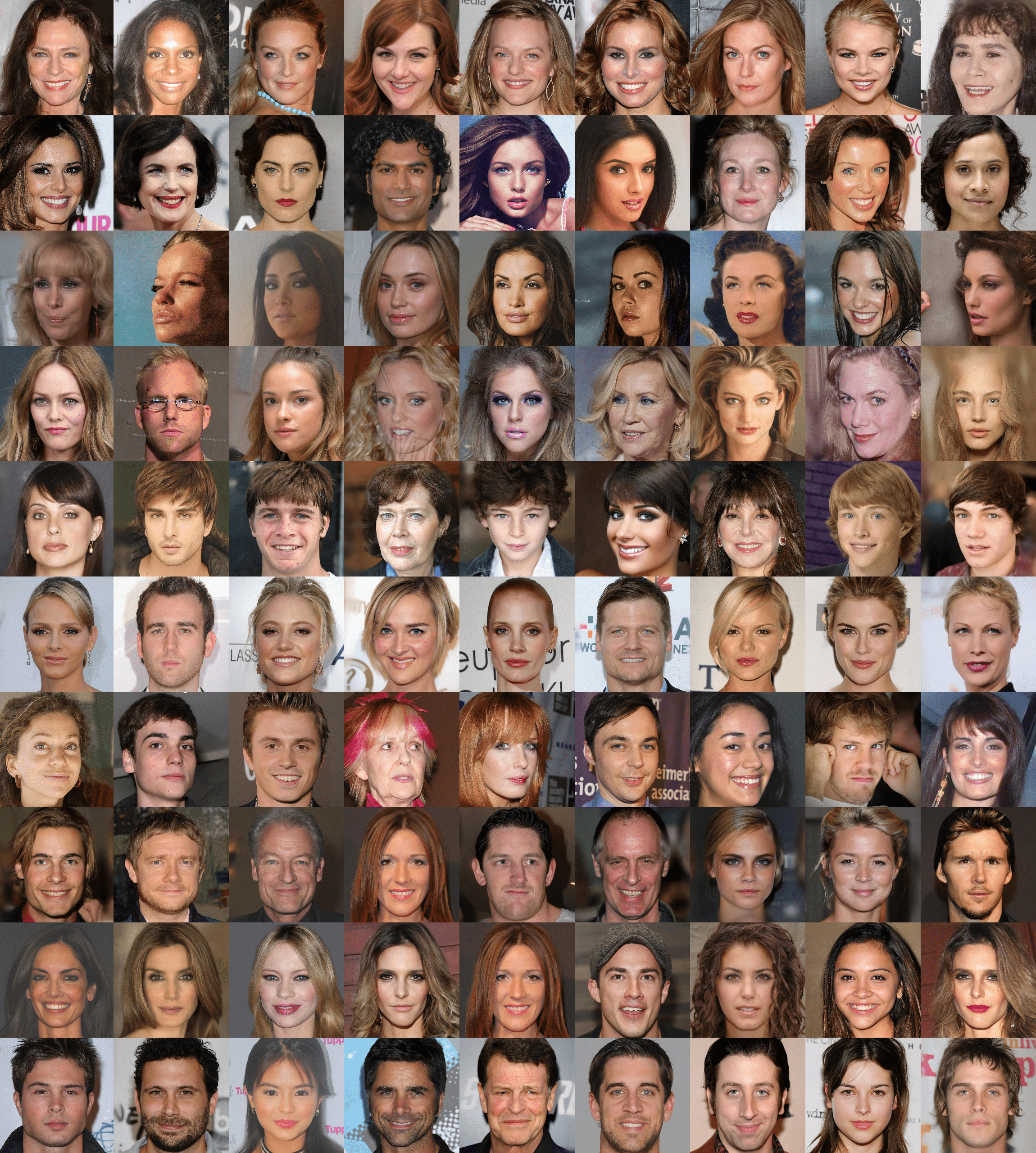}   \\
    \end{tabular} 
    \caption[]{(\includegraphics[height=0.012\textheight]{Rgb.png} Color online)~\revision{The \(k\)-nearest neighbor (kNN) test performed on images generated by the discriminator-guided DPM sampler, on the CelebA-HQdataset. The generated images are unique and distinct from the top-9 neighbors drawn from the target dataset, which suggests that the proposed approach does not memorize data.}} 
    \vspace{-1.2em}
    \label{Fig_CelebAHQ_KNN}  
    \end{center}
  \end{figure*}

  \begin{figure*}[!th]
  \begin{center}
    \begin{tabular}[b]{P{.45\linewidth}|P{.45\linewidth}}
            \includegraphics[height=0.95\linewidth]{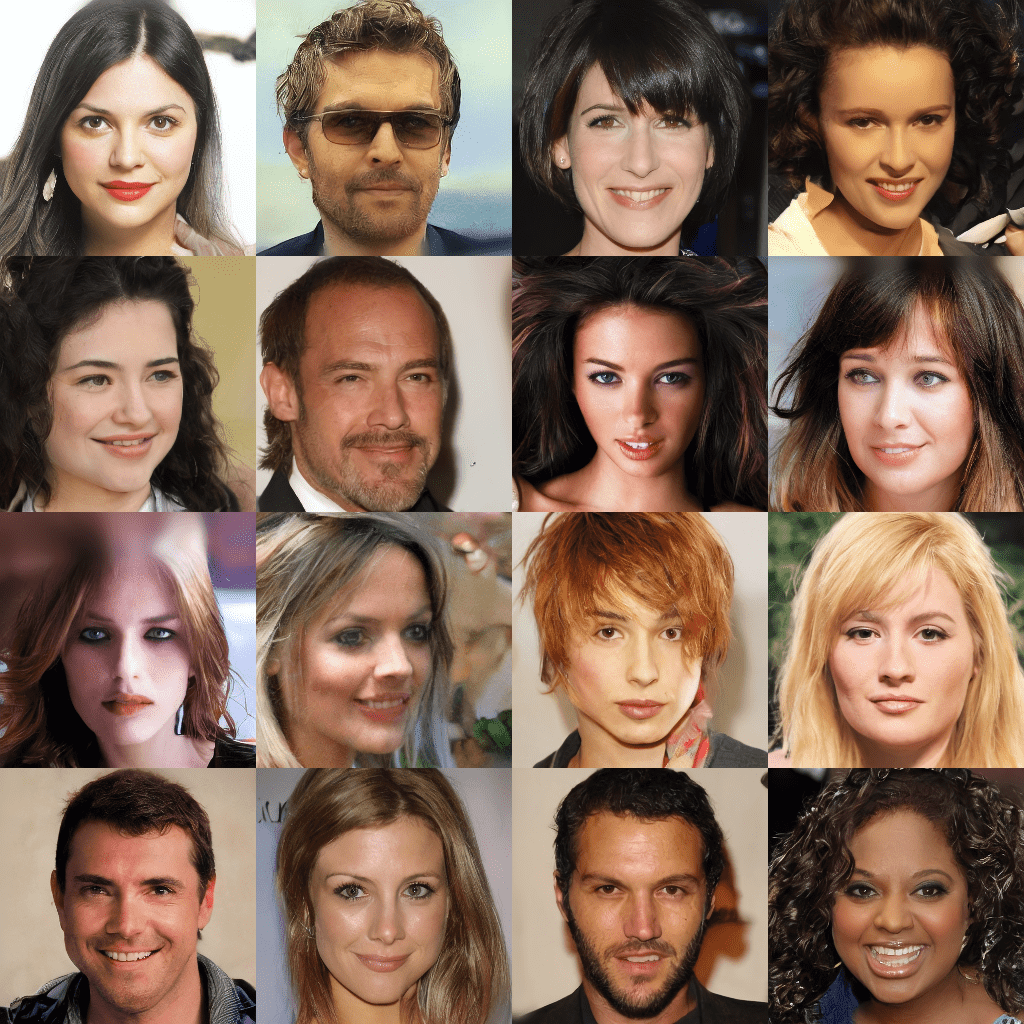}   &
            \includegraphics[height=0.95\linewidth]{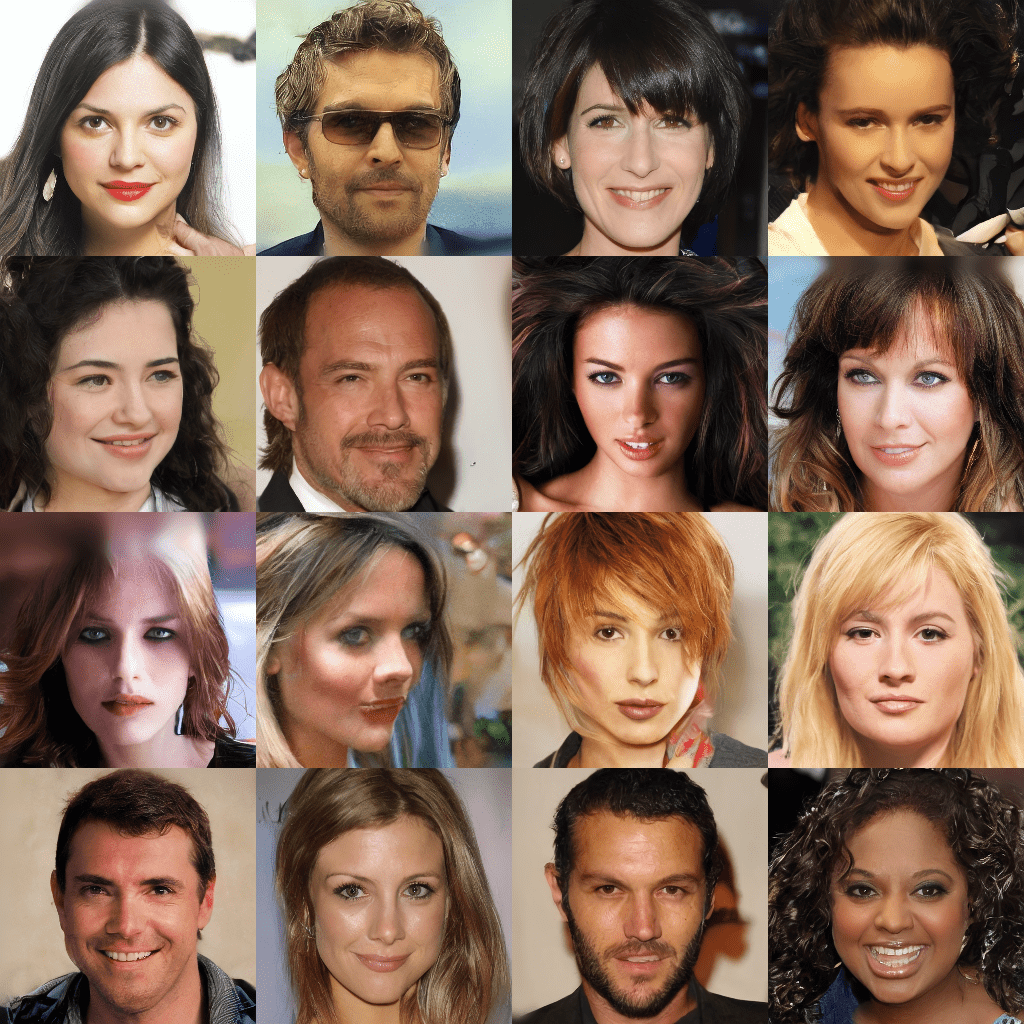}   \\
            \(M=16\) & \(M=25\) \\ \midrule
            \includegraphics[height=0.95\linewidth]{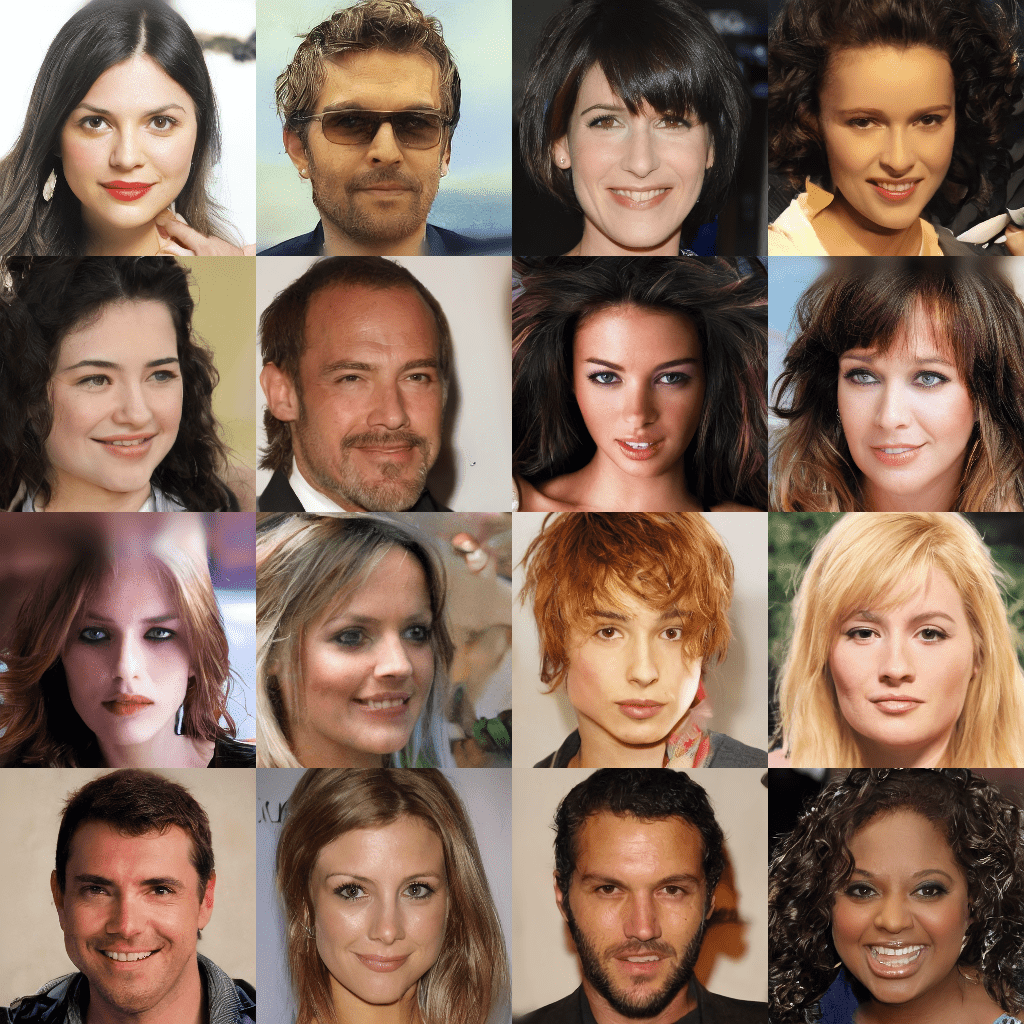}   &
            \includegraphics[height=0.95\linewidth]{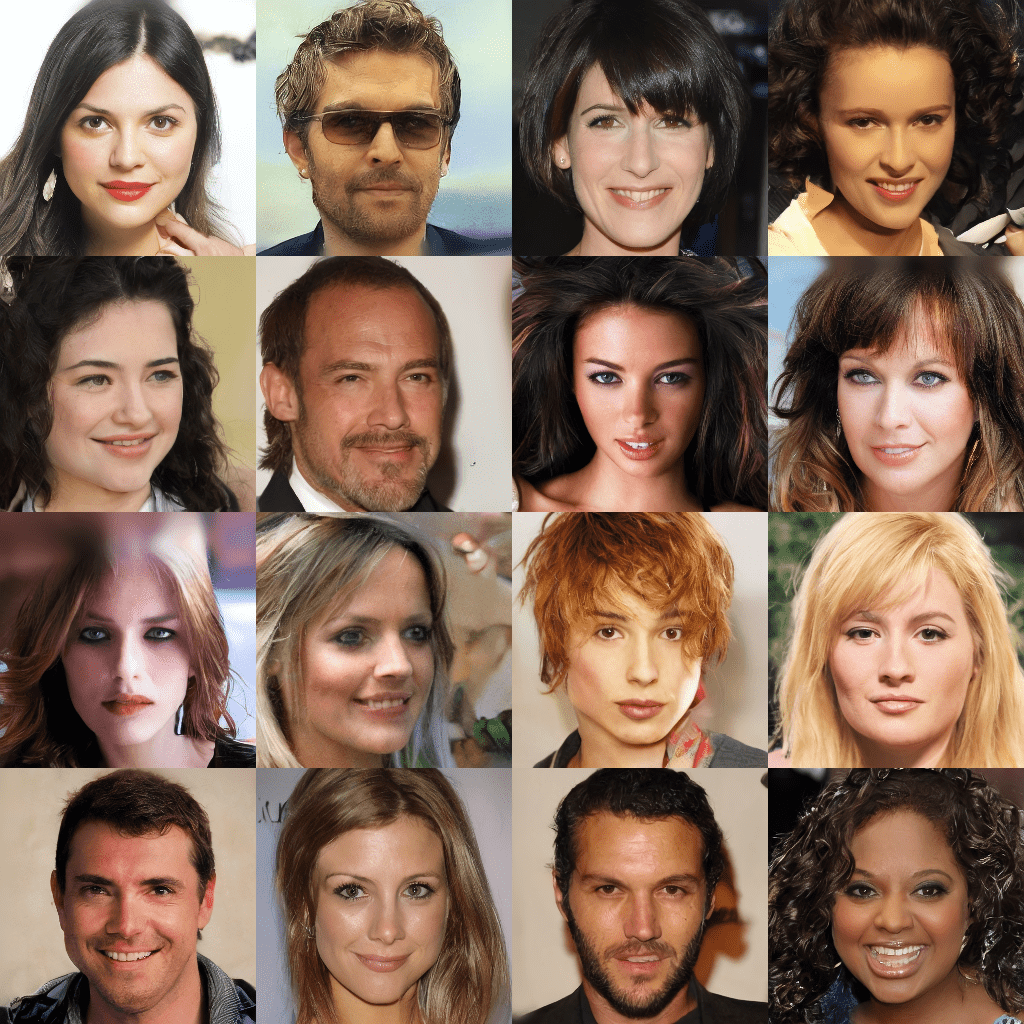}   \\
             \(M=50\) & \(M=100\) \\
    \end{tabular} 
    \caption[]{(\includegraphics[height=0.012\textheight]{Rgb.png} Color online)~\revision{A comparison of the images generated for varying numbers of centers \(M\) considered in the closed-form discriminator. We observe that the performance is generally unaffected by this choice, and using \(M=50\) is preferred, to ensure statistically, that the sample estimates converge.}} 
    \vspace{-1.2em}
    \label{Fig_CelebAHQ_M}  
    \end{center}
  \end{figure*}

\end{document}